\documentclass{article} 
\usepackage{iclr2027_conference,times}

\usepackage{amsmath,amsfonts,bm}

\def\eqref#1{equation~\ref{#1}}

\def\1{\bm{1}}

\DeclareMathAlphabet{\mathsfit}{\encodingdefault}{\sfdefault}{m}{sl}
\SetMathAlphabet{\mathsfit}{bold}{\encodingdefault}{\sfdefault}{bx}{n}

\usepackage{hyperref}
\usepackage{url}
\usepackage{cleveref}

\usepackage{booktabs}
\usepackage{multirow}
\usepackage{amssymb}

\usepackage[normalem]{ulem}
\usepackage{tabularx}
\usepackage{array}
\usepackage{makecell}
\usepackage{float}
\usepackage{rotating}
\usepackage{pifont}
\usepackage{subcaption}
\usepackage[utf8]{inputenc} 
\usepackage{soul} 
\usepackage[T1]{fontenc}    
\usepackage{hyperref}       
\usepackage{url}            
\usepackage{booktabs}       
\usepackage{amsfonts}       
\usepackage{nicefrac}       
\usepackage{microtype}      
\usepackage{xcolor}         
\usepackage{adjustbox}

\definecolor{obj}{RGB}{220,245,255}    
\definecolor{cons}{RGB}{235,220,255}   

\newcommand{\first}[1]{\colorbox{green!25}{#1}}
\newcommand{\second}[1]{\colorbox{blue!15}{#1}}
\newcommand{\third}[1]{\colorbox{gray!20}{#1}}

\newcommand{\worst}[1]{\textcolor{red}{\uline{\textcolor{black}{#1}}}}

\title{CityDeploy-Bench: Benchmarking Physics-Grounded Spatial Set Planning for Multi-Transmitter Network Deployment}

\author{%
\begin{tabular}[t]{@{}l@{\hspace{3em}}l@{}}
\textbf{Chenyang Yuan}\textsuperscript{*}
&
\textbf{Xiaoyuan Cheng}\textsuperscript{*} \\
\mdseries University of Sheffield
&
\mdseries University College London \\
\mdseries\texttt{cyuan7@sheffield.ac.uk}
&
\mdseries\texttt{ucesxc4@ucl.ac.uk} \\[0.3em]
\multicolumn{2}{@{}l@{}}{%
\normalfont\footnotesize
\textsuperscript{*}Equal contribution.}
\end{tabular}}

\iclrfinalcopy
\begin{document}

\maketitle

\begin{abstract}
Automating city-scale wireless deployment remains challenging  under complex urban propagation and network-wide interference.
We introduce \textbf{CityDeploy-Bench}, a benchmark that reframes multi-transmitter deployment as \emph{physics-grounded spatial set planning} under a unified ray-tracing verifier.
The benchmark separates utility representation from planning dynamics, enabling controlled comparison between direct scalar rewards, relational models, and higher-order interaction structures across diverse planners.
Our experiments reveal a clear transition in planning behavior as physical coupling grows.
Deployment quality becomes increasingly dependent on whether the learned utility captures collective transmitter interactions, whereas stronger search alone cannot compensate for missing relational structure.
This establishes multi-transmitter deployment as a coordination problem over physically interacting sets rather than a collection of independent spatial decisions.
We release CityDeploy-Data and the benchmark framework as a reproducible testbed for research linking decision learning with physically grounded wireless network design.
Code and implementation details are available at
\url{https://github.com/Chenyang-Yuan/CityDeploy-Bench}.
\end{abstract}

\begin{figure}[H]
    \centering
    \includegraphics[width=\linewidth]
    {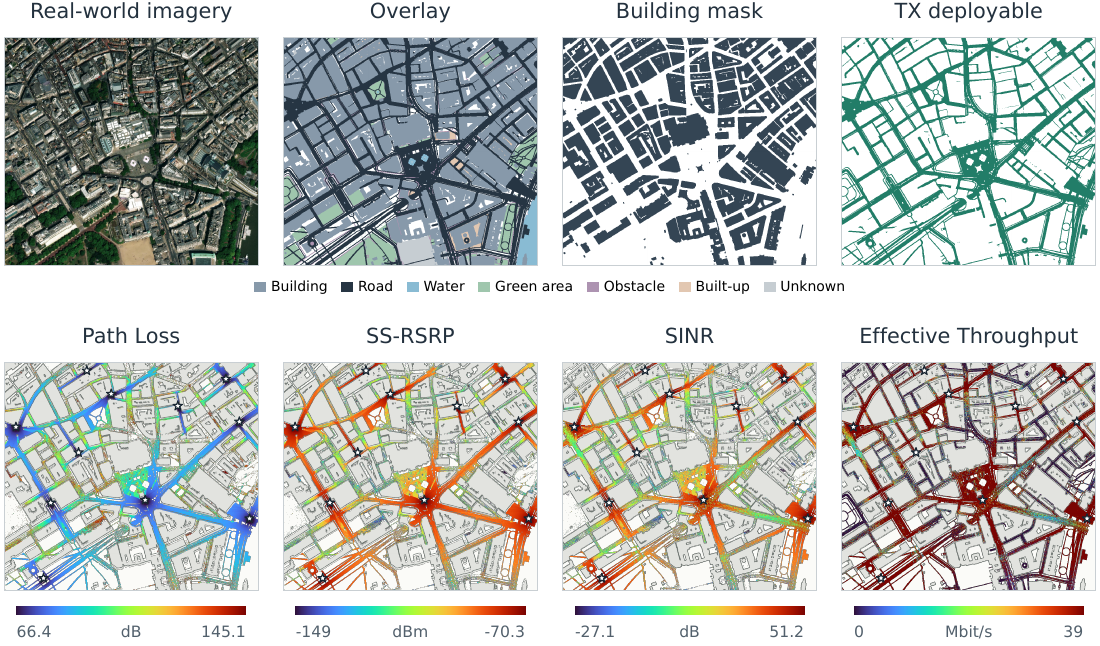}
    \caption{
    Trafalgar Square illustrates the central challenge of CityDeploy-Bench: effective multi-transmitter deployment requires coordinated decisions across the complete transmitter set under geometry-induced propagation and interference coupling.
    }
    \label{fig:citydeploy_teaser}
\end{figure}

\section{Introduction}

Wireless service in cities is governed by a strongly structured physical environment.
Buildings, roads, materials, visibility, and multipath propagation jointly determine where transmitters (TXs) can provide useful coverage and how simultaneously active TXs interfere with one another~\citep{levie2020radio,tian2021radionet}.
Designing a multi-transmitter (multi-TX) deployment therefore requires more than placing individually strong sites: the complete TX set must be coordinated with the surrounding urban geometry and with the service requirements imposed on the network.
In practice, such deployments are commonly designed through expert-driven iteration and repeatedly validated with ray-tracing tools, resulting in substantial engineering and computational cost~\citep{li2022analytics}.
Recent advances in machine learning and open-source differentiable or programmable ray tracing make automatic wireless planning increasingly feasible~\citep{hoydis2023sionna,Khan2022Digital}.

We view this problem through the lens of decision learning and formulate city-scale multi-TX deployment as \emph{physics-grounded spatial set planning}.
Given an urban scene, a planner must generate an unordered set of TX locations whose joint operation satisfies prescribed service requirements after physical verification.
This perspective differs from independent site selection.
The utility of one TX depends on the remaining deployment through serving-cell reassignment, coverage redundancy, and co-channel interference.
As deployment density grows, planning therefore changes from locating individually useful sites into coordinating a physically coupled set.
This set-level structure makes urban deployment a natural testbed for studying how learned utility representations and search algorithms interact under a high-fidelity physical evaluator.

A growing body of work has explored automatic wireless planning using optimization, learning, and generative reasoning.
Black-box and population-based optimizers search deployment configurations directly~\citep{eriksson2020scalable,Bonyadi2017Evolutionary,Deb2014GA,Qiu2024Large};
Monte Carlo methods construct reward-induced target distributions and explore them through importance weighting, sequential sampling, or Hamiltonian dynamics~\citep{piche2018probabilistic,naesseth2014sequential,lioutas2023criticsequentialmontecarlo,hoffman2014no,cardoso2022br};
and diffusion- or score-based planners progressively refine noisy candidate configurations under learned rewards or constraints~\citep{ho2020denoising,janner2022planning,dong2024diffuserlite,pan2024model,akhound2024iterated,cheng2025safe}.
Large language models have also recently been applied to wireless planning and engineering workflows under specific task formulations~\citep{Quan2025Large,li2026automatic,Modesto2026LoRaWAN,Deng2025TelePlanNet,Yuan2026LLM}.
However, these studies differ substantially in environment construction, infrastructure assumptions, physical verification, objective definitions, and evaluation metrics.
Consequently, it remains difficult to determine whether performance gains arise from the planner itself, the learned reward, the physical setting, or the evaluation protocol.

Two challenges are particularly important for a standardized benchmark.
First, \emph{spatial search is non-smooth and scene dependent}.
Urban geometry creates abrupt changes in visibility and propagation, so nearby TX coordinates can induce markedly different service responses~\citep{yuan2026planning}.
Multi-TX planning must therefore search a high-dimensional configuration space containing fragmented feasible regions and multiple locally attractive solutions.
Second, \emph{deployment utility is relational}.
A TX may improve path-loss or received-power coverage while simultaneously increasing interference or duplicating service already provided by other TXs.
The value of a deployment is thus a property of the complete set rather than the sum of independent site scores.
This raises a second question beyond search: how should the utility of an interacting TX set be represented?
Pairwise graph models provide one form of relational inductive bias~\citep{battaglia2016interaction,kipf2018neural}, while higher-order hypergraph formulations offer a natural extension when useful or harmful interactions involve multiple TXs jointly~\citep{feng2019hyper,Battiston2021physics,wang2023hyper}.

In this work, we introduce \textbf{CityDeploy-Bench}, a benchmark for physics-grounded spatial set planning in city-scale wireless networks. Figure~\ref{fig:citydeploy_overview} summarizes this pipeline,
from physics-grounded urban scenes and feasible TX sets to joint
physical evaluation and reward-guided spatial planning.
CityDeploy-Bench couples physically reconstructed urban scenes, ray-tracing-derived service responses, learned set-level utility models, and diverse planning algorithms under a common executable deployment interface.
Rather than evaluating search methods in isolation, the benchmark separates \emph{utility representation} from \emph{planning dynamics}, enabling controlled study of how relational structure and inference interact.
Our main contributions are:

\begin{itemize}
    \item[1] \textbf{A unified formulation of city-scale multi-TX deployment.}
    We formulate wireless deployment as physics-grounded spatial set planning over unordered TX configurations, with a common ray-tracing-based service utility and executable road-constrained action space.
    This establishes a standardized bridge between scenario-specific wireless engineering and learning-based decision making.

    \item[2] \textbf{CityDeploy-Data and structured utility representations.}
    We construct a geographically partitioned dataset of urban scenes and multi-TX deployments with dense physical responses and deployment-level service labels.
    On top of this interface, we benchmark pooled, pairwise, latent-relational, and higher-order hypergraph utility models, allowing the effect of reward structure to be separated from the effect of the planner.

    \item[3] \textbf{A broad benchmark of search and sampling dynamics.}
    We evaluate nine planners spanning annealed score-based search, Monte Carlo inference, and derivative-free optimization under the same learned utility interfaces and physical verification protocol.
    This design supports controlled comparison of representation--inference compatibility rather than treating each reward--planner combination as an unrelated method.

    \item[4] \textbf{Mechanistic findings on multi-TX planning.}
    Our experiments reveal that deployment difficulty is governed by physical coupling rather than TX count alone:
    densification can move a scene from a coverage-limited to an interference-limited regime;
    the useful relational order grows with coupling but is not universally maximal;
    reward representation and search dynamics exhibit strong compatibility effects;
    and high peak coverage does not necessarily imply reliable target attainment or infrastructure efficiency.

    \item[5] \textbf{Multi-faceted evaluation with physical verification.}
    We evaluate simulator-verified joint and marginal service quality, planning efficiency, infrastructure efficiency, seed stability, and geographic transfer/adaptation under consistent deployment constraints and radio assumptions.
    This provides a reproducible basis for studying learned planning beyond a single scene, objective, or optimization paradigm.
\end{itemize}

\section{Problem Formulation}

\subsection{Multi-TX Deployment and Service Utility}
\label{sec:deployment-service}

\paragraph{Deployment.}
Given an urban scene $S$, we seek an unordered set of $N$ transmitter (TX)
locations $P=\{p_i\}_{i=1}^{N}$ on the deployable road domain $\mathcal D_S$.
Feasible deployments satisfy
\begin{equation}
\mathcal F_N(S)
=
\left\{
P\subseteq\mathcal D_S:
|P|=N,\;
\|p_i-p_j\|_2\ge d_{\min}
\right\},
\label{eq:feasible-set}
\end{equation}
with $d_{\min}=20\,\mathrm{m}$.
Scene construction, coordinate conventions, and continuous-to-road mapping are
detailed in Appendices~\ref{app:data} and~\ref{app:planning}.

\paragraph{Physical response and service utility.}
For a feasible deployment, the ray-tracing operator
\begin{equation}
Y=F_{\mathrm{RT}}(S,P),\qquad
Y(x)=
\bigl(
\mathrm{PL},
\mathrm{SS\text{-}RSRP},
\mathrm{SINR},
T_{\mathrm{eff}}
\bigr)(x)
\label{eq:physical-response}
\end{equation}
combines strongest-power serving association with full-frequency-reuse
interference.
Over the road evaluation cells $\Omega_S$, we define joint-metric (JM) coverage as
\begin{equation}
U(S,P)
=
\frac{1}{|\Omega_S|}
\sum_{x\in\Omega_S}
\mathbf 1
\!\left[
Y(x)\text{ satisfies }\boldsymbol\tau
\right],
\label{eq:service-utility}
\end{equation}
where $\boldsymbol\tau$ denotes the fixed service thresholds.
For a prescribed TX cardinality $N$, the planning objective is
\begin{equation}
P_N^\star(S)
\in
\arg\max_{P\in\mathcal F_N(S)}
U(S,P).
\label{eq:deployment-objective}
\end{equation}
Because both serving association and interference depend on the complete TX set,
$U$ is inherently relational and need not improve monotonically with TX
cardinality.
The radio profile, service definitions, thresholds, and metric dependencies are
given in Appendix~\ref{app:radio}.

\begin{figure*}[t]
    \centering
    \includegraphics[width=0.99\linewidth]{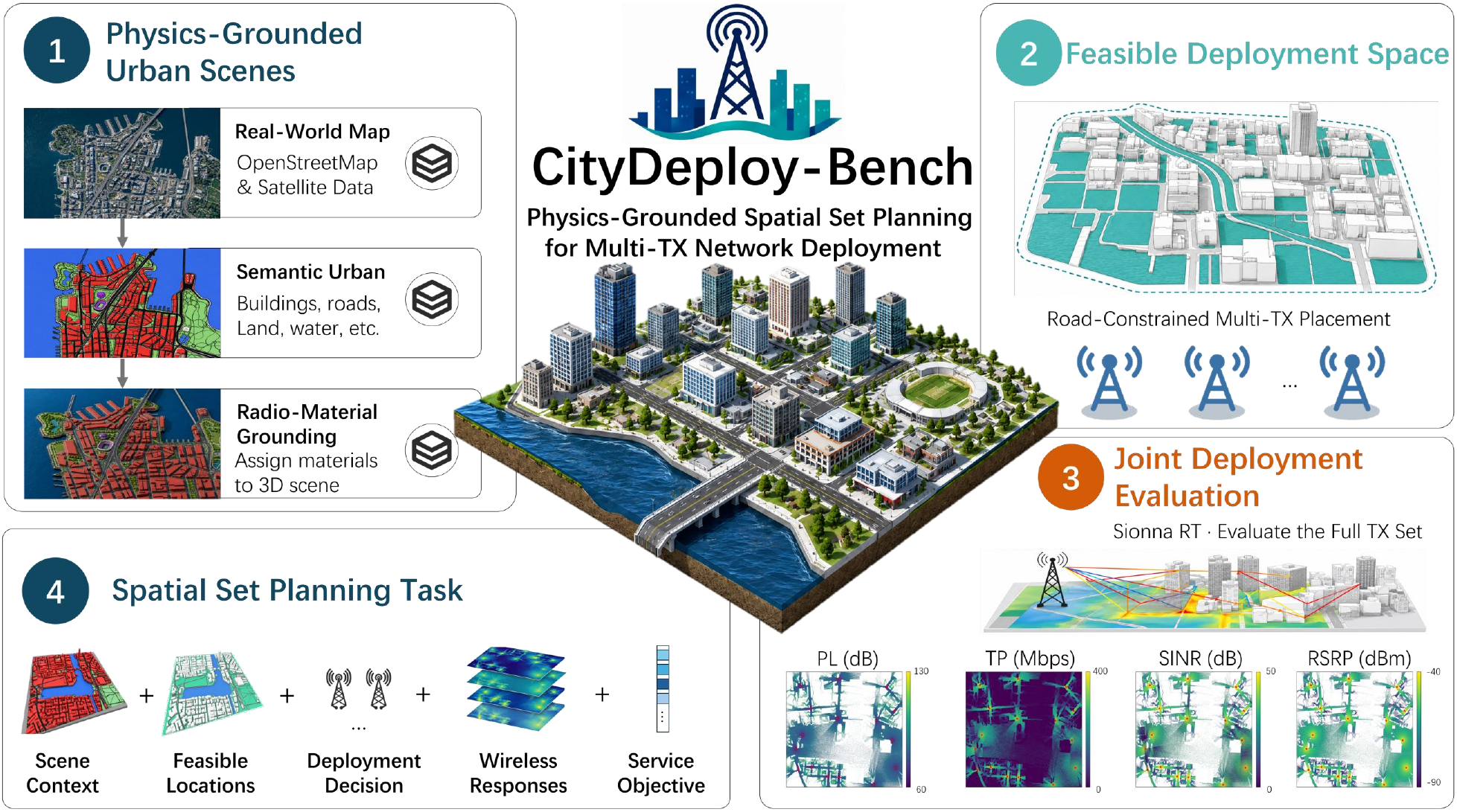}
    \caption{
\textbf{Overview of CityDeploy-Bench.}
(1) Geographic regions are converted into physics-grounded urban scenes. (2) Scene semantics define feasible multi-transmitter deployments. (3) Sionna RT evaluates the complete active transmitter set, accounting for propagation, serving association, and interference. (4) Spatial set planning selects transmitter locations under scene-specific constraints to maximize joint service coverage.
}
    \label{fig:citydeploy_overview}
\end{figure*}

\subsection{Learned Utility Representations}
\label{sec:utility-representations}

Direct evaluation of Eq.~\eqref{eq:deployment-objective} is expensive because
each utility query requires ray tracing.
We therefore learn a permutation-invariant surrogate
\begin{equation}
\widehat U_\theta(X_S,\mathbf u)
\approx
U(S,P),
\label{eq:learned-utility}
\end{equation}
where $X_S$ is the rasterized scene representation and
$\mathbf u=\{u_i\}_{i=1}^{N}$ denotes normalized TX coordinates.
The predictor is invariant to any reordering of the TX set.

We benchmark four utility representations with increasing relational structure.
The \emph{Regressor} pools independently encoded TX features without explicit
relations;
the Interaction Network (\emph{IN})~\citep{battaglia2016interaction}
introduces pairwise message passing;
the \emph{NRI} predictor~\citep{kipf2018neural} learns latent relation types;
and the Hypergraph Potential Energy Model (\emph{HPEM}) represents higher-order
TX interactions through subset potentials.

For HPEM, the predicted utility is parameterized by a learned energy,
\begin{equation}
\widehat U_\theta=-E_\theta,\qquad
E_\theta
=
E_{\mathrm{global}}
+
\sum_{k\in\mathcal K_N}
w_k
\underset{\substack{e\subseteq\mathbf u\\|e|=k}}
{\operatorname{Mean}}
E_{\theta,k}(X_S,e),
\label{eq:hypergraph-utility}
\end{equation}
where $\mathcal K_N=\{k\in\mathcal K:k\le N\}$ and
$\mathcal K\subseteq\{1,2,3,4\}$.
The learned potentials provide a structured predictive decomposition rather than
an assumed one-to-one correspondence with physical interaction energies.
Architectures, symmetry constraints, and training objectives are detailed in
Appendix~\ref{app:utility-models}.

\subsection{Reward-Guided Planning}
\label{sec:reward-guided-planning}

All planners consume the same learned utility through the penalized score
\begin{equation}
r_S(\mathbf u)
=
\widehat U_\theta(X_S,\mathbf u)
-
\lambda_{\mathrm{road}}\phi_{\mathrm{road}}(\mathbf u)
-
\lambda_{\mathrm{sep}}\phi_{\mathrm{sep}}(\mathbf u),
\label{eq:planning-score}
\end{equation}
where the penalties encourage road feasibility and the minimum TX separation.
A planner $\mathcal A$ produces continuous candidate deployments, which are
converted to an executable TX set by a common selection-and-repair operator,
\begin{equation}
\mathcal C=\mathcal A[r_S],\qquad
\widehat P=\mathcal R_S(\mathcal C)\in\mathcal F_N(S).
\label{eq:planning-interface}
\end{equation}
The selected deployment is then evaluated by $F_{\mathrm{RT}}$;
simulator outputs are used only for final physical verification and never to
rerank candidates during search.

We consider annealed score-based, Monte Carlo, and derivative-free planners,
all operating through this common scalar interface.
Their objectives, constraints, and implementations are given in
Appendix~\ref{app:planning}.

\section{Benchmarks}
\label{sec:benchmarks}

CityDeploy-Bench evaluates physics-grounded spatial planning
under realistic urban geometry and jointly simulated wireless
interactions. The benchmark couples geographically diverse
urban scenes with scene-specific deployment constraints,
set-valued planning objectives, and a common physical verifier.
Figure~\ref{fig:citydeploy_overview} summarizes the benchmark
pipeline.

\subsection{CityDeploy-Data}
\label{sec:citydeploy_data}
\begin{figure}[t]
    \centering
    \includegraphics[width=0.90\linewidth]{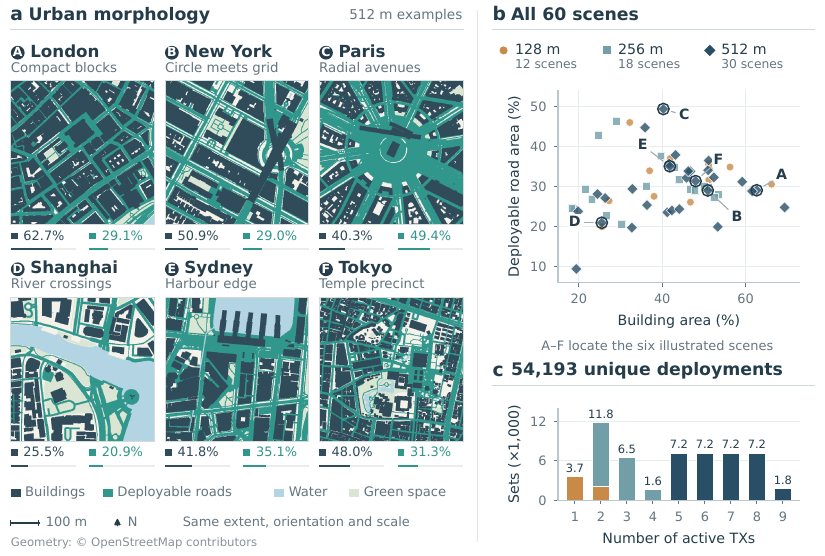}
    \caption{
    \textbf{CityDeploy-Data composition.}
    \textbf{(a)} Representative $512\,\mathrm{m}$ urban morphologies.
    \textbf{(b)} Building and deployable-road fractions across all
    60 scenes and three scene scales.
    \textbf{(c)} Distribution of 54,193 unique deployments by
    active TX count.
    }
    \label{fig:scene_morphology}
\end{figure}
\paragraph{Urban scenes and deployments.}
CityDeploy-Data contains \textbf{54,193 unique deployments}
from 60 urban scenes across six cities at 128, 256, and
512\,m scales. Each scene provides aligned urban geometry,
semantic and radio-material maps, deployable roads, and receiver
grids. Unordered TX sets spanning multiple cardinalities are
evaluated by the common ray-tracing pipeline.
Figure~\ref{fig:scene_morphology} summarizes the diversity in
urban morphology, scene scale, and deployment cardinality.

\paragraph{Supervision and geographic splits.}
Each deployment is paired with dense PL, SS-RSRP, SINR, and
effective-throughput responses and their service labels, all
recomputed for the complete active TX set. Geographically
overlapping regions are grouped before partitioning to prevent
spatial leakage. Construction, sampling, split, and integrity
details are provided in Appendix~\ref{app:data}.

\subsection{Benchmark Tasks}
\label{sec:tasks-evaluation}

\paragraph{Fixed-cardinality planning.}
Given a scene $S$ and prescribed TX cardinality $N$, planners optimize the
fixed-cardinality objective in Eq.~\eqref{eq:deployment-objective} and return one
executable deployment for physical verification.
We report simulator-verified JM coverage as the primary planning-quality metric,
together with the marginal service coverages and planning cost.
This task measures how effectively a planner uses a fixed infrastructure budget.

\paragraph{Target-coverage planning.}
Given a target JM coverage $\gamma_{\mathrm{tar}}$, the ideal objective is
\begin{equation}
N^\star_{\gamma}(S)
=
\min
\left\{
N:
\exists P\in\mathcal F_N(S),\;
U(S,P)\ge\gamma_{\mathrm{tar}}
\right\}.
\label{eq:target-cardinality-objective}
\end{equation}
In practice, each cardinality is searched independently in ascending order, and
the procedure stops at the first simulator-verified success.
Because each inner search is approximate, the returned cardinality is the first
successful tested count rather than a claimed global minimum.
We report success rate, first successful TX count, best verified JM coverage,
and cumulative planning cost.
This task measures reliable service attainment and infrastructure efficiency
under finite search.

\paragraph{Transfer and adaptation.}
We evaluate whether learned utility models remain useful when planning shifts to
new geographic scenes.
\emph{Zero-shot} evaluation applies the base checkpoint directly, while
\emph{few-shot} and \emph{full-pool} adaptation update the utility model using
increasing amounts of target-scene supervision.
Adaptation and evaluation deployments remain disjoint.
Planning performance is measured using the same fixed-cardinality or
target-coverage protocols above.
City-level generalization is claimed only when the complete target city is
excluded from base training.
Adaptation pools, provenance, stopping rules, and computational accounting are
specified in Appendices~\ref{app:model-adaptation} and~\ref{app:evaluation}.

\section{Experiments}
\label{sec:experiments}

We evaluate CityDeploy-Bench along the three planning regimes defined in
Section~\ref{sec:benchmarks}.
Rather than treating the utility model and the planner as a single method,
our experiments separate two complementary design axes:
\emph{how deployment utility is represented} and
\emph{how the resulting reward landscape is searched}.
This allows us to study not only which combinations achieve high simulator-verified
service, but also how physical coupling, relational structure, search dynamics,
and geographic shift determine planning behavior. All reported service metrics are computed on the final
road-feasible deployment using the physical simulator rather than the learned
utility model.
Detailed training configurations, planner hyperparameters, additional scenes,
target levels, and ablations are provided in the appendices.

\subsection{Experimental Setup}
\label{sec:experimental-setup}

\paragraph{Utility representations.}
We evaluate four permutation-invariant utility representations introduced in
Section~\ref{sec:utility-representations}:
the pooled \emph{Regressor}, the pairwise Interaction Network (\emph{IN}),
the latent-relation \emph{NRI} model, and the higher-order
Hypergraph Potential Energy Model (\emph{HPEM}).
These models expose the same scalar utility interface but differ in the relational
structure available to represent interactions among simultaneously active TXs.

\paragraph{Planning algorithms.}
We benchmark nine search algorithms spanning three planning families.
The annealed score-based planner uses reward gradients under a decreasing noise
schedule.
Monte Carlo methods include Gaussian and Langevin weighted sampling
(\emph{WS-G} and \emph{WS-L}), NUTS~\citep{hoffman2014no}, and the
BR-SNIS adaptation~\citep{cardoso2022br}.
Derivative-free methods include Greedy search,
particle swarm optimization (\emph{PSO}), a genetic algorithm (\emph{GA}),
and TuRBO~\citep{eriksson2020scalable}.
Each compatible planner can consume any of the learned utility representations,
making utility representation and search strategy separable experimental factors.

\paragraph{Evaluation protocol.}
For fixed-cardinality planning, we report simulator-verified JM coverage and
the associated marginal service coverages.
For target-coverage planning, we additionally report success rate, the first
successful tested TX count, best verified JM coverage, and cumulative planning cost.
Planning time, model-evaluation counts, and random-seed variability are reported
separately because heterogeneous planner iterations are not treated as equivalent
computational units.
Unless explicitly stated otherwise, results aggregate independent random seeds;
full per-scene results and implementation settings are given in the appendix.

\subsection{Findings}
\label{sec:findings}

We summarize six findings that characterize how physical coupling,
utility representation, and search interact in multi-TX planning.

\paragraph{Finding 1: More coverage does not necessarily mean better service.}
Multi-TX deployment has a non-monotone marginal structure.
To isolate the physical effect of adding one transmitter, we compare nested
deployments $S$ and $S\cup\{p\}$ from the same scene and deployment family,
while keeping all existing TX locations unchanged.

\begin{figure}[t]
    \centering
    \includegraphics[width=\linewidth]{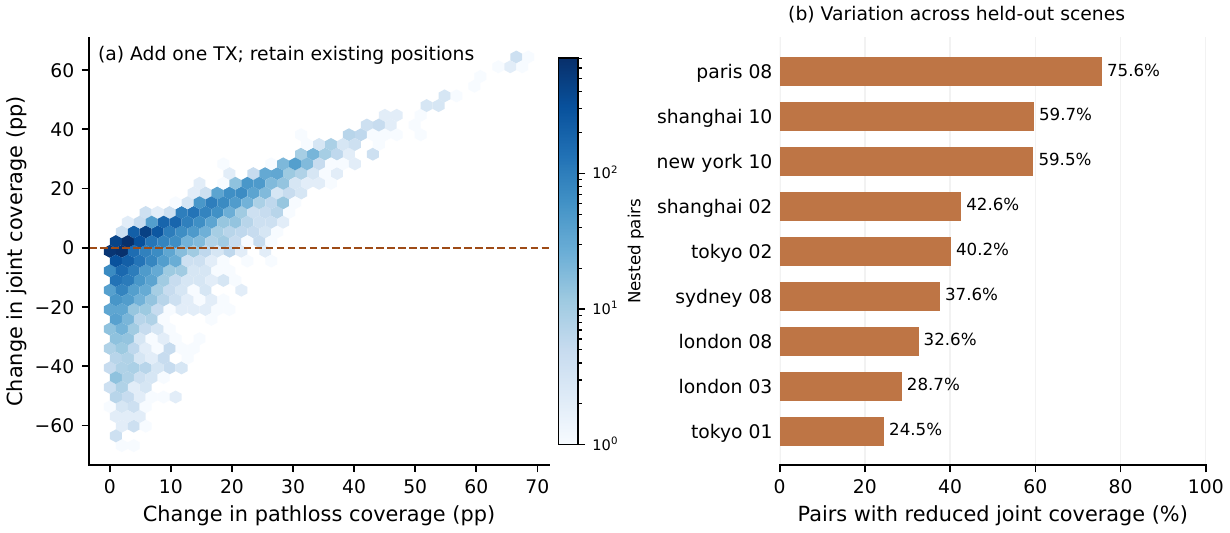}
    \caption{
    \textbf{Marginal effect of adding one transmitter.}
    We compare nested deployments $S$ and $S\cup\{p\}$ while retaining all
    existing TX positions.
    \textbf{Left:} change in path-loss coverage versus change in JM coverage
    across matched deployment pairs; density below the horizontal zero line
    shows cases where serving-link coverage improves while joint service
    deteriorates.
    \textbf{Right:} fraction of nested pairs with reduced JM coverage in each
    held-out scene.
    Across all 9,320 pairs, 42.44\% exhibit a decrease in JM coverage after the
    additional TX is activated.
    }
    \label{fig:nested-set-diagnostic}
\end{figure}

Figure~\ref{fig:nested-set-diagnostic} shows that an added TX can increase
path-loss coverage while simultaneously reducing JM coverage.
Across 9,320 nested deployment pairs from the held-out scenes, 3,955
($42.44\%$) exhibit lower joint coverage after the additional TX is activated.
This statistic is descriptive of the available nested pairs rather than an
estimated probability over arbitrary deployments.

The mechanism follows directly from the shared-spectrum physical model.
An added TX can strengthen the serving signal or cover previously weak regions,
while also introducing co-channel interference and changing serving association.
Its marginal value therefore depends on the existing deployment:
\[
U(S\cup\{p\})-U(S)
\]
can be positive or negative even when serving-link coverage improves.
Multi-TX planning must consequently evaluate transmitters as a coordinated set
rather than rank locations by individual coverage quality.

Aggregate cardinality-level trends exhibit the same coverage--interference
transition across deployment scales
(Figure~\ref{fig:tx_metric_tradeoff} in
Appendix~\ref{app:diagnostics_scaling}).

\paragraph{Finding 2: Relational complexity should match physical coupling.}
No single utility representation is uniformly preferable across deployment
regimes.
At low TX cardinality, pooled or pairwise representations can match or exceed
higher-order HPEM guidance for some planners; as the active set grows, replacing
HPEM produces increasingly systematic degradation
(Figure~\ref{fig:reward_sensitivity} in
Appendix~\ref{app:diagnostics_reward}).
The relevant representation complexity is therefore determined by the interaction
structure induced by geometry, cardinality, and interference rather than by
cardinality alone.

The controlled ablations sharpen this observation.
Increasing generic model capacity does not monotonically reduce utility-prediction
error, whereas increasing the represented interaction order improves verified
planning quality at matched inference budgets
(Appendix~\ref{app:ablation}).
Higher order is therefore useful when the deployment requires it, rather than as
a generic consequence of using a larger model.

\paragraph{Finding 3: Representation and search provide complementary gains.}
Changing the utility representation affects different planners
by substantially different amounts
(Figure~\ref{fig:reward_sensitivity}), showing that search
depends on the geometry of the learned reward landscape.
Conversely, larger inference budgets improve planning within
each tested interaction order without eliminating the advantage
of higher-order representations
(Appendix~\ref{app:ablation_inference}).
Search effort and relational structure therefore address
different bottlenecks rather than substituting for one another.

\paragraph{Finding 4: Planning performance is multi-dimensional.}
Search methods occupy different operating regions in service quality,
computational cost, and seed sensitivity.
The quality--efficiency frontier changes with deployment scale
(Figure~\ref{fig:quality_efficiency_pareto} in
Appendix~\ref{app:diagnostics_efficiency}), so no single search strategy defines
the preferred operating point across all regimes.

\begin{figure}[t]
    \centering
    \includegraphics[width=0.65\linewidth]{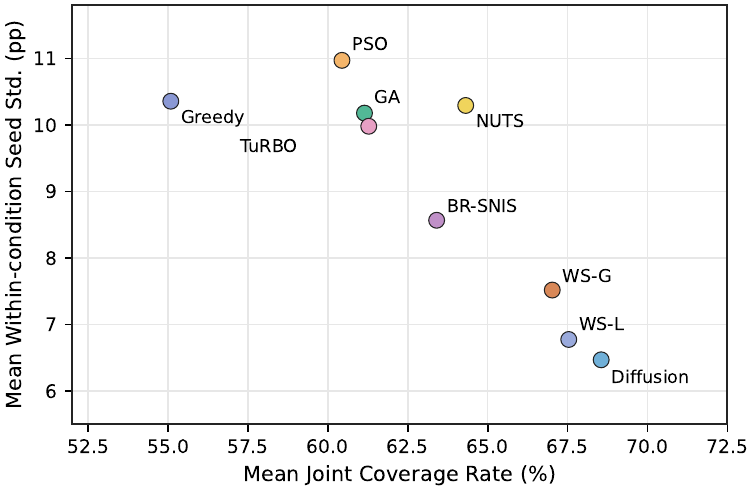}
    \caption{
    \textbf{Planning quality and random-seed stability.}
    Mean simulator-verified JM coverage is plotted against the mean
    within-condition standard deviation over five random seeds.
    Each condition is defined by city, deployment scale, TX cardinality,
    and utility representation.
    Several stochastic planners occupy the high-quality, low-variance region,
    showing that stochastic search does not inherently imply unstable planning.
    }
    \label{fig:quality_seed_stability}
\end{figure}

Stochasticity also does not imply poor repeatability.
As shown in Figure~\ref{fig:quality_seed_stability}, several sampling-based
planners combine high simulator-verified JM coverage with relatively low
within-condition variation across random seeds.
Planning methods should therefore be compared jointly in terms of service quality,
cost, and repeatability rather than by a single best-coverage value.

\paragraph{Finding 5: Peak quality does not imply reliable target attainment.}
Target-coverage planning separates peak utility from reliable
service attainment and infrastructure cost.
At the 70\% JM target, Diffusion succeeds in all 15 trials with
$2.20\pm0.45$ TXs at first success.
TuRBO reaches comparable peak JM coverage but succeeds in only
$66.67\%$ of trials, while WS-L attains a $93.33\%$ success rate
with the lowest cumulative planning time among the high-success
methods.
Thus peak quality, reliability, required infrastructure, and
search cost define distinct planning objectives.
Additional targets and scenes are reported in
Appendix~\ref{append:Simulation results with target JM coverage}.

\begin{table*}[t]
\centering
\caption{
\textbf{Target-coverage planning at 70\% JM coverage in the representative Paris scene using HPEM guidance.}
We report the first successful tested TX count (Req.\ TXs), success rate (SR),
best simulator-verified JM coverage, marginal service coverages, and cumulative
planning time.
Values are mean$\pm$std over 15 independent trials.
Required-TX statistics are computed over successful trials; failures remain in
the success-rate denominator.
}
\label{tab:target_joint_tx_paris07_70}
\setlength{\tabcolsep}{3.5pt}
\renewcommand{\arraystretch}{1.00}
\scriptsize
\begin{adjustbox}{max width=\textwidth,center}
\begin{tabular}{ccccccccc}
\toprule
\textbf{Method}
& \textbf{Req. TXs$\downarrow$}
& \textbf{SR (\%)$\uparrow$}
& \textbf{Best JM (\%)$\uparrow$}
& \textbf{PL (\%)$\uparrow$}
& \textbf{SS-RSRP (\%)$\uparrow$}
& \textbf{SINR (\%)$\uparrow$}
& \textbf{Eff. TP (\%)$\uparrow$}
& \textbf{RT (s)$\downarrow$} \\
\midrule

Diffusion
& \first{2.20$\pm$0.45}
& \first{100.00}
& \first{76.82$\pm$3.42}
& \first{87.35$\pm$5.33}
& \second{84.80$\pm$6.49}
& \first{79.49$\pm$3.15}
& 78.51$\pm$2.49
& 66.09$\pm$49.36 \\

BR-SNIS
& 3.60$\pm$1.14
& \third{86.67}
& 70.92$\pm$3.46
& \second{84.74$\pm$6.52}
& 79.25$\pm$2.95
& 73.72$\pm$3.55
& \first{83.88$\pm$3.89}
& 31.70$\pm$13.98 \\

NUTS
& 3.40$\pm$0.89
& \third{86.67}
& 74.96$\pm$3.96
& 81.92$\pm$6.39
& 82.00$\pm$6.38
& 77.87$\pm$3.73
& 81.60$\pm$4.03
& 36.09$\pm$14.33 \\

TuRBO
& \third{3.00$\pm$1.00}
& 66.67
& \second{76.39$\pm$5.92}
& 83.68$\pm$9.22
& 83.79$\pm$9.20
& \second{78.98$\pm$5.50}
& \second{82.84$\pm$5.69}
& \third{30.21$\pm$14.64} \\

PSO
& 4.40$\pm$2.61
& 80.00
& 72.78$\pm$3.66
& 79.19$\pm$2.95
& \first{87.42$\pm$5.29}
& 75.14$\pm$4.43
& 81.69$\pm$3.35
& 41.48$\pm$32.41 \\

GA
& 4.40$\pm$2.70
& 73.33
& 72.39$\pm$3.20
& 78.81$\pm$7.63
& 78.90$\pm$7.60
& 75.96$\pm$1.59
& 80.01$\pm$3.41
& 42.32$\pm$33.47 \\

Greedy
& \worst{6.20$\pm$3.83}
& \worst{40.00}
& \worst{65.91$\pm$6.89}
& \worst{75.84$\pm$5.06}
& \worst{75.96$\pm$5.02}
& \worst{70.14$\pm$6.95}
& \worst{76.29$\pm$3.68}
& \worst{100.92$\pm$186.00} \\

WS-G
& \second{2.40$\pm$0.55}
& 73.33
& \third{75.80$\pm$4.25}
& \third{83.88$\pm$3.44}
& \third{83.94$\pm$3.42}
& \third{77.91$\pm$3.98}
& \third{82.52$\pm$2.83}
& \second{18.72$\pm$7.03} \\

WS-L
& 3.60$\pm$3.05
& \second{93.33}
& 72.97$\pm$1.95
& 78.94$\pm$4.03
& 79.01$\pm$4.05
& 75.45$\pm$1.06
& 78.47$\pm$1.46
& \first{18.07$\pm$6.93} \\
\bottomrule
\end{tabular}
\end{adjustbox}
\end{table*}

\paragraph{Finding 6: Interaction structure transfers more readily than its calibration.}
Zero-shot utility models retain useful planning behavior on target scenes, but
target-scene adaptation consistently improves simulator-verified deployment quality.
Table~\ref{tab:transfer_paris12} shows this effect in a challenging 12-TX Paris
setting: all three utility representations improve from zero-shot to few-shot and
full-pool adaptation.

\begin{table*}[t]
\centering
\caption{
\textbf{Target-scene adaptation in a challenging Paris deployment with 12 TXs.}
We compare zero-shot transfer, few-shot adaptation, and full-pool adaptation
for utility models with increasing relational structure.
Results are reported as mean$\pm$std.
Adaptation consistently improves simulator-verified service, while higher-order
relational modeling retains an advantage across all three regimes.
}
\label{tab:transfer_paris12}
\setlength{\tabcolsep}{4.0pt}
\small

\resizebox{0.96\textwidth}{!}{
\begin{tabular}{llcccccc}
\toprule
\textbf{Utility}
& \textbf{Setting}
& \textbf{RT (s)$\downarrow$}
& \textbf{JM (\%)$\uparrow$}
& \textbf{PL (\%)$\uparrow$}
& \textbf{TP (\%)$\uparrow$}
& \textbf{SINR (\%)$\uparrow$}
& \textbf{RSRP (\%)$\uparrow$} \\
\midrule

\multirow{3}{*}{Regressor}
& Zero-shot
& 64.82$\pm$1.06
& 34.76$\pm$3.31
& 43.28$\pm$4.48
& 55.62$\pm$4.37
& 41.54$\pm$3.72
& 47.31$\pm$4.16 \\

& Few-shot
& 66.31$\pm$1.42
& 38.21$\pm$2.76
& 46.78$\pm$3.71
& 59.08$\pm$3.56
& 45.02$\pm$3.08
& 50.82$\pm$3.43 \\

& Full-pool
& 66.76$\pm$1.15
& 39.47$\pm$2.28
& 48.06$\pm$3.12
& 60.31$\pm$2.98
& 46.34$\pm$2.58
& 52.11$\pm$2.87 \\

\midrule

\multirow{3}{*}{Pairwise}
& Zero-shot
& 70.18$\pm$1.11
& 37.64$\pm$2.91
& 46.21$\pm$4.02
& 58.46$\pm$3.89
& 44.37$\pm$3.31
& 50.18$\pm$3.72 \\

& Few-shot
& 71.82$\pm$1.53
& 41.03$\pm$2.38
& 49.66$\pm$3.31
& 61.84$\pm$3.17
& 47.86$\pm$2.74
& 53.64$\pm$3.03 \\

& Full-pool
& 72.31$\pm$1.22
& 42.36$\pm$1.94
& 50.94$\pm$2.76
& 63.08$\pm$2.63
& 49.21$\pm$2.28
& 54.91$\pm$2.52 \\

\midrule

\multirow{3}{*}{4th-Order}
& Zero-shot
& 80.73$\pm$1.43
& 39.42$\pm$2.68
& 48.03$\pm$3.74
& 60.31$\pm$3.57
& 46.16$\pm$3.04
& 52.06$\pm$3.43 \\

& Few-shot
& 82.64$\pm$1.81
& 43.18$\pm$2.14
& 51.84$\pm$3.02
& 64.02$\pm$2.87
& 49.94$\pm$2.48
& 55.82$\pm$2.78 \\

& Full-pool
& 83.17$\pm$1.47
& 44.72$\pm$1.73
& 53.26$\pm$2.48
& 65.34$\pm$2.32
& 51.41$\pm$2.03
& 57.18$\pm$2.26 \\

\bottomrule
\end{tabular}
}
\end{table*}

The relational advantage also persists across adaptation regimes.
For example, the 4th-order model improves JM coverage from
$39.42\pm2.68\%$ in the zero-shot setting to $43.18\pm2.14\%$ after few-shot
adaptation and $44.72\pm1.73\%$ with full-pool supervision, while remaining above
the pooled and pairwise alternatives at each stage.
This suggests that part of the interaction structure transfers across scenes,
whereas its quantitative calibration remains sensitive to local geometry and
propagation.

Updating the utility model therefore provides a direct mechanism for adapting
planning without relearning the search procedure.
These experiments measure target-scene or geographic transfer unless the complete
target city is excluded from base training; we do not interpret scene-level
separation as unseen-city generalization.
Additional adaptation results are reported in
Appendix~\ref{app:model-adaptation}.

\section{Conclusion}

CityDeploy-Bench benchmarks physics-grounded spatial set
planning under a common ray-tracing verifier. Separating utility
representation from search reveals a coupling-dependent
transition: densification shifts deployment toward
interference-aware coordination, relational structure becomes
increasingly important, and additional search cannot fully
compensate for missing interactions. Target-coverage and
transfer experiments further distinguish peak quality, reliable
service attainment, and cross-scene calibration. The benchmark
provides a reproducible testbed for studying search,
representation, and transfer in physically grounded decision
making.



\bibliography{iclr2027_conference}

@article{levie2020radio,
  author={Ron Levie and Çağkan Yapar and Gitta Kutyniok and Giuseppe Caire},
  title={RadioUNet: Fast Radio Map Estimation With Convolutional Neural Networks},
  journal={IEEE Transactions on Wireless Communications},
  volume={20},
  number={6},
  pages={4001--4015},
  year={2021},
  doi={10.1109/TWC.2021.3054977}
}

@misc{tian2021radionet,
      title={RadioNet: Transformer based Radio Map Prediction Model For Dense Urban Environments}, 
      author={Yu Tian and Shuai Yuan and Weisheng Chen and Naijin Liu},
      year={2021},
      eprint={2105.07158},
      archivePrefix={arXiv},
      primaryClass={eess.SP},
      url={https://arxiv.org/abs/2105.07158}, 
}

@misc{li2022analytics,
      title={Analytics and Machine Learning Powered Wireless Network Optimization and Planning}, 
      author={Ying Li and Djordje Tujkovic and Po-Han Huang},
      year={2022},
      eprint={2209.06352},
      archivePrefix={arXiv},
      primaryClass={eess.SY},
      url={https://arxiv.org/abs/2209.06352}, 
}

@ARTICLE{Qiu2024Large,
  author={Qiu, Kehai and Bakirtzis, Stefanos and Wassell, Ian and Song, Hui and Zhang, Jie and Wang, Kezhi},
  journal={IEEE Wireless Communications Letters}, 
  title={Large Language Model-Based Wireless Network Design}, 
  year={2024},
  volume={13},
  number={12},
  pages={3340-3344},
  doi={10.1109/LWC.2024.3462556}}

@ARTICLE{Quan2025Large,
  author={Quan, Hongye and Ni, Wanli and Zhang, Tong and Ye, Xiangyu and Xie, Ziyi and Wang, Shuai and Liu, Yuanwei and Song, Hui},
  journal={IEEE Networking Letters}, 
  title={Large Language Model Agents for Radio Map Generation and Wireless Network Planning}, 
  year={2025},
  volume={7},
  number={3},
  pages={166-170},
  doi={10.1109/LNET.2025.3539829}}

@inproceedings{li2026automatic,
  author={Xiaomeng Li and Yuru Zhang and Qiang Liu and Mehmet Can Vuran and
          Nathan Huynh and Li Zhao and Mizan Rahman and Eren Erman Ozguven},
  title={Automatic Network Planning with Digital Radio Twin},
  booktitle={IEEE International Conference on Communications (ICC)},
  pages={1--6},
  year={2026},
  doi={10.1109/ICC59461.2026.11587682}
}

@ARTICLE{Modesto2026LoRaWAN,
  author={Modesto, Cláudio and Mozart, Lucas and Gonçalves, Glauco and Nahum, Cleverson and Castro, Bruno and Klautau, Aldebaro},
  journal={IEEE Open Journal of the Communications Society}, 
  title={LoRaWAN Gateway Placement for Network Planning Using Ray Tracing-Based Channel Models}, 
  year={2026},
  volume={7},
  number={},
  pages={6612-6624},
  doi={10.1109/OJCOMS.2026.3704345}}

@INPROCEEDINGS{Deng2025TelePlanNet,
  author={Deng, Zongyuan and Cai, Yujie and Liu, Qing and Mu, Shiyao and Lyu, Bin and Yang, Zhen},
  booktitle={2025 IEEE/CIC International Conference on Communications in China (ICCC)}, 
  title={TelePlanNet: An AI-Driven Framework for Efficient Telecom Network Planning}, 
  year={2025},
  volume={},
  number={},
  pages={1-6},
  doi={10.1109/ICCC65529.2025.11148630}}

@ARTICLE{Yuan2026LLM,
  author={Yuan, Chenyang and Hou, Jinbo and Yu, Gang and Qiu, Kehai and Wang, Kezhi and Hu, Haonan and Zhang, Jie},
  journal={IEEE Transactions on Mobile Computing}, 
  title={LLM-Assisted Optimisation of Multi-RIS Placement and Beamforming in Smart Warehouses}, 
  year={2026},
  volume={25},
  number={8},
  pages={12722-12734},
  doi={10.1109/TMC.2026.3674609}}

@inproceedings{hoydis2023sionna,
  author={Jakob Hoydis and Fayçal Aït Aoudia and Sebastian Cammerer and
          Merlin Nimier-David and Nikolaus Binder and Guillermo Marcus and Alexander Keller},
  title={Sionna RT: Differentiable Ray Tracing for Radio Propagation Modeling},
  booktitle={2023 IEEE Globecom Workshops (GC Wkshps)},
  pages={317--321},
  year={2023},
  doi={10.1109/GCWkshps58843.2023.10465179}
}

@ARTICLE{Khan2022Digital,
  author={Khan, Latif U. and Han, Zhu and Saad, Walid and Hossain, Ekram and Guizani, Mohsen and Hong, Choong Seon},
  journal={IEEE Communications Surveys \& Tutorials}, 
  title={Digital Twin of Wireless Systems: Overview, Taxonomy, Challenges, and Opportunities}, 
  year={2022},
  volume={24},
  number={4},
  pages={2230-2254},
  doi={10.1109/COMST.2022.3198273}}

@misc{yuan2026planning,
      title={Planning with Language and Generative Models: Toward General Reward-Guided Wireless Network Design}, 
      author={Chenyang Yuan and Xiaoyuan Cheng},
      year={2026},
      eprint={2602.00357},
      archivePrefix={arXiv},
      primaryClass={cs.LG},
      url={https://arxiv.org/abs/2602.00357}, 
}

@inproceedings{eriksson2020scalable,
  author={David Eriksson and Michael Pearce and Jacob R. Gardner and
          Ryan Turner and Matthias Poloczek},
  title={Scalable Global Optimization via Local Bayesian Optimization},
  booktitle={Advances in Neural Information Processing Systems},
  volume={32},
  year={2019}
}

@ARTICLE{Bonyadi2017Evolutionary,
  author={Bonyadi, Mohammad Reza and Michalewicz, Zbigniew},
  journal={Evolutionary Computation}, 
  title={Particle Swarm Optimization for Single Objective Continuous Space Problems: A Review}, 
  year={2017},
  volume={25},
  number={1},
  pages={1-54},
  doi={10.1162/EVCO_r_00180}}

@ARTICLE{Deb2014GA,
  author={Deb, Kalyanmoy and Jain, Himanshu},
  journal={IEEE Transactions on Evolutionary Computation}, 
  title={An Evolutionary Many-Objective Optimization Algorithm Using Reference-Point-Based Nondominated Sorting Approach, Part I: Solving Problems With Box Constraints}, 
  year={2014},
  volume={18},
  number={4},
  pages={577-601},
  doi={10.1109/TEVC.2013.2281535}}

@inproceedings{
piche2018probabilistic,
title={Probabilistic Planning with Sequential Monte Carlo methods},
author={Alexandre Pich\'{e} and Valentin Thomas and Cyril Ibrahim and Yoshua Bengio and Chris Pal},
booktitle={International Conference on Learning Representations},
year={2019},
url={https://openreview.net/forum?id=ByetGn0cYX},
}

@article{naesseth2014sequential,
  title={Sequential Monte Carlo for graphical models},
  author={Naesseth, Christian A and Lindsten, Fredrik and Sch{\"o}n, Thomas B},
  journal={Advances in neural information processing systems},
  volume={27},
  year={2014}
}

@inproceedings{lioutas2023criticsequentialmontecarlo,
  author={Vasileios Lioutas and Jonathan Wilder Lavington and Justice Sefas and
          Matthew Niedoba and Yunpeng Liu and Berend Zwartsenberg and
          Setareh Dabiri and Frank Wood and Adam Scibior},
  title={Critic Sequential Monte Carlo},
  booktitle={International Conference on Learning Representations},
  year={2023},
  url={https://openreview.net/forum?id=ObtGcyKmwna}
}

@article{cappe2007overview,
  title={An overview of existing methods and recent advances in sequential Monte Carlo},
  author={Capp{\'e}, Olivier and Godsill, Simon J and Moulines, Eric},
  journal={Proceedings of the IEEE},
  volume={95},
  number={5},
  pages={899--924},
  year={2007},
  publisher={IEEE}
}

@article{hoffman2014no,
  title={The No-U-Turn Sampler: Adaptively Setting Path Lengths in Hamiltonian Monte Carlo},
  author={Hoffman, Matthew D. and Gelman, Andrew},
  journal={Journal of Machine Learning Research},
  volume={15},
  number={47},
  pages={1593--1623},
  year={2014}
}

@article{cardoso2022br,
  title={Br-snis: bias reduced self-normalized importance sampling},
  author={Cardoso, Gabriel and Samsonov, Sergey and Thin, Achille and Moulines, Eric and Olsson, Jimmy},
  journal={Advances in Neural Information Processing Systems},
  volume={35},
  pages={716--729},
  year={2022}
}

@article{ho2020denoising,
  title={Denoising diffusion probabilistic models},
  author={Ho, Jonathan and Jain, Ajay and Abbeel, Pieter},
  journal={Advances in neural information processing systems},
  volume={33},
  pages={6840--6851},
  year={2020}
}

@inproceedings{janner2022planning,
  author={Michael Janner and Yilun Du and Joshua B. Tenenbaum and Sergey Levine},
  title={Planning with Diffusion for Flexible Behavior Synthesis},
  booktitle={Proceedings of the 39th International Conference on Machine Learning},
  series={Proceedings of Machine Learning Research},
  volume={162},
  pages={9902--9915},
  year={2022}
}

@article{dong2024diffuserlite,
  title={Diffuserlite: Towards real-time diffusion planning},
  author={Dong, Zibin and Hao, Jianye and Yuan, Yifu and Ni, Fei and Wang, Yitian and Li, Pengyi and Zheng, Yan},
  journal={Advances in Neural Information Processing Systems},
  volume={37},
  pages={122556--122583},
  year={2024}
}

@article{ubukata2024diffusion,
  title={Diffusion model for planning: A systematic literature review},
  author={Ubukata, Toshihide and Li, Jialong and Tei, Kenji},
  journal={arXiv preprint arXiv:2408.10266},
  year={2024}
}

@inproceedings{akhound2024iterated,
  author={Tara Akhound-Sadegh and Jarrid Rector-Brooks and Joey Bose and
          Sarthak Mittal and Pablo Lemos and Cheng-Hao Liu and Marcin Sendera and
          Siamak Ravanbakhsh and Gauthier Gidel and Yoshua Bengio and
          Nikolay Malkin and Alexander Tong},
  title={Iterated Denoising Energy Matching for Sampling from Boltzmann Densities},
  booktitle={Proceedings of the 41st International Conference on Machine Learning},
  series={Proceedings of Machine Learning Research},
  volume={235},
  pages={760--786},
  year={2024}
}

@article{pan2024model,
  title={Model-based diffusion for trajectory optimization},
  author={Pan, Chaoyi and Yi, Zeji and Shi, Guanya and Qu, Guannan},
  journal={Advances in Neural Information Processing Systems},
  volume={37},
  pages={57914--57943},
  year={2024}
}

@inproceedings{cheng2025safe,
 author = {Cheng, Xiaoyuan and Tang, Xiaohang and Yang, Yiming},
 booktitle = {Advances in Neural Information Processing Systems},
 doi = {10.52202/085713-4852},
 editor = {D. Belgrave and C. Zhang and H. Lin and R. Pascanu and P. Koniusz and M. Ghassemi and N. Chen},
 pages = {144946--144985},
 publisher = {Curran Associates, Inc.},
 title = {Safe and Stable Control via Lyapunov-Guided Diffusion Models},
 url = {https://proceedings.neurips.cc/paper_files/paper/2025/file/d59b4ae6646183d3848d7aaa0989cbbd-Paper-Conference.pdf},
 volume = {38},
 year = {2025}
}

@inproceedings{battaglia2016interaction,
  author={Peter W. Battaglia and Razvan Pascanu and Matthew Lai and
          Danilo Jimenez Rezende and Koray Kavukcuoglu},
  title={Interaction Networks for Learning about Objects, Relations and Physics},
  booktitle={Advances in Neural Information Processing Systems},
  volume={29},
  year={2016}
}

@inproceedings{kipf2018neural,
  author={Thomas Kipf and Ethan Fetaya and Kuan-Chieh Wang and Max Welling and Richard Zemel},
  title={Neural Relational Inference for Interacting Systems},
  booktitle={Proceedings of the 35th International Conference on Machine Learning},
  series={Proceedings of Machine Learning Research},
  volume={80},
  pages={2688--2697},
  year={2018}
}

@misc{feng2019hyper,
      title={Hypergraph Neural Networks}, 
      author={Yifan Feng and Haoxuan You and Zizhao Zhang and Rongrong Ji and Yue Gao},
      year={2019},
      eprint={1809.09401},
      archivePrefix={arXiv},
      primaryClass={cs.LG},
      url={https://arxiv.org/abs/1809.09401}, 
}

@article{Battiston2021physics,
   title={The physics of higher-order interactions in complex systems},
   volume={17},
   ISSN={1745-2481},
   url={http://dx.doi.org/10.1038/s41567-021-01371-4},
   DOI={10.1038/s41567-021-01371-4},
   number={10},
   journal={Nature Physics},
   publisher={Springer Science and Business Media LLC},
   author={Battiston, Federico and Amico, Enrico and Barrat, Alain and Bianconi, Ginestra and Ferraz de Arruda, Guilherme and Franceschiello, Benedetta and Iacopini, Iacopo and Kéfi, Sonia and Latora, Vito and Moreno, Yamir and Murray, Micah M. and Peixoto, Tiago P. and Vaccarino, Francesco and Petri, Giovanni},
   year={2021},
   month=Oct, pages={1093–1098} }

@inproceedings{wang2023hyper,
  author={Yuxin Wang and Quan Gan and Xipeng Qiu and Xuanjing Huang and David Wipf},
  title={From Hypergraph Energy Functions to Hypergraph Neural Networks},
  booktitle={Proceedings of the 40th International Conference on Machine Learning},
  series={Proceedings of Machine Learning Research},
  volume={202},
  pages={35605--35623},
  year={2023}
}
\bibliographystyle{iclr2027_conference}

\clearpage
\appendix

\section{CityDeploy-Data Construction}
\label{app:data}

CityDeploy-Data pairs urban scenes and feasible TX sets with spatial radio maps and service labels.
This appendix describes scene construction, deployment sampling, and data organization; Appendix~\ref{app:radio} specifies the radio calculations.

\subsection{Geographic Sources and Scene Construction}
\label{app:scene-construction}
\paragraph{Geographic inputs.}
The dataset contains ten scenes from each of London, New York, Paris, Shanghai, Sydney, and Tokyo.
OpenStreetMap provides building footprints, roads, and land-use features within each scene's geographic bounding box.
We project these features to a local metric frame to construct three-dimensional geometry and aligned semantic rasters, retaining the source snapshot, query record, geographic extent, and coordinate reference system.

\paragraph{Geometry and material priors.}
Building footprints are extruded using \texttt{building:height} or \texttt{height} attributes, followed by level counts at $3.2\,\mathrm{m}$ per level and a deterministic building-type prior when heights are missing.
The benchmark construction uses a flat ground surface.
Semantic classes are mapped to Sionna radio-material priors: concrete for buildings and roads, medium-dry ground for ground and green areas, and wet ground for water.
These assignments specify the simulation model rather than measured material composition.
Scene manifests record the height policy, material profile, software versions, and hashes of the source and generated assets.

\subsection{Spatial Coordinates and Deployable Regions}
\label{app:coordinates}
\paragraph{One coordinate convention.}
A scene of width $L_x$ and height $L_y$ uses a centered horizontal frame in meters.
The normalized coordinates supplied to the learning and planning modules are
\begin{equation}
u_i=\left(\frac{p_{i,x}}{L_x}+\frac12,\frac{p_{i,y}}{L_y}+\frac12\right),
\qquad
p_i=\left((u_{i,x}-\tfrac12)L_x,(u_{i,y}-\tfrac12)L_y\right).
\label{eq:coordinate-map}
\end{equation}
Thus $u_i\in[0,1]^2$, while separation constraints are always evaluated in meters.
The three scene sizes are $128\times128$, $256\times256$, and $512\times512\,\mathrm{m}^2$.

\paragraph{Deployment and evaluation domains.}
The road raster is thresholded at $0.5$ on a $1\,\mathrm{m}$ grid.
The centers of its active cells form $\mathcal D_S$; the corresponding receiver cells form $\Omega_S$.
For an $H\times W$ raster, column $c$ and row $r$ have coordinates
\begin{equation}
p_{x,c}=-\frac{L_x}{2}+(c+\tfrac12)\frac{L_x}{W},\qquad
p_{y,r}=-\frac{L_y}{2}+(r+\tfrac12)\frac{L_y}{H}.
\end{equation}
The dataset explicitly uses the road mask for service evaluation.
Other scene masks, including a general non-water evaluation mask, do not define the denominator in Eq.~\eqref{eq:service-utility}.

\subsection{Deployment Sampling and Subset Families}
\label{app:families}
\paragraph{Master deployments.}
We first sample a feasible master set from $\mathcal D_S$.
A fixed schedule allocates $50\%$ of masters to uniform sampling, $30\%$ to maximin sampling, and $20\%$ to clustered sampling.
Uniform sampling traverses a random candidate order; maximin sampling favors candidates farthest from the selected set; clustered sampling traverses candidates by distance to a randomly chosen center.
Each strategy enforces the $20\,\mathrm{m}$ separation constraint during selection.
The maximin search considers at most 4,096 candidates at each addition.
An incomplete maximin selection falls back to uniform sampling; failure to complete a feasible set raises a generation error.

\paragraph{Related subsets.}
Each master defines a \emph{family} of deployments at prescribed cardinalities (Table~\ref{tab:family-design}).
When all combinations are requested, they are enumerated.
Otherwise, subsets are sampled as prefixes of random permutations of the master, retaining unique subsets and cross-cardinality containment relations.
All subsets inherit spatial feasibility from the master.
Family links expose controlled changes in TX membership without assuming additive or monotone utility.

\begin{table}[!ht]
\centering\small
\caption{Deployment family construction. Counts describe the completed dataset. Memberships count appearances in families before deduplication. The final generation uses three blocks of the base master schedule.}
\label{tab:family-design}
\begin{tabular}{@{}lrrrr@{}}
\toprule
Scene side & Scenes & Master size & Masters/scene & Subsets per master\\
\midrule
$128\,\mathrm{m}$ & 12 & 2 & 180 & $N=1:2;\ N=2:1$\\
$256\,\mathrm{m}$ & 18 & 4 & 90 & $N=2:6;\ N=3:4;\ N=4:1$\\
$512\,\mathrm{m}$ & 30 & 9 & 60 & $N=5,6,7,8:4\text{ each};\ N=9:1$\\
\midrule
Total & 60 & & 5,580 families & 54,900 memberships\\
\bottomrule
\end{tabular}
\end{table}

\paragraph{Propagation reuse and label recomputation.}
The simulator computes a separate propagation-gain map for every TX in a master.
For a subset, we select the corresponding gain maps and recompute serving association, interference, service maps, and coverage labels using only the active TXs.
Channel reuse preserves per-TX propagation; subset labels still require a joint service calculation.

\subsection{Dataset Composition and Geographic Splits}
\label{app:data-splits}
\paragraph{Unique deployments.}
The completed dataset contains 54,193 unique scene--deployment records and 54,900 family memberships.
A deployment is identified by its scene, dataset contract, and sorted candidate indices.
When the same deployment occurs in multiple families, one physical record is retained and the membership table preserves its family links.
The counts by $N=1,\ldots,9$ are respectively 3,655; 11,838; 6,480; 1,620; 7,200; 7,200; 7,200; 7,200; and 1,800.

\paragraph{Geographic separation.}
Scenes with overlapping geographic bounding boxes are connected, and the transitive connected components define 54 geographic groups.
Each group belongs to a single split, preventing overlapping scene extents from crossing the training, validation, and test boundaries.
The fixed assignments yield the composition in Table~\ref{tab:data-splits}.
All six cities occur in the dataset splits; this partition measures transfer to held-out geographic areas within the included cities.
An evaluation that holds out an entire city requires a separate partition.

\begin{table}[!ht]
\centering\small
\caption{Dataset partition. Size columns report numbers of scenes; deployment counts refer to unique records after deduplication.}
\label{tab:data-splits}
\begin{tabular}{@{}lrrrrrr@{}}
\toprule
Split & Groups & Scenes & $128\,\mathrm{m}$ & $256\,\mathrm{m}$ & $512\,\mathrm{m}$ & Deployments\\
\midrule
Training & 36 & 42 & 8 & 12 & 22 & 38,162\\
Validation & 9 & 9 & 2 & 3 & 4 & 8,019\\
Test & 9 & 9 & 2 & 3 & 4 & 8,012\\
\midrule
Total & 54 & 60 & 12 & 18 & 30 & 54,193\\
\bottomrule
\end{tabular}
\end{table}

\subsection{Sample Schema and Data Integrity}
\label{app:data-integrity}
\paragraph{Stored objects.}
Scene features are stored once per scene; family memberships are stored separately from unique deployment records.
Each record contains identifiers, candidate indices, meter-valued and normalized coordinates, simulation provenance, cell counts, and coverage labels.
Associated files store the four service maps, serving-TX indices, evaluation and coverage masks, and per-TX gains.

\paragraph{Frozen construction contract.}
The dataset contract fixes the radio profile, coordinate convention, deployment and evaluation policies, and minimum separation.
The scene registry hashes the manifest, geometry, scene inputs, semantic masks, and candidates; the split manifest records geographic assignments.
Deduplication retains the first stored evaluation and its seed, so repeated family memberships do not represent independent Monte Carlo replications.

\section{Wireless Simulation and Service Utility}
\label{app:radio}

This appendix specifies the forward operator in Eq.~\eqref{eq:physical-response}.
Its calculation follows a single chain: per-TX propagation gains, received powers, serving association and interference, service maps, and cell-wise coverage labels.
The same definitions apply to dataset construction and the simulator evaluation of planned deployments.

\subsection{Radio and Propagation Configuration}
\label{app:radio-profile}
We use Sionna RT~\citep{hoydis2023sionna} with a fixed radio profile (Table~\ref{tab:radio-profile}).
Each TX and RX uses a single-element isotropic antenna with vertical polarization and zero azimuth orientation.
TXs are placed at a fixed height of $10\,\mathrm{m}$, and receivers lie on a horizontal plane at $1.5\,\mathrm{m}$.
Ray tracing uses $5\times10^6$ samples per TX, maximum depth 5, and random seed 42.
All active TXs use the same carrier and bandwidth.
The configured propagation mechanisms are line of sight, specular and diffuse reflection, refraction, diffraction, and edge diffraction.
Material scattering coefficients are set to $0.30$ for concrete, $0.10$ for medium-dry ground, and $0.05$ for wet ground.
The completed dataset records Sionna version 1.2.1 and the CUDA polarized ray-tracing backend.

\begin{table}[!ht]
\centering\small
\caption{Fixed radio profile used to construct service labels. Transmit power is per TX; the receiver noise figure is included in the thermal-noise calculation. }
\label{tab:radio-profile}
\begin{tabular}{@{}ll@{}}
\toprule
Quantity & Value\\
\midrule
Carrier frequency; system bandwidth $W$ & $3.65\,\mathrm{GHz}$; $20\,\mathrm{MHz}$\\
TX power $P_{\mathrm{tx,dBm}}$; TX/RX heights & $24\,\mathrm{dBm}$; $10/1.5\,\mathrm{m}$\\
TX/RX antenna & Single element, isotropic, vertical polarization\\
Receiver noise figure $F_{\mathrm{NF}}$ & $7\,\mathrm{dB}$\\
Subcarrier spacing; resource blocks & $30\,\mathrm{kHz}$; 51\\
Subcarriers per block; SSB power offset $\Delta_{\mathrm{SSB}}$ & 12; $0\,\mathrm{dB}$\\
Receiver-cell size; maximum ray depth & $1\times1\,\mathrm{m}^2$; 5\\
Samples per TX & $5\times10^6$\\
Throughput efficiency $\eta$; resource share $\rho$ & 0.65; 0.50\\
Spectral-efficiency cap $c_{\max}$ & $6\,\mathrm{bit/s/Hz}$\\
\bottomrule
\end{tabular}
\end{table}

The declared system bandwidth $W=20\,\mathrm{MHz}$ is used in both the thermal-noise and throughput calculations.
The $30\,\mathrm{kHz}$ subcarrier spacing and $51\times12=612$ active subcarriers are recorded for the reference-signal power conversion only;
we do not replace $W$ by the corresponding occupied bandwidth of $18.36\,\mathrm{MHz}$ in Eqs.~\eqref{eq:linear-sinr} or~\eqref{eq:effective-throughput}.

\subsection{Received Power and Serving Association}
\label{app:received-power}
Let $g_i(x)\geq0$ be the linear propagation gain from TX $i$ to receiver cell $x$.
The received power in watts and serving index are
\begin{equation}
q_i(x)=P_{\mathrm{tx,W}}g_i(x),\qquad
P_{\mathrm{tx,W}}=10^{(P_{\mathrm{tx,dBm}}-30)/10},\qquad
s(x)=\arg\max_i q_i(x).
\label{eq:association}
\end{equation}
Ties use the first maximum in the stored TX order; equal-power ties do not change the scalar service metrics.
For a positive serving gain, path loss and wideband received power are
\begin{equation}
\mathrm{PL}(x)=-10\log_{10}g_{s(x)}(x),\qquad
P_{\mathrm{rx,dBm}}(x)=10\log_{10}q_{s(x)}(x)+30.
\end{equation}
The reference-signal power proxy distributes wideband power over the configured subcarriers:
\begin{equation}
\mathrm{SS\text{-}RSRP}(x)=P_{\mathrm{rx,dBm}}(x)
-10\log_{10}(N_{\mathrm{RB}}N_{\mathrm{SC/RB}})+\Delta_{\mathrm{SSB}},
\label{eq:ss-rsrp}
\end{equation}
where $N_{\mathrm{RB}}N_{\mathrm{SC/RB}}=51\times12=612$.
This calculation defines a reproducible reference-signal power label; it does not simulate an explicit synchronization-signal waveform.

\subsection{Interference and Service Metrics}
\label{app:interference}
\paragraph{SINR.}
Every non-serving active TX contributes co-channel interference.
Received powers are added without coherent combining across TXs.
With $W$ expressed in hertz, receiver noise and linear SINR are
\begin{align}
P_{\mathrm{n,dBm}}&=-174+10\log_{10}W+F_{\mathrm{NF}},
&P_{\mathrm{n,W}}&=10^{(P_{\mathrm{n,dBm}}-30)/10},\\
I(x)&=\sum_{i\neq s(x)}q_i(x),
&\gamma(x)&=\frac{q_{s(x)}(x)}{I(x)+P_{\mathrm{n,W}}}.
\label{eq:linear-sinr}
\end{align}
The SINR service map is $\mathrm{SINR}(x)=10\log_{10}\gamma(x)$ in dB.
Strongest-power association and interference are recomputed whenever the active TX set changes.

\paragraph{Effective throughput.}
We convert SINR to a capped Shannon-based throughput proxy,
\begin{equation}
T_{\mathrm{eff}}(x)=\frac{W\eta\rho}{10^6}
\min\!\left\{\log_2(1+\gamma(x)),c_{\max}\right\}
\quad[\mathrm{Mbps}].
\label{eq:effective-throughput}
\end{equation}
The fixed efficiency $\eta$ and resource share $\rho$ summarize implementation losses and resource allocation.
With the configured values, the maximum is $39\,\mathrm{Mbps}$.
The proxy measures service under a common allocation assumption; it does not model traffic-dependent scheduling or user load.

\subsection{Coverage Labels and Joint Service Utility}
\label{app:coverage-labels}
\paragraph{Cell-level decisions.}
For metric $j$, let $b_j(x)$ indicate that its value is finite and satisfies the corresponding condition in Table~\ref{tab:service-thresholds}.
Marginal coverage and JM coverage are
\begin{equation}
C_j(S,P)=\frac{1}{|\Omega_S|}\sum_{x\in\Omega_S}b_j(x),\qquad
U(S,P)=\frac{1}{|\Omega_S|}\sum_{x\in\Omega_S}\prod_{j=1}^{4}b_j(x).
\label{eq:coverage-labels}
\end{equation}
Thus JM coverage is the spatial fraction of the intersection of the four service masks, rather than a product or average of the four marginal fractions.
All road cells remain in the denominator, including cells without a valid serving link.
Before metric computation, non-finite and non-positive propagation gains are replaced by zero.
When no positive serving power exists, the logarithmic metrics are marked invalid and effective throughput is zero, so the cell fails the joint criterion.

\begin{table}[!ht]
\centering\small
\caption{Service conditions defining $\boldsymbol\tau$. These are per-cell radio requirements; a target JM coverage is a separate deployment-level requirement.}
\label{tab:service-thresholds}
\begin{tabular}{@{}lll@{}}
\toprule
Service map & Condition & Interpretation\\
\midrule
Path loss & $\mathrm{PL}\leq106\,\mathrm{dB}$ & Serving-link attenuation\\
SS-RSRP & $\mathrm{SS\text{-}RSRP}\geq-110\,\mathrm{dBm}$ & Reference-signal power proxy\\
SINR & $\mathrm{SINR}\geq5\,\mathrm{dB}$ & Interference and noise tolerance\\
Effective throughput & $T_{\mathrm{eff}}\geq10\,\mathrm{Mbps}$ & Rate under the fixed resource share\\
\bottomrule
\end{tabular}
\end{table}

\paragraph{Dependence between service conditions.}
The four service maps represent distinct radio quantities, but the corresponding binary threshold tests are not independent under the fixed benchmark profile.

First, because transmit power and the reference-signal conversion are fixed,
\begin{equation}
\mathrm{SS\text{-}RSRP}
=
24-\mathrm{PL}-10\log_{10}(612)
=
-\mathrm{PL}-3.87
\quad[\mathrm{dBm}].
\end{equation}
Hence
\[
\mathrm{PL}\leq106\,\mathrm{dB}
\quad\Longrightarrow\quad
\mathrm{SS\text{-}RSRP}\geq-109.87\,\mathrm{dBm},
\]
which is slightly stronger than the benchmark requirement of
$-110\,\mathrm{dBm}$.

Second, $T_{\mathrm{eff}}$ is a deterministic monotone function of SINR before the spectral-efficiency cap.
At the benchmark SINR threshold,
\[
\mathrm{SINR}=5\,\mathrm{dB}
\quad\Longrightarrow\quad
T_{\mathrm{eff}}\approx13.37\,\mathrm{Mbps},
\]
which exceeds the $10\,\mathrm{Mbps}$ throughput requirement.

Therefore, for valid cells under the present radio profile and thresholds,
\begin{equation}
\mathbf 1_{\mathrm{JM}}(x)
=
\mathbf 1[\mathrm{PL}(x)\le106]
\,
\mathbf 1[\mathrm{SINR}(x)\ge5].
\label{eq:jm-reduced}
\end{equation}
This equivalence is specific to the chosen profile and thresholds; it is not part of the general definition of JM coverage.
We retain all four service maps and marginal labels because they expose distinct physical/service quantities and allow the same benchmark interface to support alternative radio profiles or threshold settings.

\section{Related Work}\label{app:related-work}

\subsection{Optimization-Based Models}

Optimization-based methods directly search the deployment
space using network utility as black-box feedback. Bayesian
optimization improves evaluation efficiency by fitting a
probabilistic surrogate to the objective, while TuRBO
restricts modeling and candidate selection to adaptive local
trust regions for scalable high-dimensional search
~\citep{eriksson2020scalable}. Heuristic methods instead
explore candidate configurations through local or
population-based updates, including particle-swarm and
evolutionary optimization
~\citep{Bonyadi2017Evolutionary,Deb2014GA,Qiu2024Large}.
These approaches provide strong training-free planning
baselines, but their effectiveness increasingly depends on
whether repeated objective evaluations can resolve fragmented
and highly coupled multi-TX configuration spaces.

\subsection{Monte Carlo Sampling}

Monte Carlo methods reinterpret planning as stochastic
inference over a target distribution whose probability
increases with deployment utility. Sequential Monte Carlo
represents this distribution with weighted particles that are
iteratively propagated, reweighted, and resampled
~\citep{piche2018probabilistic,naesseth2014sequential,
lioutas2023criticsequentialmontecarlo}. Importance-sampling
methods approximate the same target from weighted proposal
samples, with BR-SNIS reducing the bias induced by
self-normalization~\citep{cardoso2022br}. Gradient-informed
MCMC methods such as NUTS instead exploit Hamiltonian
dynamics to explore continuous high-probability regions
~\citep{hoffman2014no}. These methods provide complementary
mechanisms for exploring multi-modal deployment distributions,
but their inference remains governed by the utility
distribution defined by the underlying reward.

\subsection{Diffusion Sampling}

Diffusion-based planners generate solutions through iterative
stochastic refinement, providing a flexible mechanism for
modeling multi-modal planning distributions
~\citep{ho2020denoising,janner2022planning,
ubukata2024diffusion,dong2024diffuserlite}. Reward- and
energy-guided formulations further steer this reverse process
toward task objectives or physical constraints
~\citep{akhound2024iterated,pan2024model,cheng2025safe}.
For wireless deployment, recent work formulates AP placement
as sampling from a reward-induced Gibbs distribution and
shows that deployment quality can be strongly limited by
reward estimation accuracy~\citep{yuan2026planning}. This
makes wireless planning a particularly relevant setting for
examining whether stronger generative inference can compensate
for imperfect representations of physical deployment utility.

\subsection{Relational Graph and Structured Reward Modeling}

Relational models represent multi-entity systems through
objects and their interactions rather than independent entity
features. Interaction Networks explicitly decompose
object-centric and relation-centric computation, while Neural
Relational Inference learns latent interaction graphs directly
from observations
~\citep{battaglia2016interaction,kipf2018neural}. These
formulations provide a natural basis for multi-TX planning,
where the utility of one transmitter depends on the remaining
configuration through coverage complementarity, redundancy,
and interference.

Standard graph formulations primarily encode pairwise
relations. However, many physical systems exhibit collective
effects that arise only from groups of interacting entities
~\citep{Battiston2021physics}. Hypergraph neural networks
represent such dependencies through hyperedges connecting
multiple entities~\citep{feng2019hyper}, while hypergraph
energy formulations provide a principled way to express these
higher-order relations as structured potentials
~\citep{wang2023hyper}. For wireless deployment, this view
extends the reward from a scalar assessment of a complete TX
set to an interaction-aware utility representation over
individual, pairwise, and higher-order TX relations. The open
question is therefore not simply whether relational modeling
improves planning, but which interaction structure provides
the appropriate reward representation under different urban
environments and deployment scales.

\section{Utility Models and Training Details}
\label{app:utility-models}

This appendix specifies the predictors in Section~\ref{sec:utility-representations}, using the saved base-model architectures and training configurations.

\subsection{Scene and TX-Set Inputs}
\label{app:model-inputs}
\paragraph{Scene rasters.}
The input contains 11 channels: buildings, roads, green areas, water, obstacles, built-up areas, explicitly open areas, unknown features, background, TX feasibility, and the stored evaluation mask.
Rasters retain their native $128^2$, $256^2$, or $512^2$ resolution; batches share one scene and therefore one spatial shape.
The stored evaluation-mask feature can represent non-water areas; labels use the road-only domain $\Omega_S$ (Appendix~\ref{app:coverage-labels}).

\paragraph{Spatial encoding.}
A convolutional encoder produces a dense feature field $D_S$ and a global scene embedding $g_S$.
At each normalized TX location $u_i$, bilinear interpolation extracts a local feature $\ell_i=D_S(u_i)$ using border padding and \texttt{align\_corners=False}.
Physical dimensions $(L_x,L_y)$ accompany the coordinates, and an active-TX mask excludes padding from mean/max reductions.
All base configurations use the cardinality feature $c=N/9$.
For Reward, IN, and NRI, the map-size feature is $s=\tfrac12\log_2((L_x,L_y)/128)$, applied elementwise.
HPEM instead uses
\begin{equation}
s_H=\left[\log\frac{(L_x,L_y)}{512},\;
\frac{(L_x,L_y)}{\sqrt{L_x^2+L_y^2}}\right].
\label{eq:model-map-features}
\end{equation}

\begin{table}[!ht]
\centering\small
\caption{Saved base-model architectures. Parameter counts include auxiliary heads. IN and NRI share the scene encoder structure but have different relation modules.}
\label{tab:model-architecture}
\begin{tabularx}{\linewidth}{@{}l*{4}{>{\centering\arraybackslash}X}@{}}
\toprule
Component & Reward & IN & NRI & HPEM\\
\midrule
Convolution channels & 32/64/96 & 32/64/48 & 32/64/48 & 32/64/48\\
Kernel sizes & 5/3/3 & 5/3/3 & 5/3/3 & 5/3/3\\
Strides & 2/2/2 & 2/2/1 & 2/2/1 & 2/2/1\\
GroupNorm groups & 4/8/8 & 8/8/8 & 8/8/8 & 8/8/8\\
Local / global scene width & 96/96 & 48/96 & 48/96 & 48/128\\
TX embedding width & 128 & 128 & 128 & 192\\
Relation / hyperedge width & -- & 128 & 128 & 160\\
Output-head hidden widths & 256/128/64 & 192/192/192 & 192/192/192 & 192/192\\
Output nonlinearity & Sigmoid & Sigmoid & Sigmoid & None\\
Output-head dropout & 0.2 & 0.2 & 0.2 & 0\\
Auxiliary scalar heads & 0 & 0 & 0 & 5\\
Trainable parameters & 244,641 & 521,649 & 833,909 & 1,270,778\\
\bottomrule
\end{tabularx}
\end{table}

\subsection{Set-Pooled and Relational Predictors}
\label{app:set-relational}
\paragraph{Reward Predictor.}
A two-layer TX encoder maps $[u_i,\ell_i]$ to $h_i\in\mathbb R^{128}$.
The output head receives the masked mean and maximum of $\{h_i\}$ together with $g_S$, $s$, and $c$.
Its sigmoid output lies in $[0,1]$.
Convolutions use GroupNorm and SiLU; the head uses SiLU and dropout.

\paragraph{Interaction Network.}
The IN node encoder receives $[u_i,\ell_i,g_S,s,c]$.
For receiver $i$ and sender $j\neq i$, it constructs
\begin{equation}
a_{ij}=[h_i,h_j,\delta_{ij},|\delta_{ij}|,\|\delta_{ij}\|_2],
\qquad \delta_{ij}=u_j-u_i.
\label{eq:pair-descriptor}
\end{equation}
A shared relation MLP gives $v_{ij}=f_{\mathrm{rel}}(a_{ij})$; node updates use mean incoming messages:
\begin{equation}
h'_i=f_{\mathrm{node}}\!\left(h_i,\frac{1}{N-1}\sum_{j\neq i}v_{ij},g_S,s,c\right),\qquad N>1.
\end{equation}
The readout combines mean/max-pooled updated nodes, the mean of all valid directed relation features, and the scene context.
There is one message-passing block and no self-loop; for $N=1$, the relation aggregate is zero.

\paragraph{NRI-style predictor.}
NRI encodes $a_{ij}$ as four logits and averages opposite directions before computing type probabilities.
Type zero contributes no message; other types use separate MLPs:
\begin{equation}
v_{ij}=\sum_{k=1}^{3}q_{ij,k}\,f_k(a_{ij}).
\label{eq:nri-messages}
\end{equation}
Training uses soft Gumbel--Softmax samples at temperature $0.5$.
Evaluation uses deterministic softmax probabilities without Gumbel noise, yielding stable predictions and coordinate gradients.
Although the logits are symmetrized, directed message features can differ between $i\!\rightarrow\!j$ and $j\!\rightarrow\!i$.
The IN aggregation pattern provides the readout; edge types have no observed physical labels.

\subsection{Hypergraph Potential Energy Model}
\label{app:hpem-architecture}
\paragraph{TX embeddings.}
HPEM augments positions with Fourier features
\begin{equation}
\psi(u_i)=\left[u_i,\{\sin(2\pi f u_i),\cos(2\pi f u_i)\}_{f\in\{1,2,4\}}\right].
\end{equation}
A shared MLP encodes $[\psi(u_i),\ell_i,g_S,s_H,c]$ into $h_i\in\mathbb R^{192}$.
MLPs in this model have two SiLU hidden layers followed by a linear output layer.
The singleton energy is obtained from a scalar head on each node, followed by a masked mean.

\paragraph{Symmetric subset features.}
For each enabled order $k\geq2$, all active $k$-TX combinations are enumerated.
Order-specific encoders and mean/max pooling produce symmetric member features.
This is concatenated with an eight-dimensional geometric descriptor: normalized centroid (2), coordinate standard deviations (2), and mean, standard deviation, minimum, and maximum of within-subset pair distances (4).
Distances are computed in meters and divided by the map diagonal; standard deviations use the population convention.
The centroid's scene feature, $g_S$, $s_H$, and $c$ complete the input to an order-specific feature MLP and energy head.

\paragraph{Order aggregation and global readout.}
Let $\mathcal H_k(P)$ contain all active subsets of size $k$.
The model averages their energies,
\begin{equation}
\overline E_k=
\begin{cases}
|\mathcal H_k(P)|^{-1}\sum_{e\in\mathcal H_k(P)}E_{\theta,k}(X_S,e),&\mathcal H_k(P)\neq\varnothing,\\
0,&\mathcal H_k(P)=\varnothing,
\end{cases}
\label{eq:order-aggregation}
\end{equation}
and uses zero for disabled orders.
With $\overline h$ denoting the masked mean of node embeddings, the global and final energies are
\begin{equation}
E_{\mathrm{global}}=f_g([\overline h,g_S,s_H,c,\overline E_1,\ldots,\overline E_4]),\qquad
E_\theta=E_{\mathrm{global}}+\sum_{k\in\mathcal K}w_k\overline E_k.
\label{eq:hpem-global}
\end{equation}
The base model enables orders 1--4 with $w_k=1$.
Its prediction $\widehat U_\theta=-E_\theta$ has no sigmoid or inference-time clipping.
Order means control subset-count scaling, and the global branch couples their contributions.
Five auxiliary heads read the same global input through a separate shared trunk and predict JM, path-loss, SS-RSRP, SINR, and effective-throughput coverage.

\paragraph{Permutation invariance.}
Shared encoders, exhaustive active pairs/subsets, and symmetric reductions preserve predictions when coordinates and masks are permuted together, up to floating-point error.
This holds in deterministic evaluation mode; independent NRI training-time Gumbel draws need not agree across forward passes.
Coordinate gradients exist almost everywhere.

\subsection{Training Objectives and Sampling}
\label{app:model-training}
\paragraph{Supervision.}
Let $y_b=U(S_b,P_b)$ be the JM label for deployment $b$.
Reward and IN minimize $\mathcal L_{\mathrm{MSE}}=\operatorname{Mean}_b(\widehat U_b-y_b)^2$.
NRI adds a regularizer toward a uniform four-type distribution:
\begin{equation}
\mathcal L_{\mathrm{NRI}}=\mathcal L_{\mathrm{MSE}}+10^{-4}\mathcal L_{\mathrm{edge}},\qquad
\mathcal L_{\mathrm{edge}}=
\operatorname{Mean}_{b}\operatorname{Mean}_{i\neq j\text{ active}}
\sum_{k=0}^{3}q_{bij,k}\log\frac{q_{bij,k}}{1/4}.
\end{equation}
The regularizer uses the sampled soft probabilities during training and is zero for a deployment without valid TX pairs.

\paragraph{HPEM objectives.}
The energy model combines utility regression, ranking, and auxiliary supervision:
\begin{align}
\mathcal L_{\mathrm{HPEM}}&=\operatorname{MSE}(-E_\theta,y)
+0.25\mathcal L_{\mathrm{rank}}+0.25\mathcal L_{\mathrm{aux}},\\
\mathcal L_{\mathrm{rank}}&=\frac1{|\mathcal Q|}\sum_{(a,b)\in\mathcal Q}
\left[0.02+\operatorname{sign}(y_a-y_b)(E_a-E_b)\right]_+.
\label{eq:hpem-ranking}
\end{align}
Here $\mathcal Q$ contains within-batch pairs with $a<b$ and $|y_a-y_b|>10^{-4}$.
The ranking loss is zero if $\mathcal Q$ is empty.
The auxiliary loss averages the five coverage-head MSEs.
Planning uses the main prediction $-E_\theta$.

\paragraph{Batch and sample units.}
Batches contain one scene and one TX cardinality, so ranking compares deployments at a common scene and size.
Pairs need not belong to the same master family, and the ranking loss does not compare different cardinalities.
TX order is randomly permuted during training.
The HPEM loader expands family memberships, while Reward, IN, and NRI use unique physical deployments.
Repeated membership therefore changes some HPEM sample weights.
The geographic split is shared, while sampling weights, capacity, and auxiliary supervision differ.

\subsection{Base-Model Configuration}
\label{app:base-model-config}
All base models use AdamW with $(\beta_1,\beta_2)=(0.9,0.999)$ and $\epsilon=10^{-8}$.
The saved runs use seed 42, batch size at most 32, gradient-norm clipping at 5, and a maximum of 100 epochs.
HPEM uses cosine learning-rate decay over the configured training duration and selects the checkpoint with the lowest total validation objective.
The remaining predictors use validation-MSE plateau scheduling and early stopping.
Table~\ref{tab:base-training} gives the remaining settings.

\begin{table}[!ht]
\centering\small
\caption{Training settings from the saved base runs. HPEM uses a different selection loss and sample weighting from the other predictors.}
\label{tab:base-training}
\begin{tabularx}{\linewidth}{@{}lXX@{}}
\toprule
Setting & HPEM & Reward / IN / NRI\\
\midrule
Learning rate & $3\times10^{-4}$ & $10^{-4}$\\
Weight decay & $10^{-4}$ & $10^{-5}$\\
Learning-rate schedule & Cosine, period 100, minimum 0 & Plateau, factor 0.5, patience 5, minimum $10^{-6}$\\
Early-stopping patience & Disabled & 15 epochs\\
Checkpoint criterion & Strictly lower total validation loss & Validation MSE improvement $>10^{-5}$\\
Sampling unit & Family membership & Unique deployment\\
Auxiliary objectives & Ranking and five coverage heads & None / none / edge KL\\
\bottomrule
\end{tabularx}
\end{table}

\subsection{Target-Scene Adaptation}
\label{app:model-adaptation}
Adaptation updates all parameters with the base losses.
From a pool of 990 unique deployments, separate from 200 evaluation examples, 297 examples are used for 30 epochs, followed by all 990 for 70 additional epochs.
AdamW is initialized once and retained between stages: learning rates are $3\times10^{-5}$ for HPEM and $10^{-5}$ otherwise, with weight decay $10^{-5}$, batch size 32, clipping at 5, and seed 42.
There is no scheduler or early stopping; end-of-stage checkpoints are saved without evaluation-based selection.
The retained figures call full-pool fine-tuning \emph{In-distribution}; target-scene and target-city provenance remain separate properties.

\section{Planning Algorithms and Implementation}
\label{app:planning}

All planners share a learned utility interface and a final mapping to road deployments.
They differ in how they propose and update continuous TX coordinates.
This appendix specifies the search and selection procedures.

\subsection{Shared Planning Interface and Constraints}
\label{app:planning-interface}
\paragraph{Utility and search score.}
For fixed $S$ and $N$, write $\mathbf u=(u_1,\ldots,u_N)\in[0,1]^{N\times2}$.
The model receives the scene features $X_S$, normalized coordinates, and a mask identifying active TXs, and returns $\widehat U_\theta(X_S,\mathbf u)$.
The scalar prediction approximates the simulator utility $U(S,P)$.
During continuous search, all planners evaluate
\begin{equation}
r_S(\mathbf u)=\widehat U_\theta(X_S,\mathbf u)
-\lambda_{\mathrm{road}}\phi_{\mathrm{road}}(\mathbf u)
-\lambda_{\mathrm{sep}}\phi_{\mathrm{sep}}(\mathbf u).
\label{eq:search-score}
\end{equation}
Let $m_S(u)$ be the bilinear interpolation of the binary road feasibility mask and $[a]_+=\max(a,0)$.
The penalties are
\begin{align}
\phi_{\mathrm{road}}(\mathbf u)&=\frac1N\sum_i[1-m_S(u_i)]_+,\\
\phi_{\mathrm{sep}}(\mathbf u)&=\binom{N}{2}^{-1}\sum_{i<j}
\left[1-\frac{\|p_i-p_j\|_2}{d_{\min}}\right]_+^2,\quad N\geq2,
\label{eq:search-penalties}
\end{align}
with $\phi_{\mathrm{sep}}=0$ for $N=1$ and $p_i$ recovered using Eq.~\eqref{eq:coordinate-map}.
Mask interpolation uses zero padding and cell-centered coordinates (\texttt{align\_corners=False}).
The score $r_S$ is a search objective, not a physical coverage fraction.
Soft penalties encourage feasibility during search; the discrete checks in Appendix~\ref{app:selection} enforce it on the returned deployment.

\paragraph{Bounded and unconstrained coordinates.}
Reflection acts elementwise as $R(a)=1-|1-(a\bmod2)|$.
The diffusion planner uses it to keep updates in the unit cube.
For Monte Carlo methods using unconstrained coordinates, let $\mathbf u=\operatorname{sigmoid}(z)$.
Their unnormalized log target is
\begin{equation}
\ell_\beta(z)=\beta r_S(\operatorname{sigmoid}(z))
+\sum_{k=1}^{2N}\log\bigl(u_k(1-u_k)\bigr),
\label{eq:logistic-target}
\end{equation}
where $\beta>0$ controls concentration around high scores.
The Jacobian term preserves the intended density $\pi_\beta(\mathbf u)\propto\exp(\beta r_S(\mathbf u))$ with respect to uniform volume on the cube.
Numerical logit conversion clips coordinates to $[10^{-5},1-10^{-5}]$.

\subsection{Diffusion-Based Planning}
\label{app:diffusion}
\paragraph{Reward-induced score.}
The diffusion planner constructs a score from the learned utility and its coordinate gradients; it does not require a separately trained denoiser.
For Gaussian perturbations $\epsilon\sim\mathcal N(0,I)$, define the smoothed positive function
\begin{equation}
h_\sigma(\mathbf u)=\mathbb E_\epsilon
\left[\exp\!\left(\beta r_S(R(\mathbf u+\sigma\epsilon))\right)\right].
\end{equation}
At each step, $M$ perturbations $v_m=\mathbf u+\sigma\epsilon_m$ estimate its log gradient:
\begin{equation}
w_m=\frac{\exp(\beta r_S(R(v_m)))}{\sum_{a=1}^{M}\exp(\beta r_S(R(v_a)))},\qquad
\widehat s_\sigma(\mathbf u)=\beta\sum_{m=1}^{M}w_m\nabla_{v_m}r_S(R(v_m)).
\label{eq:diffusion-score}
\end{equation}
Gradients pass through reflection, and antithetic perturbations pair $\epsilon$ with $-\epsilon$.
Equation~\eqref{eq:diffusion-score} is a self-normalized Monte Carlo approximation, with finite-sample error.

\paragraph{Annealed updates.}
Noise scales decrease geometrically from $\sigma_{\max}$ to $\sigma_{\min}$ over $T$ steps.
Starting from $R(\tfrac12+\sigma_{\max}\epsilon)$, the planner applies a variance-exploding reverse-diffusion-style update,
\begin{equation}
\mathbf u_{t+1}=R\!\left(\mathbf u_t+\Delta_t\widehat s_{\sigma_t}(\mathbf u_t)
+\sqrt{\Delta_t}\,\xi_t\right),\qquad
\Delta_t=\sigma_t^2-\sigma_{t+1}^2,
\label{eq:diffusion-update}
\end{equation}
where $\xi_t\sim\mathcal N(0,I)$ and the last stochastic term is disabled by default.
The implementation retains the best visited state of each trajectory.
This finite-step search does not establish exact sampling from $\pi_\beta$.

\subsection{Monte Carlo Planning}
\label{app:monte-carlo}
\paragraph{Annealed SMC.}
Sequential Monte Carlo evolves $L$ particles through temperatures $\beta_t=\beta(t/T)^a$, starting from uniform samples on the unit cube~\citep{cappe2007overview}.
Before mutation, particle log weights are incremented by
\begin{equation}
\log\widetilde w_t^{(l)}=\log w_{t-1}^{(l)}
+(\beta_t-\beta_{t-1})r_S(\mathbf u_{t-1}^{(l)}).
\end{equation}
After normalization, systematic resampling is triggered when
$\mathrm{ESS}=1/\sum_l(w_t^{(l)})^2$ falls below $\kappa L$; resampled weights are reset to uniform.
Mutation operates in logit coordinates with target $\ell_{\beta_t}$.
\emph{SMC-Gaussian} uses a Gaussian random walk and a Metropolis--Hastings acceptance test.
\emph{SMC-Langevin} proposes
$z'=z+\tfrac12 s_t^2\nabla\ell_{\beta_t}(z)+s_t\epsilon$ and includes the forward and reverse proposal densities in the acceptance ratio.
The proposal scale $s_t$ decreases geometrically.
A score-ranked archive retains high-scoring particle states across temperatures.

\paragraph{NUTS.}
The No-U-Turn sampler~\citep{hoffman2014no} also uses $\ell_\beta$ in logit coordinates.
It samples Gaussian momenta with an identity mass matrix and recursively extends a leapfrog trajectory until a no-U-turn condition, a divergence test, or the maximum tree depth stops the expansion.
Each chain adapts its step size by dual averaging during warmup.
The planner archives only post-warmup states, and final deployment selection uses their search scores.

\paragraph{BR-SNIS adaptation.}
The implementation uses the iterated sampling--importance resampling mechanism underlying BR-SNIS~\citep{cardoso2022br}.
Each chain combines its retained deployment with $M$ new uniform proposals on the cube.
It selects the next retained state from these $M+1$ candidates with probabilities proportional to $\exp(\beta r_S)$.
All candidates are eligible for the score-ranked archive.
This adapts a sampling mechanism to deployment search: the returned plan is a concrete candidate, and no weighted average of TX coordinates is used.

\subsection{Derivative-Free Optimization}
\label{app:derivative-free}
\paragraph{Greedy construction.}
A scrambled Sobol sequence provides a finite pool of normalized TX locations.
Beam search grows deployments from one TX to $N$ TXs, scoring all single-location extensions and retaining the highest-scoring beam at each cardinality.
Coordinate-replacement rounds then refine the complete sets using the same pool.
The method therefore queries the utility model on partial as well as complete deployments.
It is a search heuristic for the learned score; no submodularity assumption is imposed.

\paragraph{Particle swarm optimization.}
PSO maintains a population of deployments, their personal best states, and a global or ring-neighborhood best state.
Velocity updates combine inertia with random cognitive and social attraction terms.
The inertia weight decreases linearly, velocities are clipped, and positions are clamped to the unit cube.
An out-of-bounds velocity component is reversed and halved.
The archive retains the best visited deployments.

\paragraph{Genetic algorithm.}
The GA applies tournament selection, simulated binary crossover, and polynomial mutation to flattened $2N$-dimensional coordinates.
Elitism carries high-scoring deployments into the next generation, and offspring remain within $[0,1]^{2N}$.
When no positive mutation probability is specified, the per-coordinate probability is $1/(2N)$.

\paragraph{TuRBO.}
The TuRBO-1 adaptation~\citep{eriksson2020scalable} fits a Gaussian process with an ARD Mat\'ern-$5/2$ kernel to standardized search scores.
A trust region is centered on the current best point, with side lengths scaled by the learned length scales.
Thompson samples over Sobol candidates select the next evaluation batch.
Consecutive improvements expand the trust region; consecutive failures shrink it, and a region below the minimum size triggers a new Sobol initialization.
All restarts share one explicit score-evaluation budget, and the final archive pools evaluated candidates across restarts.

\subsection{Candidate Selection and Feasibility Repair}
\label{app:selection}
\paragraph{Selection order.}
Each planner returns a collection of candidate deployments and their continuous search scores.
The shared procedure sorts these scores in descending order and maps every TX to its nearest road candidate in meters.
A mapped set is accepted only if its candidate locations are distinct and all pairwise distances satisfy $d_{\min}$, up to a $10^{-6}\,\mathrm{m}$ numerical tolerance.
The first valid mapped set is returned.

\paragraph{Repair fallback.}
When a mapped set fails, the procedure attempts a constrained assignment of its TXs to road candidates.
It tests cyclic rotations of both the original and reversed TX orders.
For each order, each TX greedily takes its nearest unused candidate that is sufficiently separated from already assigned TXs.
Among successful orders, the assignment with the smallest total squared displacement is retained.
The first successful repair in score order is saved as a fallback while the scan continues for a directly valid nearest-candidate mapping.
The fallback is used only when the entire scan yields no such mapping; if neither is available, the procedure reports failure.
This finite assignment heuristic need not find the globally nearest feasible deployment.

\paragraph{Final evaluation.}
The selected coordinates define the executable deployment $P\in\mathcal F_N(S)$.
Its unpenalized model prediction is recomputed, and its service maps and $U(S,P)$ are then obtained from the simulator.
The selection order is based on scores before mapping; candidates are not reranked by post-mapping predictions or simulator outcomes.

\subsection{Algorithm Configurations}
\label{app:planner-config}
Table~\ref{tab:planner-defaults} records implementation defaults; run configurations specify overrides, population size, and seeds.
These reference settings do not imply equal compute budgets across methods.
The single-run interface defaults to $\lambda_{\mathrm{road}}=\lambda_{\mathrm{sep}}=10$.
The TX cardinality $N$, particle count $L$, and number of score evaluations are separate quantities.

\begin{table}[!ht]
\centering\small
\caption{Reference implementation defaults. $d=2N$ is the coordinate dimension. SMC-Gaussian and SMC-Langevin share the annealing settings; only their mutation kernels differ.}
\label{tab:planner-defaults}
\begin{tabularx}{\linewidth}{@{}lX@{}}
\toprule
Planner & Default configuration\\
\midrule
Diffusion & 100 steps; $M=10$; $\beta=1$; $\sigma:0.35\rightarrow0.005$; antithetic noise; reflection; deterministic final step\\
SMC & 100 temperatures; $\beta=1$; $a=1$; $\kappa=0.5$; one mutation per temperature; $s:0.45\rightarrow0.08$\\
NUTS & 50 warmup and 50 retained iterations per chain; $\beta=1$; initial step size 0.15; target acceptance 0.8; maximum depth 7; divergence threshold 1,000\\
BR-SNIS & 100 rounds; 32 new uniform proposals per chain and round; $\beta=1$\\
Greedy & 256 candidate locations; one coordinate-replacement round; scoring batches of at most 4,096 deployments\\
PSO & 100 steps; inertia $0.9\rightarrow0.4$; cognitive/social coefficients 1.49618; velocity limit 0.2; global topology\\
GA & 100 generations; tournament size 3; elite fraction 0.1; crossover probability 0.9; mutation probability $1/d$; crossover/mutation indices 15/20\\
TuRBO & 200 score evaluations; $\max(2d,8)$ initial points; batch size 4; 256 candidates; 25 GP fitting steps; trust-region length 0.8, bounds $[2^{-7},1.6]$; 3 successes to expand; $\max(4,\lceil d/4\rceil)$ failures to shrink\\
\bottomrule
\end{tabularx}
\end{table}

\paragraph{Recorded outputs.}
The evaluation interface stores selected coordinates, post-mapping predictions, simulated service maps and utility, configurations, and search traces.
Score evaluations, gradients, and simulator calls are distinct operations: diffusion uses multiple perturbed model evaluations per update, while NUTS trajectory lengths vary.
Iteration counts alone therefore do not specify equal computation.

\section{Evaluation Protocols and Metrics}
\label{app:evaluation}

This appendix consolidates the task metrics used in Section~\ref{sec:tasks-evaluation}.
Physical service labels are defined once in Appendix~\ref{app:radio}; learned search scores and final selection are defined in Appendix~\ref{app:planning}.

\subsection{Fixed-Cardinality Evaluation}
For every scene, cardinality, method, and seed, the planner returns one deployment after road mapping and separation checks.
The simulator produces $U(S,P)$ and the four marginal coverage fractions $C_j(S,P)$.
Reported percentages are $100U$ and $100C_j$.
Higher coverage is preferred; lower planning time is preferred.
Comparisons use the same scene, active TX count, deployment constraints, radio profile, and service thresholds.

\subsection{Target Coverage and Failure Handling}
For a trial $r$, let $\mathcal N_r$ be the consecutively tested TX counts and $U_{r,N}$ the verified utility of its selected deployment at count $N$.
The success indicator and first successful count are
\begin{equation}
s_r=\mathbf1\!\left[\exists N\in\mathcal N_r:U_{r,N}\geq\gamma_{\mathrm{tar}}\right],\qquad
\widehat N_r=\min\{N\in\mathcal N_r:U_{r,N}\geq\gamma_{\mathrm{tar}}\}.
\end{equation}
The value $\widehat N_r$ is defined only for successful trials.
Success rate is $100R^{-1}\sum_{r=1}^{R}s_r$, and Best JM is $100\max_{N\in\mathcal N_r}U_{r,N}$ for a trial with verified deployments.
The search stops at its first success or exhausts the configured count range.
A missing smaller-count result does not establish minimality across that gap.
When reporting required TX counts over successful seeds, their conditional nature is explicit and success rate is reported alongside them.

\subsection{Transfer and Adaptation Provenance}
Geographic-group test scenes, external scenes, and held-out cities have different relationships to the training data.
The six-city split in Appendix~\ref{app:data-splits} provides geographic-group separation.
For target adaptation, the saved scene IDs, base-checkpoint split, and disjoint adaptation/evaluation record lists determine the transfer setting.
The retained adaptation configurations use \texttt{paris\_11}, \texttt{new\_york\_12}, and \texttt{tokyo\_01}; the first two are outside the base dataset, while \texttt{tokyo\_01} belongs to its test split.
All three cities occur in base training.
The labels Zero-shot, Few-shot, and In-distribution in the retained figures refer to adaptation stages; their precise training procedure is given in Appendix~\ref{app:model-adaptation}.

\subsection{Computation and Aggregation}
The single-deployment evaluator measures GPU-synchronized planner execution separately from road mapping and ray tracing.
For target coverage, cumulative planning time adds the completed planner calls across tested TX counts; ray-tracing time and end-to-end wall time are separate fields.
Model evaluations count candidate-level calls, including the perturbed candidates and gradient-bearing calls used by the search method.
Iteration counts, population sizes, model evaluations, and physical simulations are therefore reported as distinct budget quantities.

Repeated trials are summarized with means and standard deviations where available, using matched scene, cardinality, and radio conditions.
Target-coverage failures contribute to success rate rather than disappearing from the comparison.
Numerical results and their existing captions are retained in the experimental sections; this appendix defines the quantities needed to interpret them.

\section{Additional Results and Visualizations}\label{app:additional-results}
\subsection{Urban Scene Visualizations across Six Cities}
\label{app:urban_scene_visualizations}

We present three landmark-centered examples from each of the six cities.
Each example covers a $512\,\mathrm{m}\times512\,\mathrm{m}$ region.
From left to right, each row shows geographically aligned real-world imagery,
the semantic overlay, the building mask, and the TX-deployable mask.
Coordinates are expressed in meters in the local scene frame.
The examples illustrate different urban layouts, including public squares,
dense commercial districts, circular intersections, and waterfront areas.
Imagery acquisition dates may differ from those of the OpenStreetMap
snapshots; the original masks are displayed without modification.
Imagery and map-data credits are retained in each row.

\begingroup
\captionsetup[subfigure]{font=small,labelformat=parens,
  justification=centering,singlelinecheck=false,skip=3pt}
\renewcommand{\thesubfigure}{\roman{subfigure}}

\newcommand{\AutoPlanSceneRow}[2]{%
  \begin{subfigure}{\linewidth}
    \centering
    \includegraphics[width=\linewidth]{#1_scene_overview.pdf}
    \caption{#2 (\texttt{\detokenize{#1}}).}
    \label{fig:landmark:#1}
  \end{subfigure}\par\vspace{4pt}
}

\begin{figure}[htbp]
  \centering
  \AutoPlanSceneRow{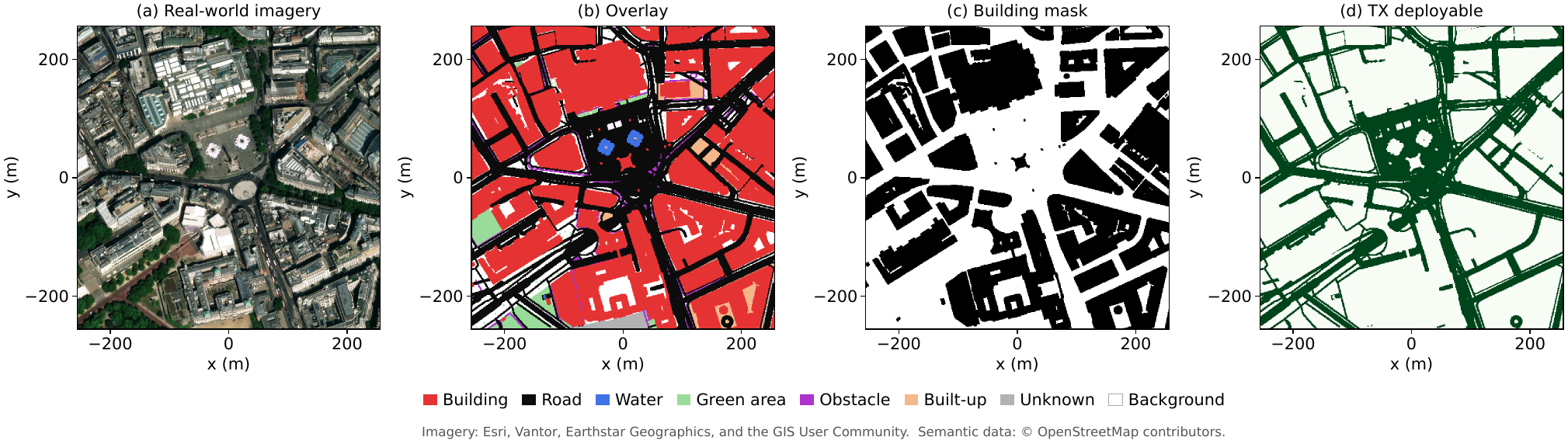}{Trafalgar Square}
  \AutoPlanSceneRow{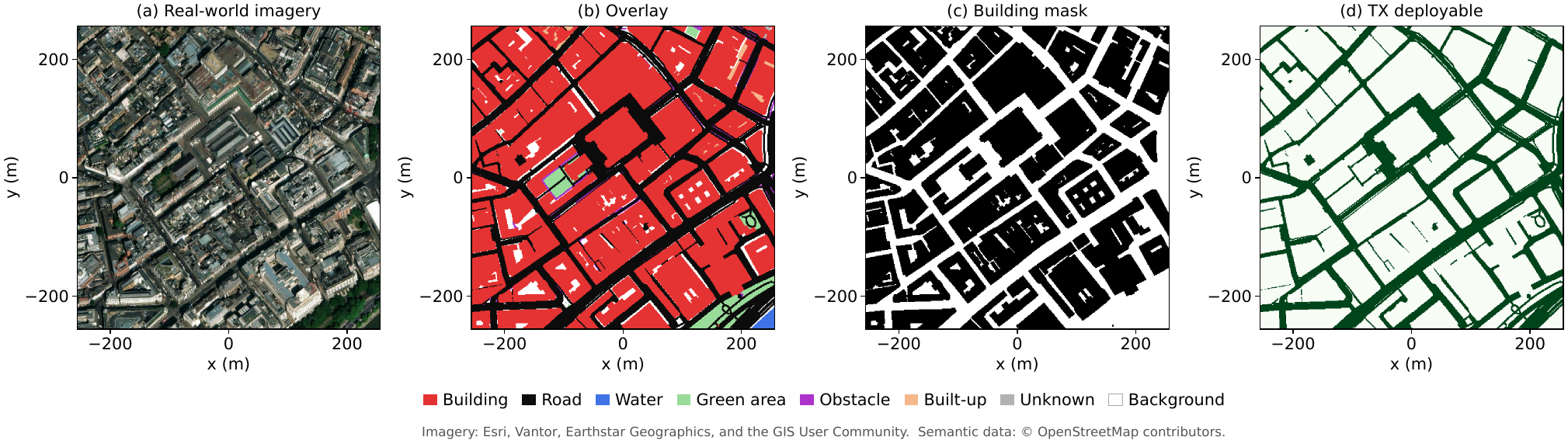}{Covent Garden Market / Royal Opera House}
  \AutoPlanSceneRow{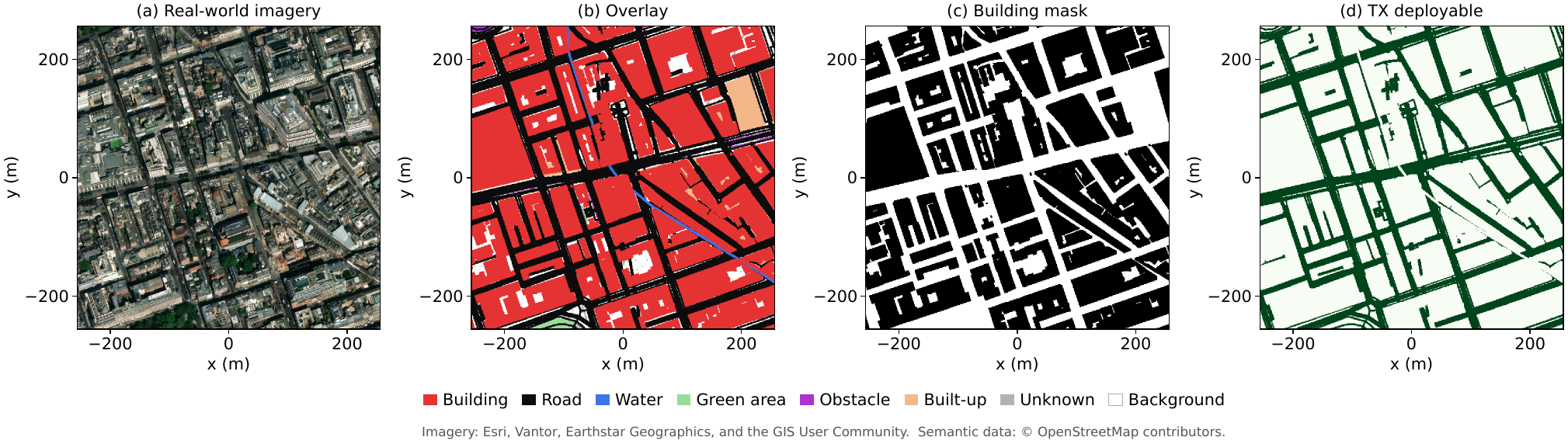}{Oxford Street / Bond Street}
  \caption{Urban scene examples in London. Each row covers a
  $512\,\mathrm{m}\times512\,\mathrm{m}$ region and shows real-world imagery,
  the semantic overlay, the building mask, and the TX-deployable mask,
  from left to right.}
  \label{fig:landmark_london}
\end{figure}
\clearpage

\begin{figure}[htbp]
  \centering
  \AutoPlanSceneRow{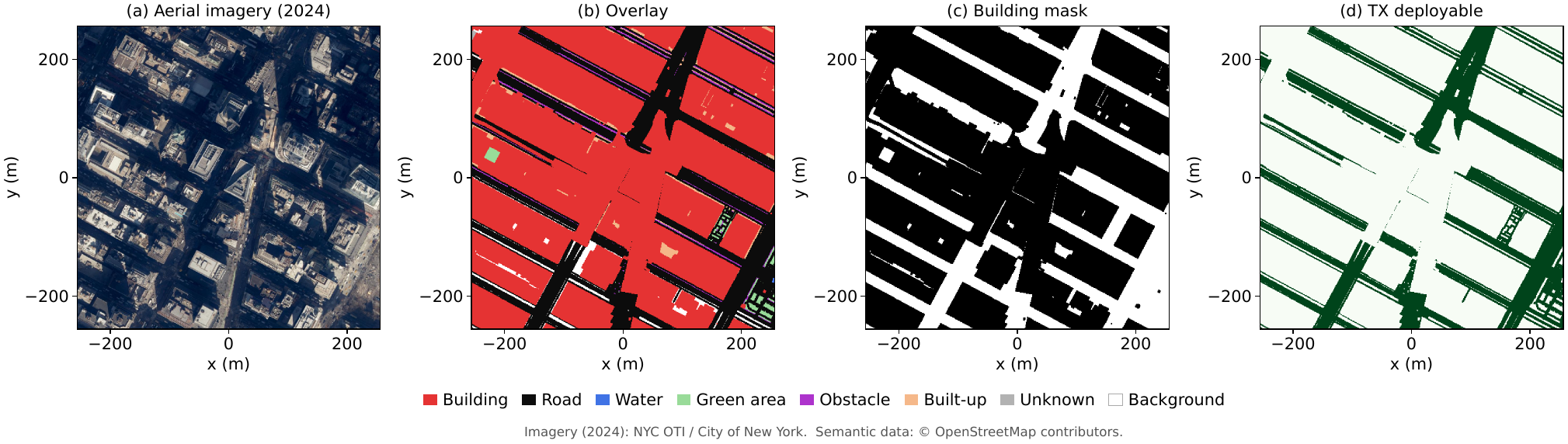}{Times Square south / 42nd Street}
  \AutoPlanSceneRow{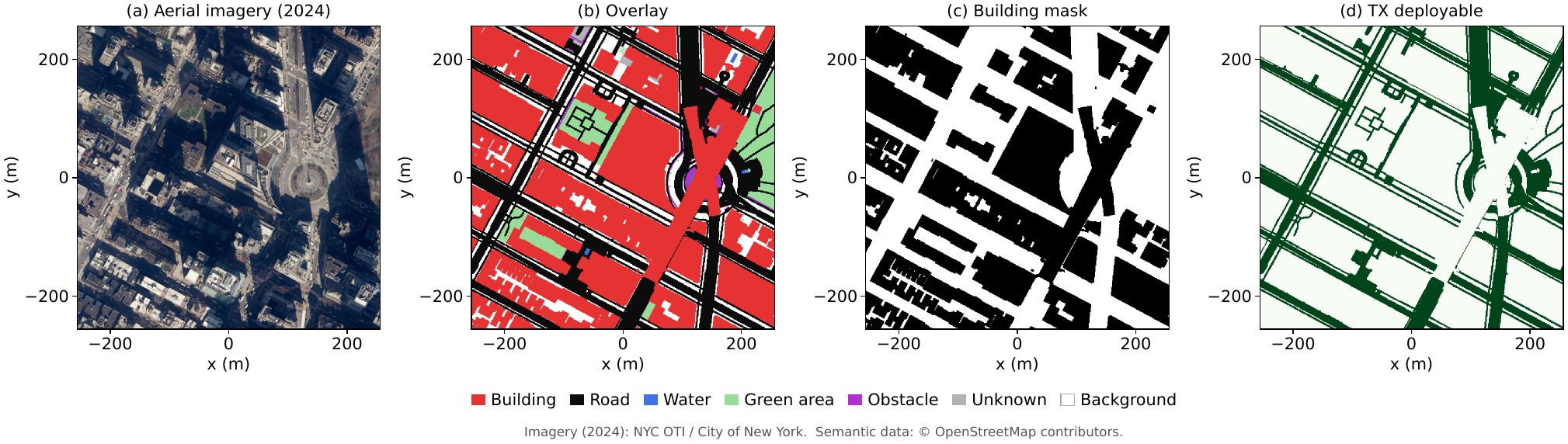}{Columbus Circle}
  \AutoPlanSceneRow{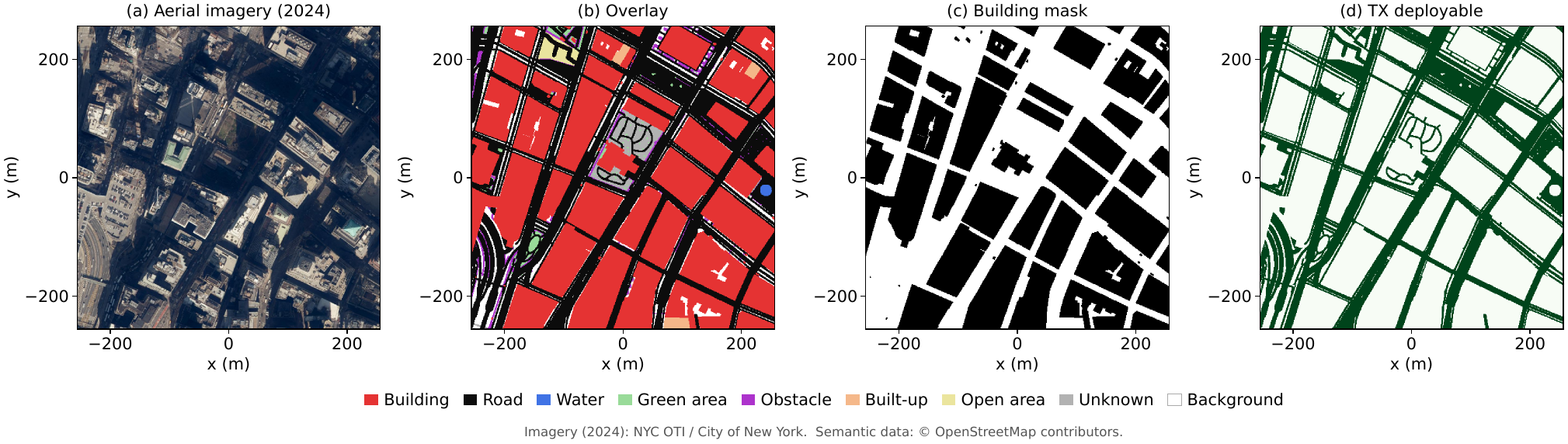}{Wall Street / New York Stock Exchange}
  \caption{Urban scene examples in New York. Each row covers a
  $512\,\mathrm{m}\times512\,\mathrm{m}$ region and shows real-world imagery,
  the semantic overlay, the building mask, and the TX-deployable mask,
  from left to right. The Times Square example shows its southern portion.}
  \label{fig:landmark_new_york}
\end{figure}
\clearpage

\begin{figure}[htbp]
  \centering
  \AutoPlanSceneRow{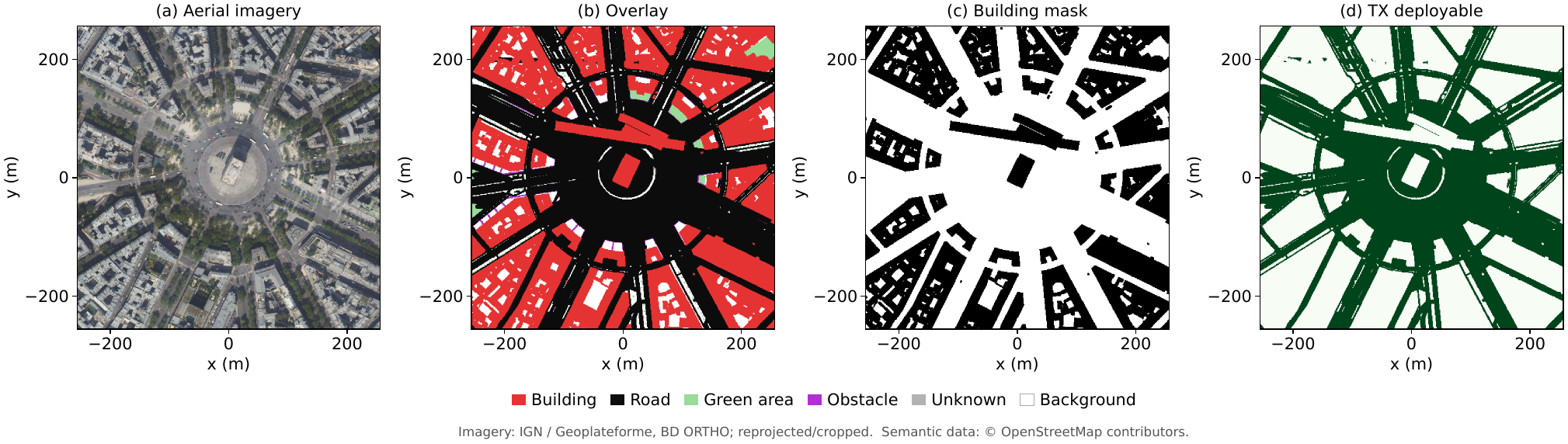}{Arc de Triomphe / Place Charles-de-Gaulle}
  \AutoPlanSceneRow{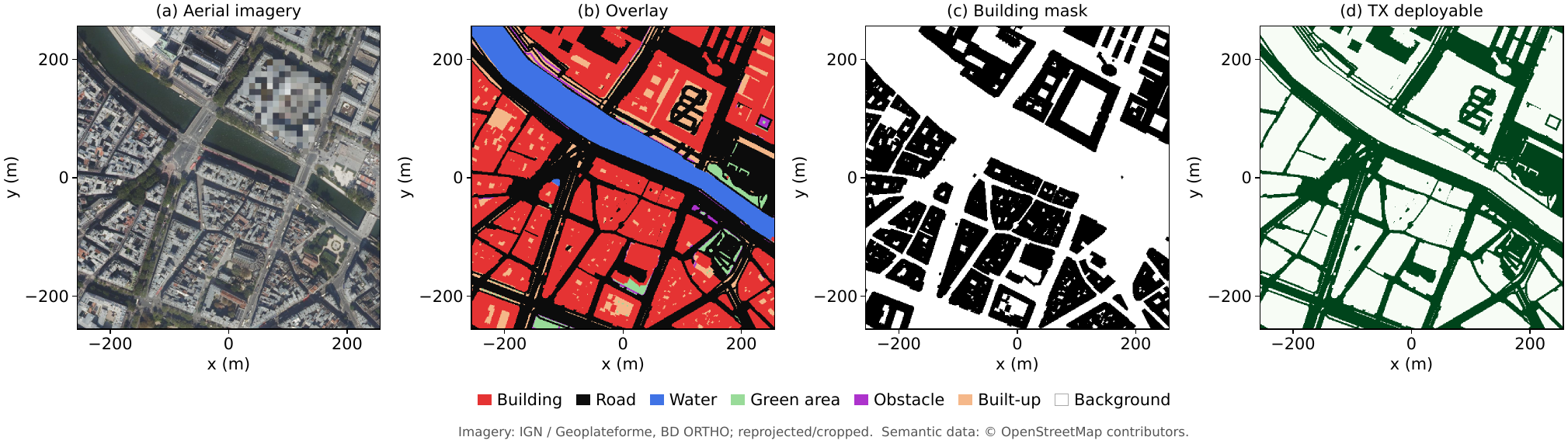}{Place Saint-Michel / Sainte-Chapelle surroundings}
  \AutoPlanSceneRow{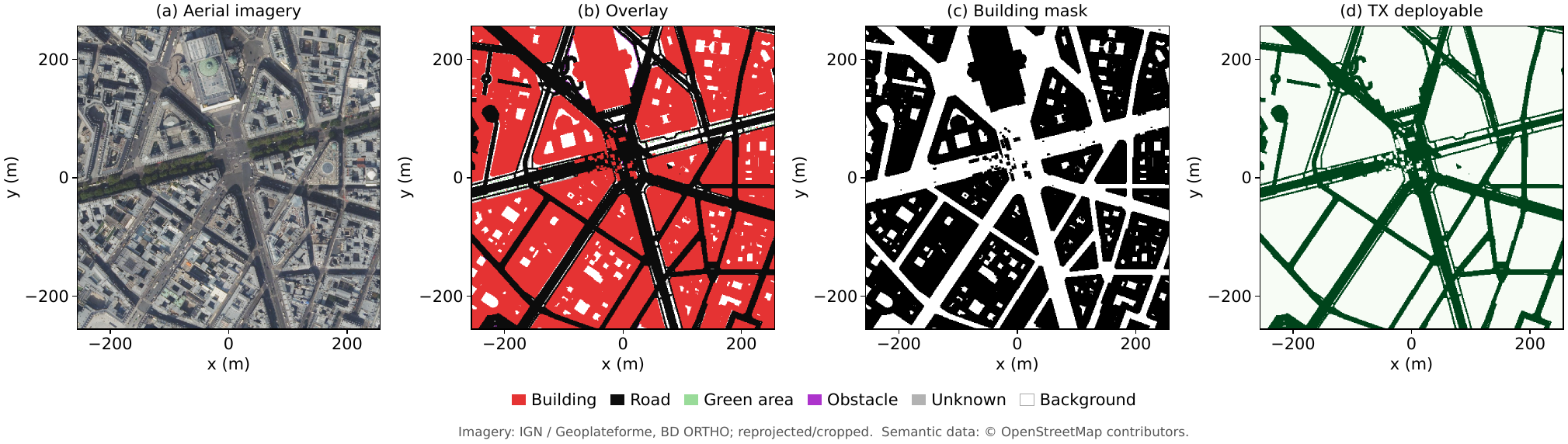}{Palais Garnier / Place de l'Op\'era}
  \caption{Urban scene examples in Paris. Each row covers a
  $512\,\mathrm{m}\times512\,\mathrm{m}$ region and shows real-world imagery,
  the semantic overlay, the building mask, and the TX-deployable mask,
  from left to right. The pixelated patch in the second row is present
  in the source imagery.}
  \label{fig:landmark_paris}
\end{figure}
\clearpage

\begin{figure}[htbp]
  \centering
  \AutoPlanSceneRow{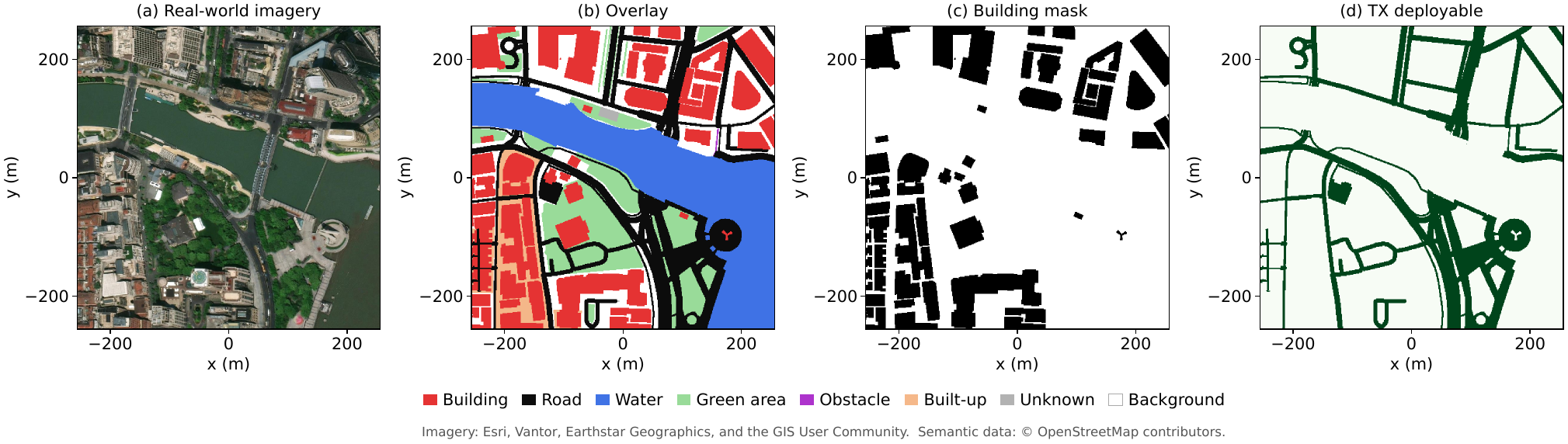}{Waibaidu Bridge / Huangpu Park}
  \AutoPlanSceneRow{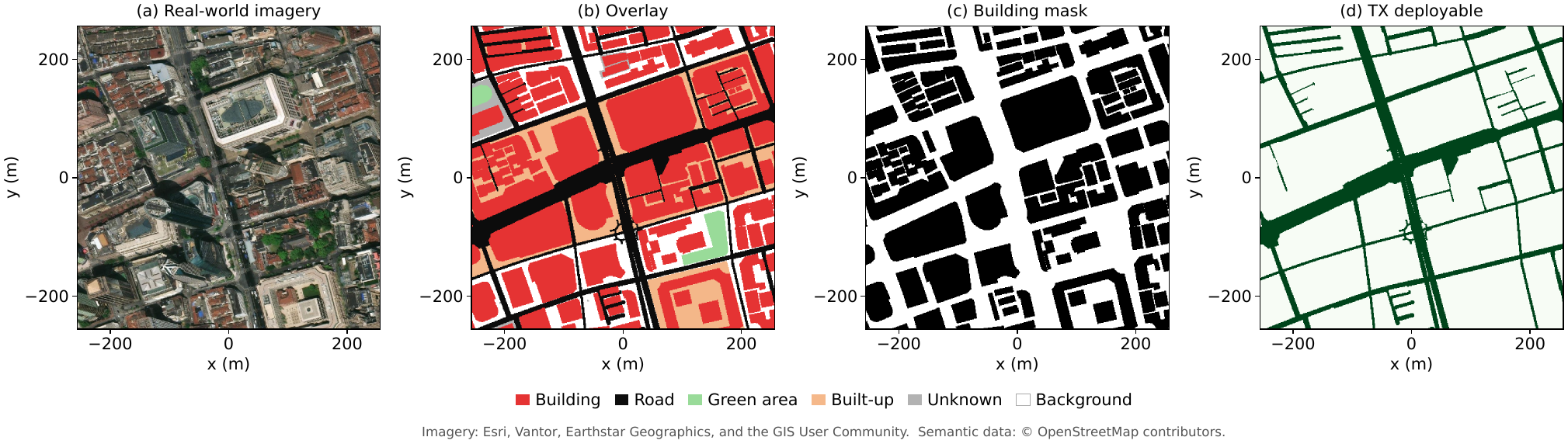}{East Nanjing Road pedestrian street}
  \AutoPlanSceneRow{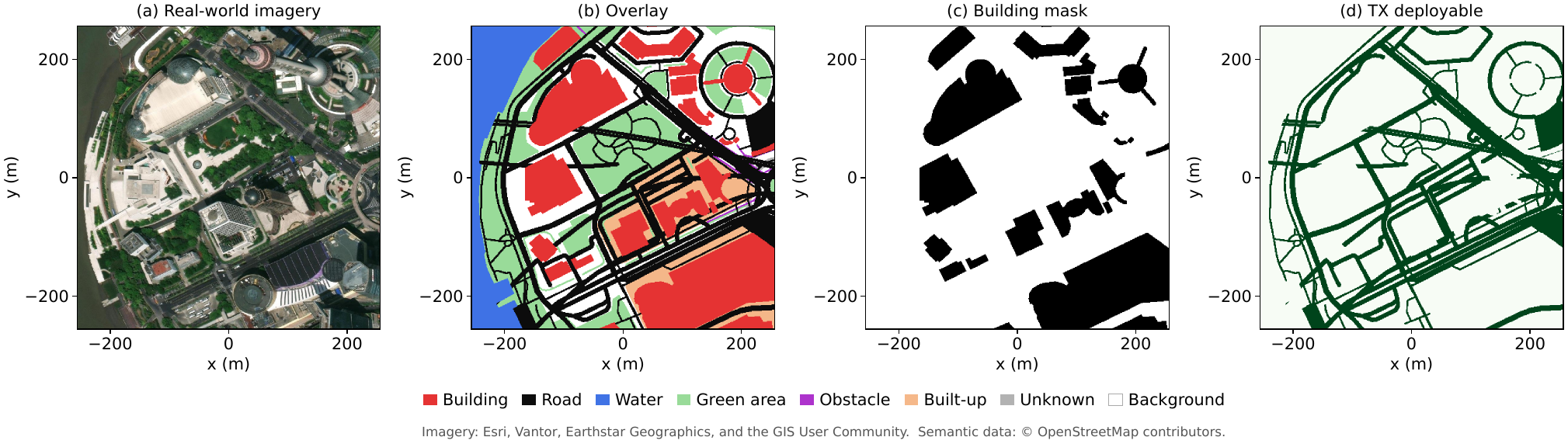}{Oriental Pearl Tower / Lujiazui waterfront}
  \caption{Urban scene examples in Shanghai. Each row covers a
  $512\,\mathrm{m}\times512\,\mathrm{m}$ region and shows real-world imagery,
  the semantic overlay, the building mask, and the TX-deployable mask,
  from left to right.}
  \label{fig:landmark_shanghai}
\end{figure}
\clearpage

\begin{figure}[htbp]
  \centering
  \AutoPlanSceneRow{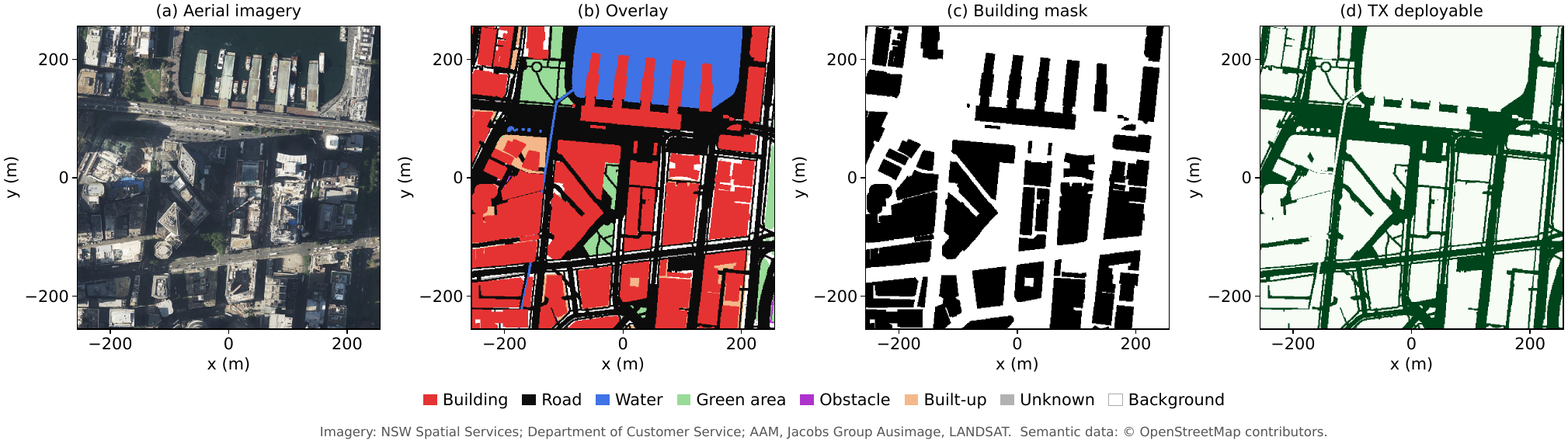}{Circular Quay / Customs House}
  \AutoPlanSceneRow{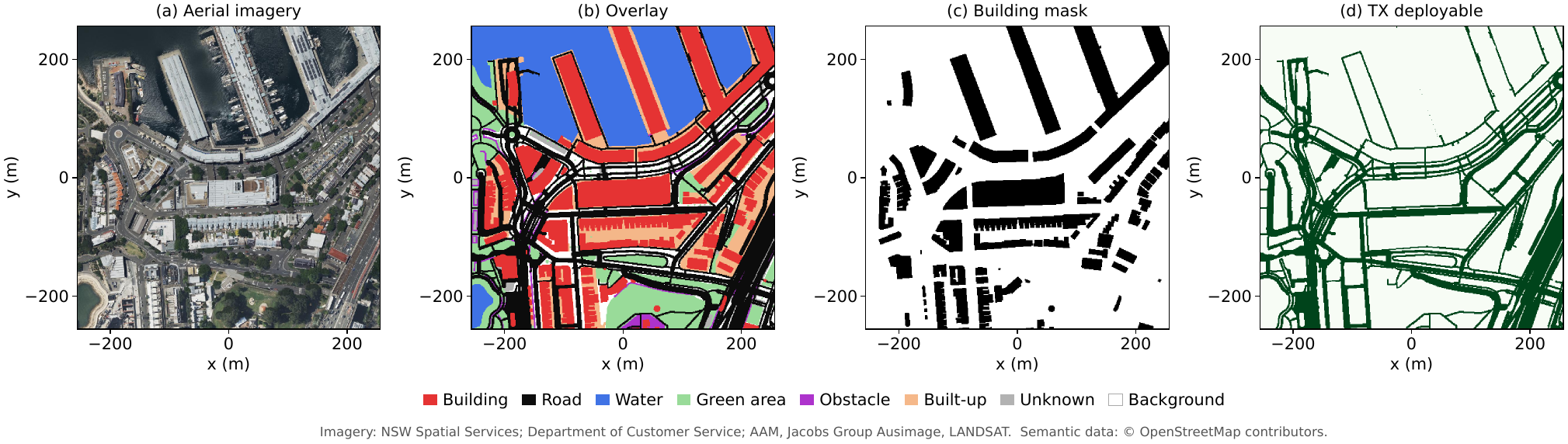}{Observatory Hill / The Rocks surroundings}
  \AutoPlanSceneRow{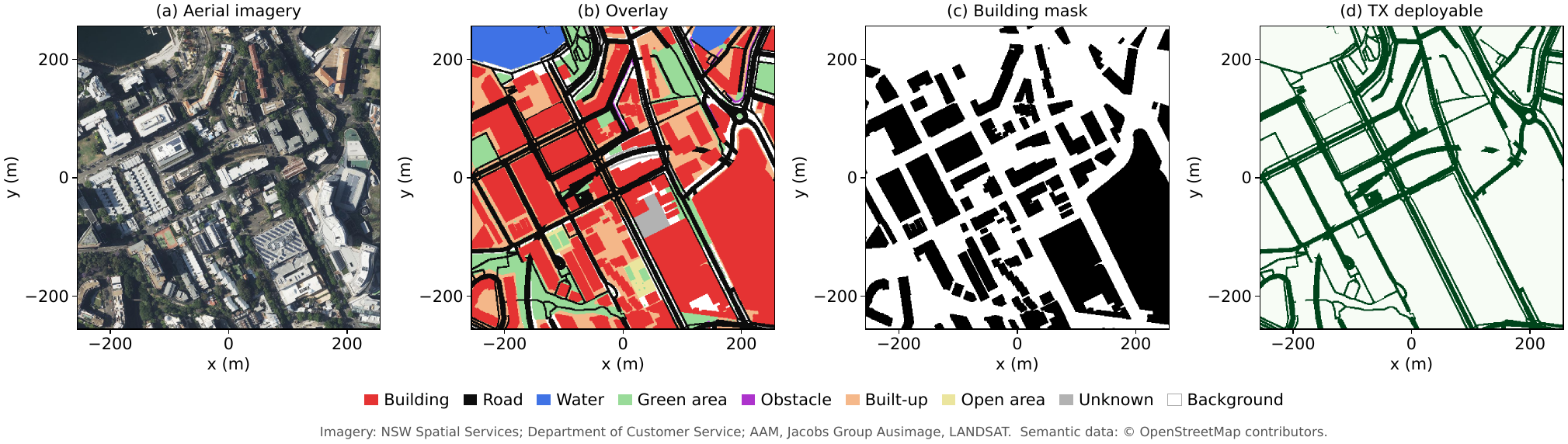}{The Star / Darling Island, Pyrmont}
  \caption{Urban scene examples in Sydney. Each row covers a
  $512\,\mathrm{m}\times512\,\mathrm{m}$ region and shows real-world imagery,
  the semantic overlay, the building mask, and the TX-deployable mask,
  from left to right.}
  \label{fig:landmark_sydney}
\end{figure}
\clearpage

\begin{figure}[htbp]
  \centering
  \AutoPlanSceneRow{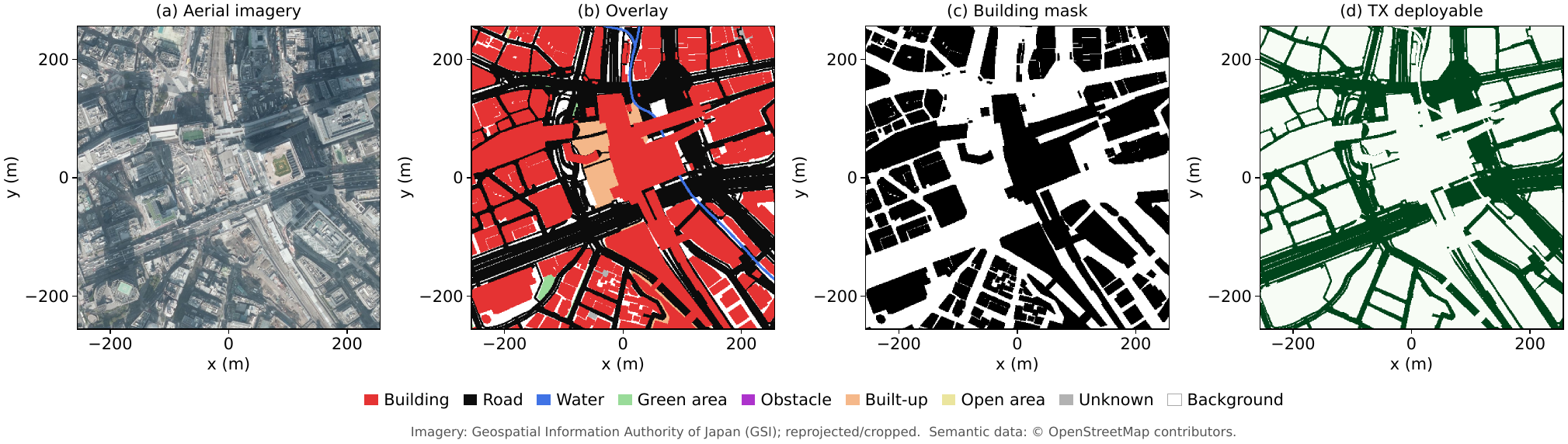}{Shibuya Scramble Crossing}
  \AutoPlanSceneRow{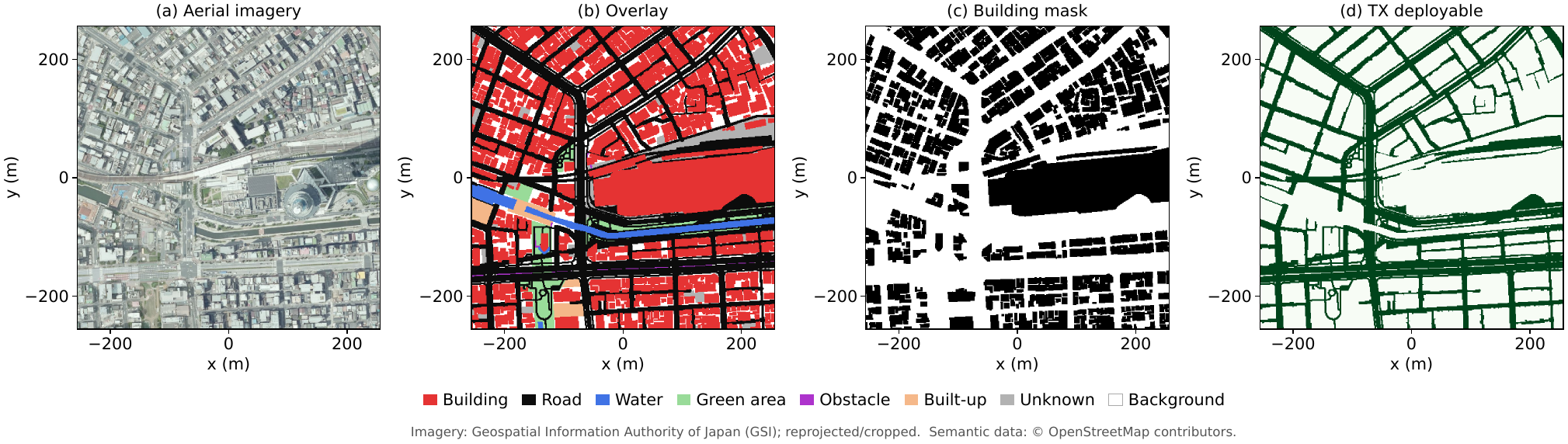}{Tokyo Skytree / Tokyo Solamachi}
  \AutoPlanSceneRow{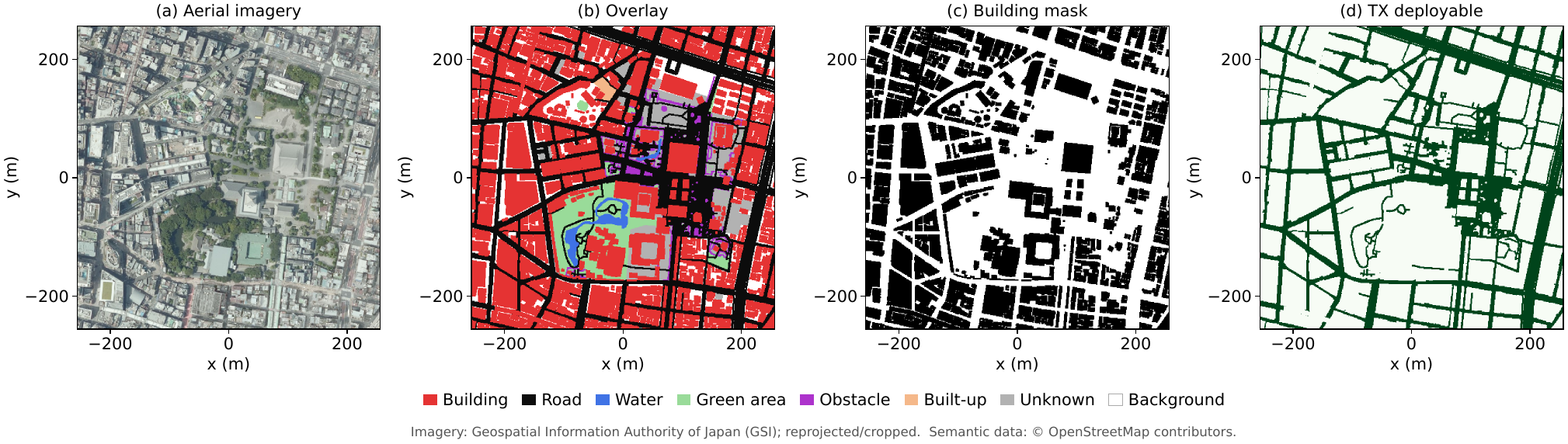}{Sens\=o-ji / Nakamise}
  \caption{Urban scene examples in Tokyo. Each row covers a
  $512\,\mathrm{m}\times512\,\mathrm{m}$ region and shows real-world imagery,
  the semantic overlay, the building mask, and the TX-deployable mask,
  from left to right.}
  \label{fig:landmark_tokyo}
\end{figure}
\clearpage
\endgroup


\clearpage


\clearpage
\subsection{Qualitative Comparison of Deployment Layouts and Radio Maps}
\label{app:deployment_radio_maps}

We visualize deployment layouts and Sionna-evaluated radio maps
for nine sampling methods across six urban scenes.
All methods use HPEM as the reward model and deploy
$N_{\mathrm{TX}}=9$ transmitters within a
$512\,\mathrm{m}\times512\,\mathrm{m}$ region.

From top to bottom, the rows correspond to Diffusion, WS-L, WS-G,
NUTS, BR-SNIS, PSO, GA, Greedy, and TuRBO.
From left to right, the columns show joint coverage, path loss,
SS-RSRP, SINR, and effective throughput.
Stars indicate TX locations, and each panel title reports
the corresponding coverage rate.
In the joint coverage maps, red and blue denote road locations
that satisfy and fail the joint coverage criterion, respectively.

Each figure uses a common seed across all nine methods and
shows individual deployments rather than seed-averaged results.
Seeds were selected based on availability across methods,
not deployment performance.
Continuous metric maps retain their original per-panel color scales;
their numerical colorbar values should therefore be consulted
when comparing panels.

\begin{figure}[p]
    \centering
    \includegraphics[
        width=\linewidth,
        height=0.86\textheight,
        keepaspectratio
    ]{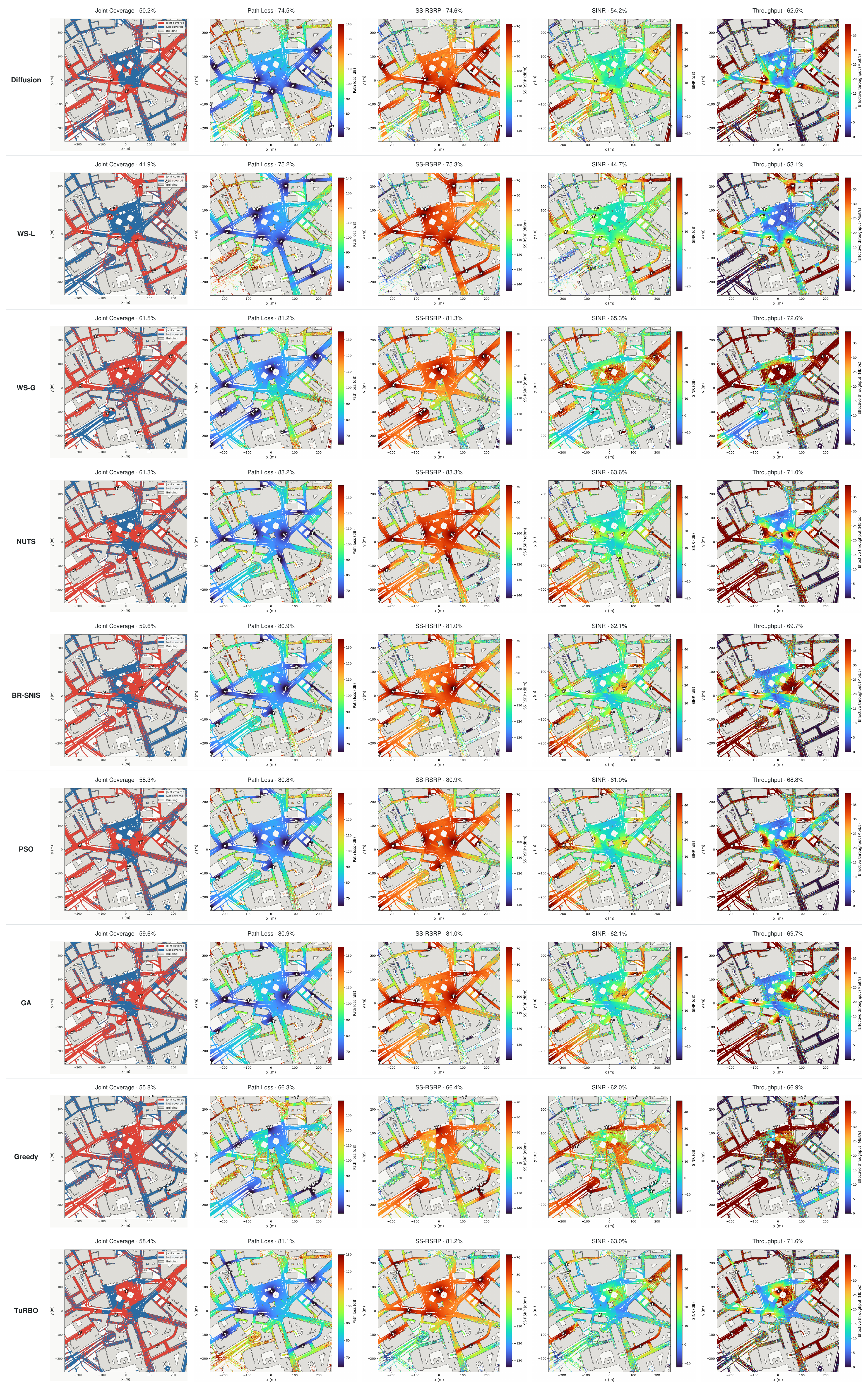}
    \caption{
    Deployment layouts and radio maps for nine HPEM-guided sampling
    methods in London.
    The scene is \texttt{london\_01}, with
    $N_{\mathrm{TX}}=9$ over a
    $512\,\mathrm{m}\times512\,\mathrm{m}$ region.
    All rows use random seed $62$.
    Results are individual Sionna-evaluated deployments.
    }
    \label{fig:appendix_radio_maps_london}
\end{figure}
\clearpage

\begin{figure}[p]
    \centering
    \includegraphics[
        width=\linewidth,
        height=0.86\textheight,
        keepaspectratio
    ]{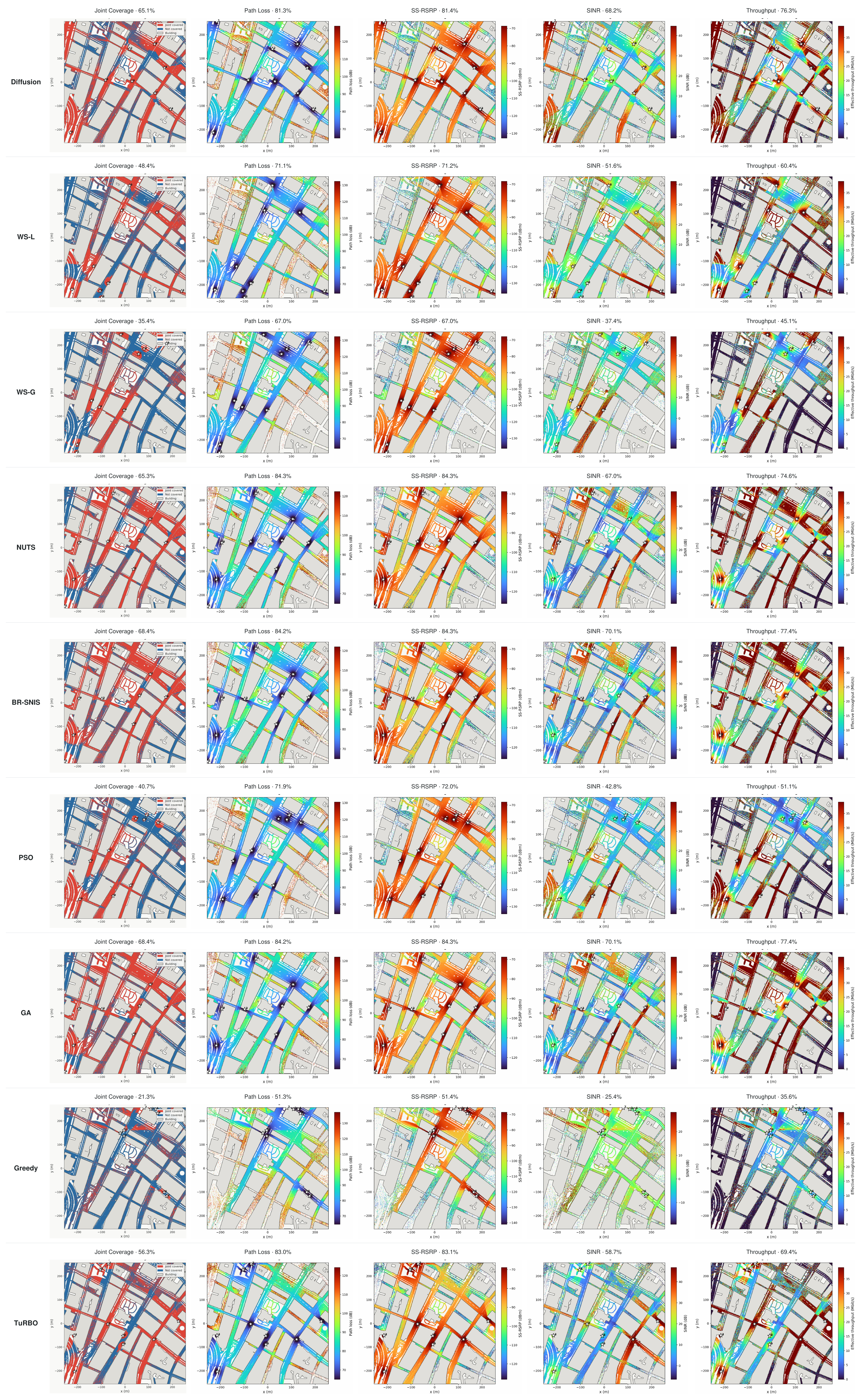}
    \caption{
    Deployment layouts and radio maps for nine HPEM-guided sampling
    methods in New York.
    The scene is \texttt{new\_york\_04}, with
    $N_{\mathrm{TX}}=9$ over a
    $512\,\mathrm{m}\times512\,\mathrm{m}$ region.
    All rows use random seed $62$.
    Results are individual Sionna-evaluated deployments.
    }
    \label{fig:appendix_radio_maps_new_york}
\end{figure}
\clearpage

\begin{figure}[p]
    \centering
    \includegraphics[
        width=\linewidth,
        height=0.86\textheight,
        keepaspectratio
    ]{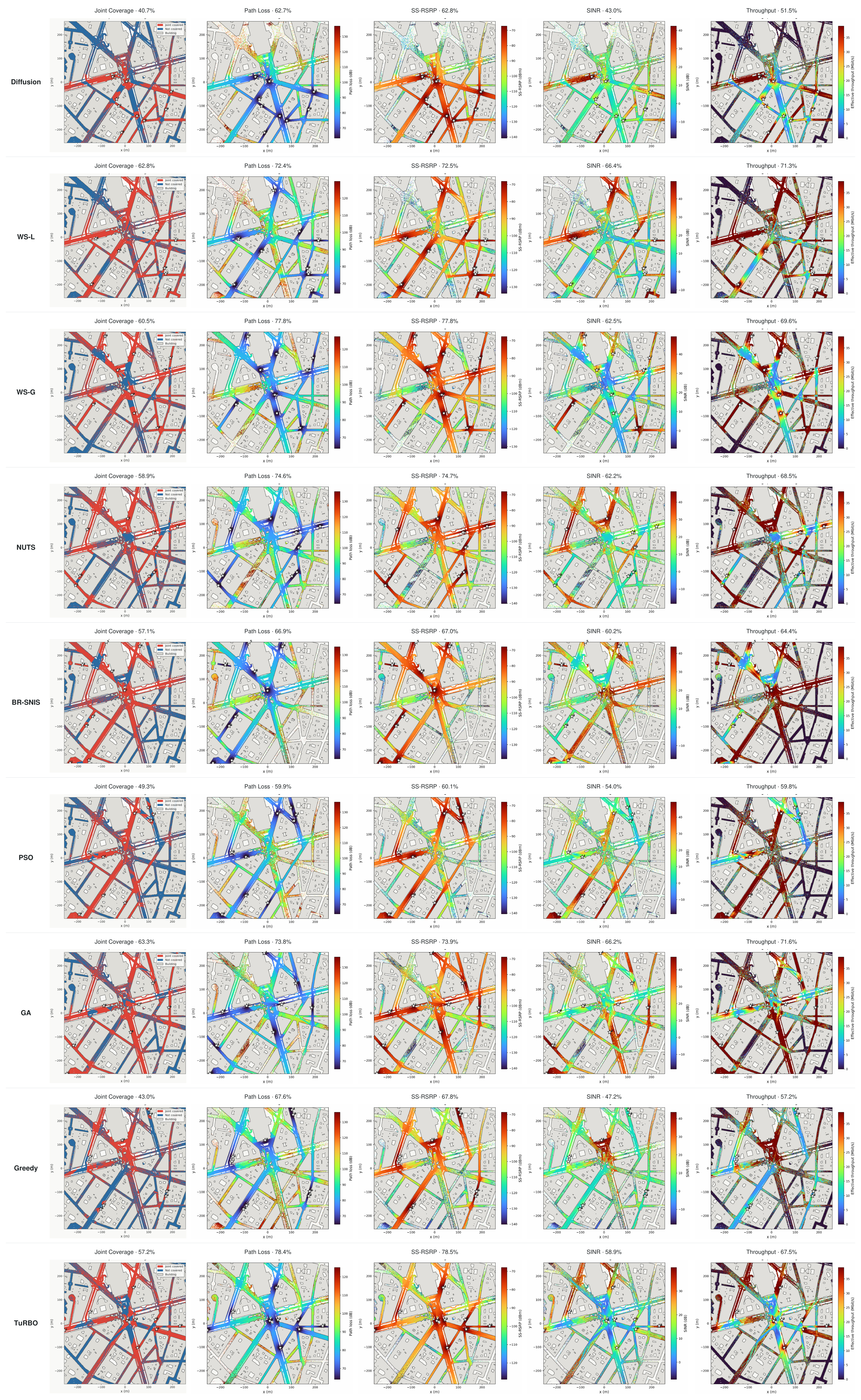}
    \caption{
    Deployment layouts and radio maps for nine HPEM-guided sampling
    methods in Paris.
    The scene is \texttt{paris\_05}, with
    $N_{\mathrm{TX}}=9$ over a
    $512\,\mathrm{m}\times512\,\mathrm{m}$ region.
    All rows use random seed $62$.
    Results are individual Sionna-evaluated deployments.
    }
    \label{fig:appendix_radio_maps_paris}
\end{figure}
\clearpage

\begin{figure}[p]
    \centering
    \includegraphics[
        width=\linewidth,
        height=0.86\textheight,
        keepaspectratio
    ]{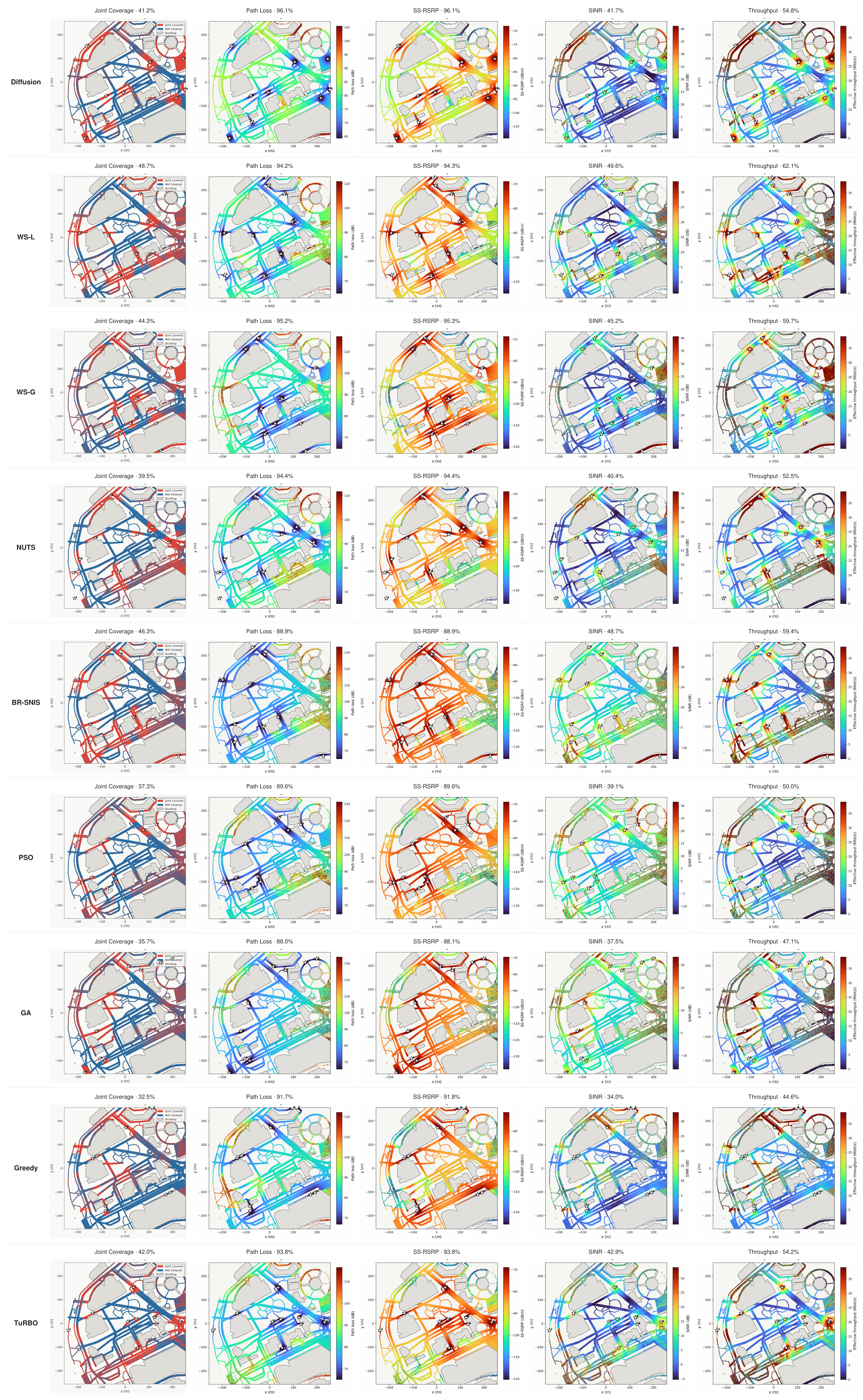}
    \caption{
    Deployment layouts and radio maps for nine HPEM-guided sampling
    methods in Shanghai.
    The scene is \texttt{shanghai\_05}, with
    $N_{\mathrm{TX}}=9$ over a
    $512\,\mathrm{m}\times512\,\mathrm{m}$ region.
    All rows use random seed $62$.
    Results are individual Sionna-evaluated deployments.
    }
    \label{fig:appendix_radio_maps_shanghai}
\end{figure}
\clearpage

\begin{figure}[p]
    \centering
    \includegraphics[
        width=\linewidth,
        height=0.86\textheight,
        keepaspectratio
    ]{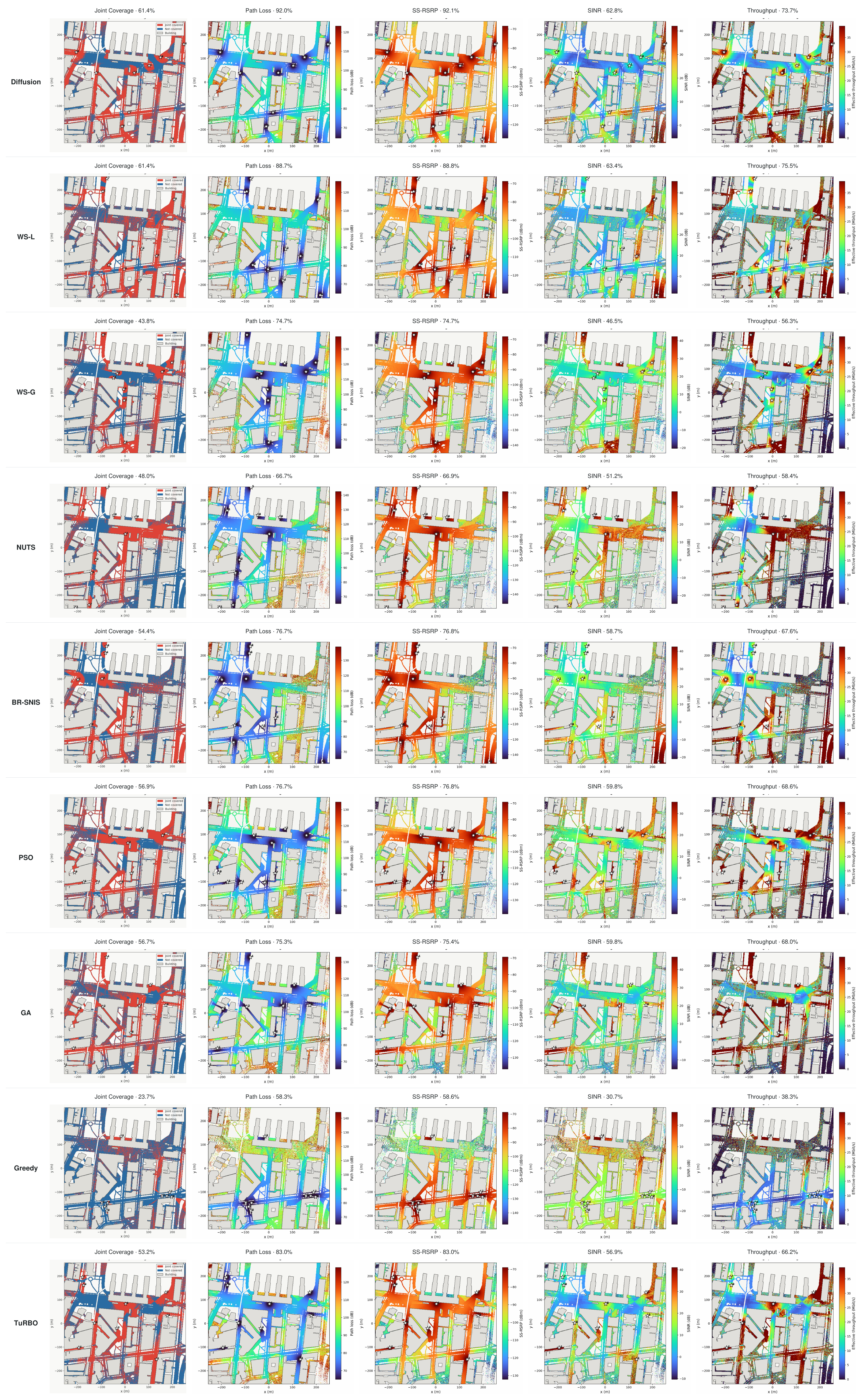}
    \caption{
    Deployment layouts and radio maps for nine HPEM-guided sampling
    methods in Sydney.
    The scene is \texttt{sydney\_01}, with
    $N_{\mathrm{TX}}=9$ over a
    $512\,\mathrm{m}\times512\,\mathrm{m}$ region.
    All rows use random seed $62$.
    Results are individual Sionna-evaluated deployments.
    }
    \label{fig:appendix_radio_maps_sydney}
\end{figure}
\clearpage

\begin{figure}[p]
    \centering
    \includegraphics[
        width=\linewidth,
        height=0.86\textheight,
        keepaspectratio
    ]{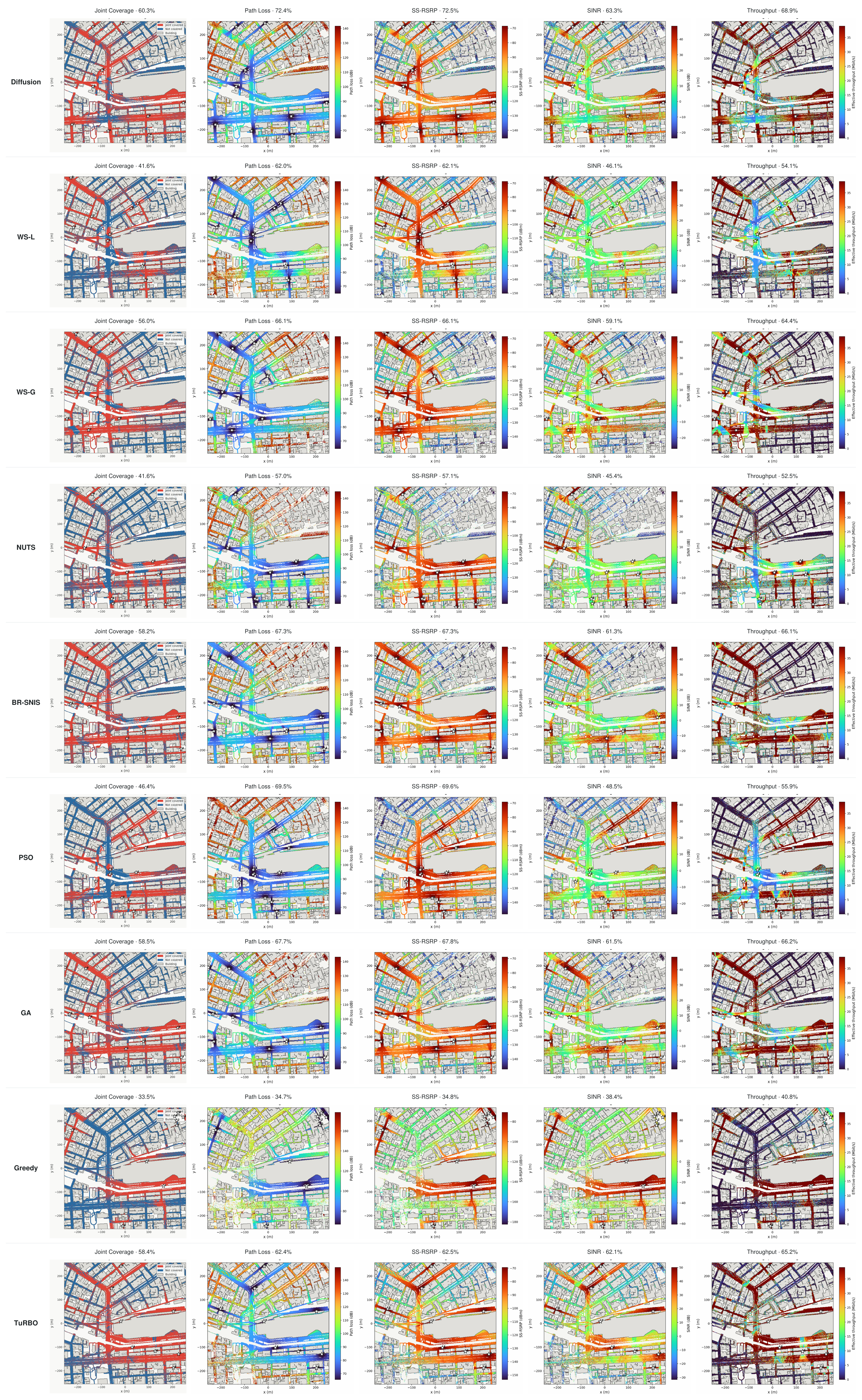}
    \caption{
    Deployment layouts and radio maps for nine HPEM-guided sampling
    methods in Tokyo.
    The scene is \texttt{tokyo\_03}, with
    $N_{\mathrm{TX}}=9$ over a
    $512\,\mathrm{m}\times512\,\mathrm{m}$ region.
    All rows use random seed $42$.
    Results are individual Sionna-evaluated deployments.
    }
    \label{fig:appendix_radio_maps_tokyo}
\end{figure}

\clearpage
\section{Simulation results with fixed Tx number}
\label{append:Simulation results with fixed Tx number}
\begin{table*}[ht]
\centering
\caption{
\textbf{Performance comparison across different TX deployment scales in the $128\,\mathrm{m}\times128\,\mathrm{m}$ Sydney scene.}
The simulation results are reported as mean$\pm$std. 
For each BS scale and evaluation metric, the \textit{top three} methods are highlighted in 
\first{(1st)}, \second{(2nd)}, and \third{(3rd)}, respectively, 
while the \worst{worst} result in each column is underlined.
}
\label{tab:illustrative_main_results}
\setlength{\tabcolsep}{6pt}
\renewcommand{\arraystretch}{1.00}
\scriptsize

\begin{adjustbox}
{width=\textwidth,totalheight=0.92\textheight,keepaspectratio}

\begin{tabular}{cccccccc}
\toprule
\textbf{TXs}
& \textbf{Method}
& \textbf{RT (s)$\downarrow$}
& \textbf{JM (\%)$\uparrow$}
& \textbf{PL (\%)$\uparrow$}
& \textbf{TP (\%)$\uparrow$}
& \textbf{SINR (\%)$\uparrow$}
& \textbf{RSRP (\%)$\uparrow$} \\
\midrule

\multirow{9}{*}{1}

& Diffusion
& \second{4.82$\pm$0.21}
& \second{91.34$\pm$0.84}
& \second{94.62$\pm$0.72}
& \first{96.48$\pm$0.58}
& \second{93.11$\pm$0.91}
& \second{95.08$\pm$0.69} \\

& WS-G
& 5.46$\pm$0.28
& 90.42$\pm$0.96
& 93.74$\pm$0.87
& 95.48$\pm$0.72
& 92.18$\pm$1.00
& 94.16$\pm$0.84 \\

& WS-L
& 6.91$\pm$0.37
& 90.08$\pm$1.01
& 93.41$\pm$0.91
& 95.16$\pm$0.78
& 91.87$\pm$1.04
& 93.89$\pm$0.88 \\

\addlinespace[1.5pt]

& NUTS
& 8.14$\pm$0.52
& \third{90.91$\pm$0.93}
& \third{94.20$\pm$0.81}
& \second{96.11$\pm$0.66}
& \third{92.72$\pm$0.97}
& \third{94.73$\pm$0.78} \\

& BR-SNIS
& \third{5.13$\pm$0.24}
& 88.96$\pm$1.18
& 92.36$\pm$1.08
& 94.22$\pm$0.93
& 90.71$\pm$1.21
& 92.65$\pm$1.04 \\

\addlinespace[1.5pt]

& Greedy
& \first{1.79$\pm$0.13}
& \worst{86.72$\pm$0.69}
& \worst{90.18$\pm$0.63}
& \worst{92.04$\pm$0.58}
& \worst{88.27$\pm$0.71}
& \worst{90.56$\pm$0.61} \\

& PSO
& 13.72$\pm$0.88
& 89.84$\pm$1.03
& 93.05$\pm$0.94
& 94.91$\pm$0.82
& 91.43$\pm$1.07
& 93.46$\pm$0.90 \\

& GA
& \worst{15.18$\pm$1.03}
& 87.63$\pm$1.16
& 91.08$\pm$1.05
& 92.91$\pm$0.91
& 89.21$\pm$1.19
& 91.44$\pm$1.01 \\

& TuRBO
& 10.96$\pm$0.71
& \first{91.70$\pm$0.79}
& \first{94.91$\pm$0.69}
& \third{95.96$\pm$0.63}
& \first{93.46$\pm$0.86}
& \first{95.31$\pm$0.71} \\

\midrule

\multirow{9}{*}{2}

& Diffusion
& \second{14.62$\pm$0.47}
& \first{95.08$\pm$0.62}
& \second{97.01$\pm$0.54}
& \first{98.21$\pm$0.42}
& \first{96.37$\pm$0.67}
& \second{97.48$\pm$0.51} \\

& WS-G
& 16.18$\pm$0.61
& 94.21$\pm$0.78
& 96.42$\pm$0.70
& 97.61$\pm$0.57
& 95.42$\pm$0.82
& 96.71$\pm$0.66 \\

& WS-L
& 20.06$\pm$0.79
& 93.88$\pm$0.84
& 96.11$\pm$0.75
& 97.34$\pm$0.63
& 95.08$\pm$0.88
& 96.39$\pm$0.71 \\

\addlinespace[1.5pt]

& NUTS
& 23.42$\pm$0.88
& \third{94.62$\pm$0.71}
& \third{96.72$\pm$0.63}
& \second{97.94$\pm$0.49}
& \second{96.01$\pm$0.73}
& \third{97.22$\pm$0.58} \\

& BR-SNIS
& \third{15.37$\pm$0.54}
& 92.94$\pm$0.91
& 95.19$\pm$0.82
& 96.56$\pm$0.69
& 94.04$\pm$0.95
& 95.51$\pm$0.79 \\

\addlinespace[1.5pt]

& Greedy
& \first{5.61$\pm$0.29}
& \worst{90.84$\pm$0.66}
& \worst{93.21$\pm$0.61}
& \worst{94.86$\pm$0.54}
& \worst{91.93$\pm$0.69}
& \worst{93.67$\pm$0.59} \\

& PSO
& 48.76$\pm$1.91
& 93.76$\pm$0.80
& 95.94$\pm$0.72
& 97.18$\pm$0.60
& 94.91$\pm$0.84
& 96.22$\pm$0.68 \\

& GA
& \worst{55.42$\pm$2.34}
& 91.92$\pm$0.98
& 94.24$\pm$0.88
& 95.61$\pm$0.76
& 93.02$\pm$1.02
& 94.48$\pm$0.85 \\

& TuRBO
& 34.91$\pm$1.21
& \second{94.91$\pm$0.66}
& \first{97.16$\pm$0.57}
& \third{97.82$\pm$0.52}
& \third{95.82$\pm$0.76}
& \first{97.61$\pm$0.55} \\

\bottomrule
\end{tabular}

\end{adjustbox}
\end{table*}

\begin{table*}[ht]
\centering
\caption{
\textbf{Performance comparison across different TX deployment scales in the $256\,\mathrm{m}\times256\,\mathrm{m}$ Paris scene.}
The simulation results are reported as mean$\pm$std. 
For each BS scale and evaluation metric, the \textit{top three} methods are highlighted in 
\first{(1st)}, \second{(2nd)}, and \third{(3rd)}, respectively, 
while the \worst{worst} result in each column is underlined.
}
\label{tab:illustrative_sydney256}
\setlength{\tabcolsep}{6pt}
\renewcommand{\arraystretch}{1.00}
\scriptsize

\begin{adjustbox}
{width=\textwidth,totalheight=0.92\textheight,keepaspectratio}

\begin{tabular}{cccccccc}
\toprule
\textbf{TXs}
& \textbf{Method}
& \textbf{RT (s)$\downarrow$}
& \textbf{JM (\%)$\uparrow$}
& \textbf{PL (\%)$\uparrow$}
& \textbf{TP (\%)$\uparrow$}
& \textbf{SINR (\%)$\uparrow$}
& \textbf{RSRP (\%)$\uparrow$} \\
\midrule

\multirow{9}{*}{2}

& Diffusion
& \second{18.74$\pm$0.62}
& \second{76.82$\pm$1.12}
& \second{84.91$\pm$1.06}
& \first{89.12$\pm$0.91}
& \second{80.63$\pm$1.21}
& \second{86.47$\pm$1.02} \\

& WS-G
& 21.06$\pm$0.78
& 75.31$\pm$1.34
& 83.42$\pm$1.26
& 87.61$\pm$1.08
& 79.02$\pm$1.42
& 84.93$\pm$1.21 \\

& WS-L
& 24.18$\pm$0.91
& 74.96$\pm$1.41
& 83.08$\pm$1.32
& 87.28$\pm$1.14
& 78.67$\pm$1.48
& 84.57$\pm$1.27 \\

\addlinespace[1.5pt]

& NUTS
& 29.73$\pm$1.06
& \third{76.31$\pm$1.19}
& \third{84.47$\pm$1.11}
& \second{88.71$\pm$0.96}
& \third{80.21$\pm$1.26}
& \third{86.02$\pm$1.07} \\

& BR-SNIS
& \third{20.11$\pm$0.71}
& 73.18$\pm$1.47
& 81.27$\pm$1.39
& 85.42$\pm$1.21
& 76.89$\pm$1.53
& 82.74$\pm$1.34 \\

\addlinespace[1.5pt]

& Greedy
& \first{6.87$\pm$0.34}
& \worst{68.42$\pm$0.93}
& \worst{76.21$\pm$0.88}
& \worst{80.53$\pm$0.79}
& \worst{71.94$\pm$1.01}
& \worst{77.82$\pm$0.86} \\

& PSO
& 61.34$\pm$2.18
& 74.13$\pm$1.39
& 82.17$\pm$1.31
& 86.36$\pm$1.13
& 77.86$\pm$1.46
& 83.68$\pm$1.25 \\

& GA
& \worst{69.82$\pm$2.64}
& 70.91$\pm$1.56
& 78.86$\pm$1.47
& 83.11$\pm$1.29
& 74.48$\pm$1.62
& 80.39$\pm$1.42 \\

& TuRBO
& 44.36$\pm$1.52
& \first{77.03$\pm$1.07}
& \first{85.18$\pm$1.01}
& \third{88.46$\pm$0.98}
& \first{80.94$\pm$1.18}
& \first{86.73$\pm$0.99} \\

\midrule

\multirow{9}{*}{3}

& Diffusion
& \second{28.61$\pm$0.88}
& \first{83.06$\pm$0.94}
& \second{89.48$\pm$0.89}
& \first{92.63$\pm$0.76}
& \first{86.39$\pm$1.03}
& \second{90.74$\pm$0.86} \\

& WS-G
& 32.14$\pm$1.07
& 81.56$\pm$1.16
& 88.02$\pm$1.08
& 91.17$\pm$0.92
& 84.86$\pm$1.22
& 89.27$\pm$1.04 \\

& WS-L
& 36.72$\pm$1.24
& 81.21$\pm$1.21
& 87.68$\pm$1.13
& 90.84$\pm$0.97
& 84.51$\pm$1.28
& 88.91$\pm$1.09 \\

\addlinespace[1.5pt]

& NUTS
& 44.53$\pm$1.41
& \third{82.47$\pm$1.01}
& \third{89.03$\pm$0.95}
& \second{92.21$\pm$0.81}
& \second{85.98$\pm$1.08}
& \third{90.28$\pm$0.91} \\

& BR-SNIS
& \third{30.47$\pm$0.96}
& 79.36$\pm$1.28
& 85.79$\pm$1.21
& 88.96$\pm$1.05
& 82.67$\pm$1.35
& 87.14$\pm$1.17 \\

\addlinespace[1.5pt]

& Greedy
& \first{10.41$\pm$0.46}
& \worst{74.21$\pm$0.84}
& \worst{80.46$\pm$0.81}
& \worst{83.72$\pm$0.73}
& \worst{77.43$\pm$0.91}
& \worst{81.82$\pm$0.78} \\

& PSO
& 91.58$\pm$2.91
& 80.31$\pm$1.21
& 86.72$\pm$1.14
& 89.91$\pm$0.98
& 83.61$\pm$1.28
& 88.08$\pm$1.09 \\

& GA
& \worst{103.46$\pm$3.42}
& 76.63$\pm$1.41
& 82.94$\pm$1.34
& 86.17$\pm$1.16
& 79.84$\pm$1.48
& 84.27$\pm$1.28 \\

& TuRBO
& 65.82$\pm$2.06
& \second{82.84$\pm$0.97}
& \first{89.71$\pm$0.91}
& \third{91.94$\pm$0.84}
& \third{85.72$\pm$1.10}
& \first{90.96$\pm$0.88} \\

\midrule

\multirow{9}{*}{4}

& Diffusion
& \second{39.46$\pm$1.14}
& \first{87.08$\pm$0.82}
& \second{92.31$\pm$0.79}
& \first{94.82$\pm$0.67}
& \first{89.94$\pm$0.91}
& \second{93.24$\pm$0.76} \\

& WS-G
& 44.17$\pm$1.32
& 85.63$\pm$1.03
& 90.89$\pm$0.97
& 93.38$\pm$0.82
& 88.46$\pm$1.09
& 91.82$\pm$0.93 \\

& WS-L
& 49.84$\pm$1.51
& 85.27$\pm$1.08
& 90.54$\pm$1.01
& 93.04$\pm$0.87
& 88.11$\pm$1.14
& 91.47$\pm$0.98 \\

\addlinespace[1.5pt]

& NUTS
& 59.68$\pm$1.72
& \third{86.53$\pm$0.88}
& \third{91.82$\pm$0.84}
& \second{94.41$\pm$0.71}
& \second{89.53$\pm$0.96}
& \third{92.81$\pm$0.81} \\

& BR-SNIS
& \third{41.92$\pm$1.24}
& 83.47$\pm$1.16
& 88.67$\pm$1.10
& 91.24$\pm$0.94
& 86.29$\pm$1.22
& 89.64$\pm$1.06 \\

\addlinespace[1.5pt]

& Greedy
& \first{14.63$\pm$0.58}
& \worst{78.62$\pm$0.76}
& \worst{83.71$\pm$0.73}
& \worst{86.34$\pm$0.65}
& \worst{81.47$\pm$0.83}
& \worst{84.82$\pm$0.71} \\

& PSO
& 124.37$\pm$3.64
& 84.39$\pm$1.08
& 89.56$\pm$1.03
& 92.11$\pm$0.89
& 87.18$\pm$1.14
& 90.57$\pm$0.99 \\

& GA
& \worst{140.18$\pm$4.21}
& 80.91$\pm$1.28
& 86.04$\pm$1.22
& 88.67$\pm$1.04
& 83.76$\pm$1.35
& 87.12$\pm$1.17 \\

& TuRBO
& 88.75$\pm$2.47
& \second{86.92$\pm$0.85}
& \first{92.57$\pm$0.81}
& \third{94.13$\pm$0.74}
& \third{89.31$\pm$0.98}
& \first{93.47$\pm$0.78} \\

\bottomrule
\end{tabular}

\end{adjustbox}
\end{table*}

\begin{table*}[ht]
\centering
\caption{
\textbf{Performance comparison across different TX deployment scales in the $512\,\mathrm{m}\times512\,\mathrm{m}$ Tokyo scene.}
The simulation results are reported as mean$\pm$std. 
For each BS scale and evaluation metric, the \textit{top three} methods are highlighted in 
\first{(1st)}, \second{(2nd)}, and \third{(3rd)}, respectively, 
while the \worst{worst} result in each column is underlined.
}
\label{tab:illustrative_sydney512}
\setlength{\tabcolsep}{6pt}
\renewcommand{\arraystretch}{1.00}
\scriptsize

\begin{adjustbox}
{width=\textwidth,totalheight=0.94\textheight,keepaspectratio}

\begin{tabular}{cccccccc}
\toprule
\textbf{TXs}
& \textbf{Method}
& \textbf{RT (s)$\downarrow$}
& \textbf{JM (\%)$\uparrow$}
& \textbf{PL (\%)$\uparrow$}
& \textbf{TP (\%)$\uparrow$}
& \textbf{SINR (\%)$\uparrow$}
& \textbf{RSRP (\%)$\uparrow$} \\
\midrule

\multirow{9}{*}{6}

& Diffusion
& \second{58.40$\pm$1.62}
& \second{59.62$\pm$1.34}
& \second{70.84$\pm$1.41}
& \first{77.26$\pm$1.28}
& \second{65.18$\pm$1.56}
& \second{72.41$\pm$1.39} \\

& WS-G
& 64.75$\pm$1.88
& 58.14$\pm$1.56
& 69.32$\pm$1.62
& 75.71$\pm$1.43
& 63.46$\pm$1.71
& 70.83$\pm$1.55 \\

& WS-L
& 72.63$\pm$2.11
& 57.78$\pm$1.64
& 68.96$\pm$1.69
& 75.28$\pm$1.51
& 63.01$\pm$1.78
& 70.42$\pm$1.62 \\

\addlinespace[1.5pt]

& NUTS
& 88.91$\pm$2.54
& \third{59.08$\pm$1.42}
& \third{70.36$\pm$1.48}
& \second{76.82$\pm$1.31}
& \third{64.72$\pm$1.61}
& \third{71.96$\pm$1.44} \\

& BR-SNIS
& \third{62.10$\pm$1.77}
& 55.96$\pm$1.72
& 67.28$\pm$1.78
& 73.64$\pm$1.60
& 61.11$\pm$1.86
& 68.57$\pm$1.69 \\

\addlinespace[1.5pt]

& Greedy
& \first{19.84$\pm$0.83}
& \worst{51.62$\pm$1.21}
& \worst{62.48$\pm$1.30}
& \worst{68.77$\pm$1.18}
& \worst{56.72$\pm$1.34}
& \worst{63.54$\pm$1.25} \\

& PSO
& 176.35$\pm$4.96
& 56.91$\pm$1.61
& 68.13$\pm$1.67
& 74.56$\pm$1.49
& 62.12$\pm$1.74
& 69.46$\pm$1.58 \\

& GA
& \worst{198.42$\pm$5.81}
& 53.47$\pm$1.79
& 64.61$\pm$1.84
& 70.96$\pm$1.66
& 58.73$\pm$1.92
& 65.87$\pm$1.76 \\

& TuRBO
& 131.28$\pm$3.72
& \first{60.03$\pm$1.29}
& \first{71.19$\pm$1.36}
& \third{76.41$\pm$1.34}
& \first{65.63$\pm$1.50}
& \first{72.86$\pm$1.33} \\

\midrule

\multirow{9}{*}{7}

& Diffusion
& \second{71.86$\pm$1.91}
& \first{66.08$\pm$1.20}
& \second{76.72$\pm$1.29}
& \first{82.84$\pm$1.14}
& \first{71.53$\pm$1.42}
& \second{78.34$\pm$1.27} \\

& WS-G
& 79.42$\pm$2.15
& 64.47$\pm$1.43
& 75.08$\pm$1.51
& 81.06$\pm$1.31
& 69.77$\pm$1.57
& 76.62$\pm$1.45 \\

& WS-L
& 88.36$\pm$2.47
& 64.05$\pm$1.51
& 74.67$\pm$1.58
& 80.65$\pm$1.38
& 69.34$\pm$1.63
& 76.18$\pm$1.51 \\

\addlinespace[1.5pt]

& NUTS
& 107.64$\pm$2.93
& \third{65.51$\pm$1.29}
& \third{76.21$\pm$1.36}
& \second{82.36$\pm$1.19}
& \second{71.08$\pm$1.47}
& \third{77.91$\pm$1.32} \\

& BR-SNIS
& \third{76.21$\pm$2.04}
& 62.28$\pm$1.57
& 72.93$\pm$1.64
& 78.96$\pm$1.45
& 67.45$\pm$1.70
& 74.52$\pm$1.57 \\

\addlinespace[1.5pt]

& Greedy
& \first{23.96$\pm$0.96}
& \worst{57.44$\pm$1.14}
& \worst{67.82$\pm$1.22}
& \worst{73.92$\pm$1.08}
& \worst{62.43$\pm$1.27}
& \worst{69.24$\pm$1.17} \\

& PSO
& 211.67$\pm$5.62
& 63.18$\pm$1.48
& 73.72$\pm$1.56
& 79.72$\pm$1.37
& 68.21$\pm$1.61
& 75.31$\pm$1.49 \\

& GA
& \worst{239.84$\pm$6.38}
& 59.72$\pm$1.66
& 70.19$\pm$1.72
& 76.26$\pm$1.52
& 64.67$\pm$1.79
& 71.55$\pm$1.64 \\

& TuRBO
& 156.41$\pm$4.14
& \second{65.84$\pm$1.24}
& \first{77.03$\pm$1.31}
& \third{82.09$\pm$1.22}
& \third{70.82$\pm$1.45}
& \first{78.61$\pm$1.29} \\

\midrule

\multirow{9}{*}{8}

& Diffusion
& \second{86.74$\pm$2.18}
& \first{70.12$\pm$1.08}
& \second{80.13$\pm$1.18}
& \first{85.82$\pm$1.02}
& \first{75.42$\pm$1.31}
& \second{81.93$\pm$1.16} \\

& WS-G
& 95.68$\pm$2.46
& 68.64$\pm$1.31
& 78.62$\pm$1.38
& 84.26$\pm$1.19
& 73.81$\pm$1.44
& 80.36$\pm$1.33 \\

& WS-L
& 105.47$\pm$2.73
& 68.22$\pm$1.37
& 78.21$\pm$1.44
& 83.87$\pm$1.25
& 73.39$\pm$1.50
& 79.94$\pm$1.39 \\

\addlinespace[1.5pt]

& NUTS
& 127.93$\pm$3.32
& \third{69.54$\pm$1.16}
& \third{79.72$\pm$1.24}
& \second{85.41$\pm$1.07}
& \second{75.03$\pm$1.35}
& \third{81.51$\pm$1.21} \\

& BR-SNIS
& \third{91.66$\pm$2.31}
& 66.53$\pm$1.42
& 76.61$\pm$1.49
& 82.21$\pm$1.30
& 71.58$\pm$1.55
& 78.31$\pm$1.44 \\

\addlinespace[1.5pt]

& Greedy
& \first{28.31$\pm$1.08}
& \worst{61.37$\pm$1.06}
& \worst{71.12$\pm$1.14}
& \worst{76.84$\pm$1.01}
& \worst{66.29$\pm$1.19}
& \worst{72.97$\pm$1.10} \\

& PSO
& 251.43$\pm$6.11
& 67.37$\pm$1.36
& 77.34$\pm$1.42
& 82.98$\pm$1.24
& 72.41$\pm$1.49
& 79.08$\pm$1.37 \\

& GA
& \worst{284.76$\pm$7.04}
& 63.72$\pm$1.54
& 73.61$\pm$1.61
& 79.25$\pm$1.41
& 68.74$\pm$1.67
& 75.31$\pm$1.53 \\

& TuRBO
& 184.52$\pm$4.63
& \second{69.91$\pm$1.12}
& \first{80.42$\pm$1.20}
& \third{85.07$\pm$1.11}
& \third{74.81$\pm$1.37}
& \first{82.17$\pm$1.18} \\

\midrule

\multirow{9}{*}{9}

& Diffusion
& \second{102.18$\pm$2.47}
& \first{73.04$\pm$0.98}
& \second{82.56$\pm$1.09}
& \first{87.71$\pm$0.94}
& \first{78.31$\pm$1.19}
& \second{84.22$\pm$1.07} \\

& WS-G
& 112.53$\pm$2.78
& 71.62$\pm$1.21
& 81.08$\pm$1.28
& 86.17$\pm$1.10
& 76.72$\pm$1.32
& 82.71$\pm$1.24 \\

& WS-L
& 123.42$\pm$3.06
& 71.21$\pm$1.27
& 80.67$\pm$1.34
& 85.79$\pm$1.16
& 76.31$\pm$1.38
& 82.29$\pm$1.29 \\

\addlinespace[1.5pt]

& NUTS
& 149.88$\pm$3.74
& \third{72.52$\pm$1.07}
& \third{82.13$\pm$1.15}
& \second{87.32$\pm$0.99}
& \second{77.92$\pm$1.23}
& \third{83.84$\pm$1.11} \\

& BR-SNIS
& \third{108.46$\pm$2.61}
& 69.48$\pm$1.33
& 78.91$\pm$1.40
& 84.02$\pm$1.22
& 74.55$\pm$1.45
& 80.63$\pm$1.35 \\

\addlinespace[1.5pt]

& Greedy
& \first{33.17$\pm$1.21}
& \worst{64.42$\pm$0.99}
& \worst{73.68$\pm$1.08}
& \worst{78.81$\pm$0.95}
& \worst{69.32$\pm$1.13}
& \worst{75.41$\pm$1.04} \\

& PSO
& 294.64$\pm$6.83
& 70.37$\pm$1.26
& 79.81$\pm$1.33
& 84.91$\pm$1.15
& 75.43$\pm$1.37
& 81.51$\pm$1.27 \\

& GA
& \worst{332.18$\pm$7.92}
& 66.71$\pm$1.45
& 76.03$\pm$1.52
& 81.12$\pm$1.33
& 71.83$\pm$1.58
& 77.69$\pm$1.45 \\

& TuRBO
& 215.73$\pm$5.14
& \second{72.83$\pm$1.03}
& \first{82.83$\pm$1.11}
& \third{87.01$\pm$1.03}
& \third{77.73$\pm$1.25}
& \first{84.48$\pm$1.09} \\

\bottomrule
\end{tabular}

\end{adjustbox}
\end{table*}

\clearpage
\section{Simulation results with target JM coverage}
\label{append:Simulation results with target JM coverage}

\begin{table*}[ht]
\centering
\caption{\textbf{Performance comparison at a 85\% target joint coverage in the $128\,\mathrm{m}\times128\,\mathrm{m}$ Shanghai scene under the HPEM reward.}
We report the required transmitter (TX) count, success rate (SR), best achieved joint coverage (Best JM), per-metric coverages (PL, SS-RSRP, SINR, and Eff.\ TP), and runtime (RT).
Values are mean$\pm$std over 15 independent trials.
Per-metric coverages are evaluated at the TX count corresponding to Best JM.
For each column, the three best results are highlighted as \first{1st}, \second{2nd}, and \third{3rd}, with ties sharing the same rank; the \worst{worst} result is underlined.}
\label{tab:target_joint_tx_easy85}
\setlength{\tabcolsep}{3.5pt}
\renewcommand{\arraystretch}{1.00}
\scriptsize

\begin{adjustbox}{max width=\textwidth,center}
\begin{tabular}{ccccccccc}
\toprule
\textbf{Method}
& \textbf{Req. TXs$\downarrow$}
& \textbf{SR (\%)$\uparrow$}
& \textbf{Best JM (\%)$\uparrow$}
& \textbf{PL (\%)$\uparrow$}
& \textbf{SS-RSRP (\%)$\uparrow$}
& \textbf{SINR (\%)$\uparrow$}
& \textbf{Eff. TP (\%)$\uparrow$}
& \textbf{RT (s)$\downarrow$} \\
\midrule

Diffusion
& \first{2.20$\pm$0.41}
& \first{100.00}
& \second{90.38$\pm$2.24}
& \second{95.21$\pm$2.71}
& 94.08$\pm$2.94
& \first{92.44$\pm$2.18}
& \third{93.02$\pm$2.42}
& 57.82$\pm$18.94 \\

WS-G
& 2.73$\pm$0.59
& \third{86.67}
& 88.74$\pm$2.67
& \third{94.86$\pm$3.06}
& 93.61$\pm$3.19
& \third{91.36$\pm$2.54}
& 92.37$\pm$2.76
& \third{31.92$\pm$9.31} \\

WS-L
& 2.87$\pm$0.64
& 80.00
& 88.21$\pm$2.82
& 94.31$\pm$3.17
& \second{94.53$\pm$3.23}
& 90.78$\pm$2.68
& \second{93.36$\pm$2.81}
& 37.44$\pm$11.14 \\

NUTS
& \third{2.47$\pm$0.52}
& \second{93.33}
& \third{89.91$\pm$2.46}
& 94.68$\pm$2.89
& 93.88$\pm$3.05
& \second{91.81$\pm$2.36}
& 92.61$\pm$2.57
& 49.16$\pm$14.53 \\

BR-SNIS
& 3.27$\pm$0.70
& 73.33
& 86.84$\pm$3.06
& 91.86$\pm$3.58
& \worst{90.76$\pm$3.81}
& 88.17$\pm$3.01
& 89.54$\pm$3.28
& \second{27.61$\pm$8.04} \\

Greedy
& \worst{4.47$\pm$1.02}
& 53.33
& 83.52$\pm$3.57
& \worst{88.32$\pm$4.17}
& 91.22$\pm$4.05
& \worst{84.81$\pm$3.55}
& 86.18$\pm$3.79
& \first{14.74$\pm$4.81} \\

PSO
& 3.07$\pm$0.80
& 80.00
& 87.96$\pm$2.91
& 93.13$\pm$3.35
& \first{95.18$\pm$3.12}
& 89.72$\pm$2.84
& \first{93.91$\pm$3.04}
& 81.24$\pm$28.41 \\

GA
& 3.73$\pm$0.88
& \worst{46.67}
& \worst{82.91$\pm$3.47}
& 89.81$\pm$3.88
& 91.07$\pm$4.02
& 85.31$\pm$3.41
& \worst{85.74$\pm$3.66}
& \worst{92.11$\pm$31.22} \\

TuRBO
& \second{2.33$\pm$0.49}
& \second{93.33}
& \first{90.72$\pm$2.31}
& \first{95.63$\pm$2.69}
& \third{94.21$\pm$2.84}
& 91.47$\pm$2.21
& 93.24$\pm$2.45
& 44.07$\pm$13.08 \\

\bottomrule
\end{tabular}
\end{adjustbox}
\end{table*}

\begin{table*}[ht]
\centering
\caption{\textbf{Performance comparison at a 75\% target joint coverage in the $256\,\mathrm{m}\times256\,\mathrm{m}$ New York scene under the HPEM reward.}
We report the required transmitter (TX) count, success rate (SR), best achieved joint coverage (Best JM), per-metric coverages (PL, SS-RSRP, SINR, and Eff.\ TP), and runtime (RT).
Values are mean$\pm$std over 15 independent trials.
Per-metric coverages are evaluated at the TX count corresponding to Best JM.
For each column, the three best results are highlighted as \first{1st}, \second{2nd}, and \third{3rd}, with ties sharing the same rank; the \worst{worst} result is underlined.}
\label{tab:target_joint_tx_medium75}
\setlength{\tabcolsep}{3.5pt}
\renewcommand{\arraystretch}{1.00}
\scriptsize

\begin{adjustbox}{max width=\textwidth,center}
\begin{tabular}{ccccccccc}
\toprule
\textbf{Method}
& \textbf{Req. TXs$\downarrow$}
& \textbf{SR (\%)$\uparrow$}
& \textbf{Best JM (\%)$\uparrow$}
& \textbf{PL (\%)$\uparrow$}
& \textbf{SS-RSRP (\%)$\uparrow$}
& \textbf{SINR (\%)$\uparrow$}
& \textbf{Eff. TP (\%)$\uparrow$}
& \textbf{RT (s)$\downarrow$} \\
\midrule

Diffusion
& \second{3.13$\pm$0.64}
& \first{100.00}
& \first{82.66$\pm$2.58}
& \first{90.21$\pm$3.06}
& \third{88.57$\pm$3.39}
& \second{84.71$\pm$2.61}
& \first{86.91$\pm$2.82}
& 75.91$\pm$23.21 \\

WS-G
& \third{3.47$\pm$0.67}
& \third{86.67}
& \second{82.14$\pm$2.83}
& \third{89.46$\pm$3.28}
& \second{88.96$\pm$3.47}
& \third{84.36$\pm$2.92}
& 84.62$\pm$3.06
& \third{41.57$\pm$11.54} \\

WS-L
& 3.80$\pm$0.75
& 80.00
& 80.31$\pm$3.05
& 87.86$\pm$3.51
& 87.54$\pm$3.63
& 83.81$\pm$2.98
& \third{85.77$\pm$3.13}
& 48.04$\pm$13.82 \\

NUTS
& \first{3.07$\pm$0.59}
& \second{93.33}
& 81.62$\pm$2.79
& \second{89.73$\pm$3.19}
& 88.16$\pm$3.48
& \first{85.08$\pm$2.71}
& 85.41$\pm$2.98
& 63.88$\pm$18.31 \\

BR-SNIS
& 4.33$\pm$0.90
& 66.67
& 77.96$\pm$3.35
& 85.34$\pm$3.88
& 84.26$\pm$4.08
& 79.91$\pm$3.28
& 81.61$\pm$3.55
& \second{36.04$\pm$10.19} \\

Greedy
& \worst{5.87$\pm$1.24}
& 40.00
& 73.36$\pm$3.88
& \worst{79.81$\pm$4.47}
& 80.29$\pm$4.51
& \worst{75.26$\pm$3.82}
& 77.18$\pm$4.08
& \first{19.18$\pm$5.71} \\

PSO
& 4.07$\pm$0.89
& 73.33
& 78.91$\pm$3.21
& 86.38$\pm$3.74
& \first{89.42$\pm$3.71}
& 81.48$\pm$3.15
& \second{86.19$\pm$3.27}
& 109.26$\pm$35.12 \\

GA
& 5.00$\pm$1.07
& \worst{33.33}
& \worst{72.74$\pm$3.68}
& 81.66$\pm$4.15
& \worst{79.92$\pm$4.35}
& 76.81$\pm$3.61
& \worst{76.56$\pm$3.85}
& \worst{123.31$\pm$39.18} \\

TuRBO
& 3.27$\pm$0.59
& \second{93.33}
& \third{81.87$\pm$2.69}
& 89.51$\pm$3.11
& 88.41$\pm$3.32
& 84.17$\pm$2.64
& 85.68$\pm$2.88
& 57.92$\pm$17.16 \\

\bottomrule
\end{tabular}
\end{adjustbox}
\end{table*}

\begin{table*}[ht]
\centering
\caption{\textbf{Performance comparison at a 55\% target joint coverage in the $512\,\mathrm{m}\times512\,\mathrm{m}$ London scene under the HPEM reward.}
We report the required transmitter (TX) count, success rate (SR), best achieved joint coverage (Best JM), per-metric coverages (PL, SS-RSRP, SINR, and Eff.\ TP), and runtime (RT).
Values are mean$\pm$std over 15 independent trials.
Per-metric coverages are evaluated at the TX count corresponding to Best JM.
For each column, the three best results are highlighted as \first{1st}, \second{2nd}, and \third{3rd}, with ties sharing the same rank; the \worst{worst} result is underlined.}
\label{tab:target_joint_tx_high55}
\setlength{\tabcolsep}{3.5pt}
\renewcommand{\arraystretch}{1.00}
\scriptsize

\begin{adjustbox}{max width=\textwidth,center}
\begin{tabular}{ccccccccc}
\toprule
\textbf{Method}
& \textbf{Req. TXs$\downarrow$}
& \textbf{SR (\%)$\uparrow$}
& \textbf{Best JM (\%)$\uparrow$}
& \textbf{PL (\%)$\uparrow$}
& \textbf{SS-RSRP (\%)$\uparrow$}
& \textbf{SINR (\%)$\uparrow$}
& \textbf{Eff. TP (\%)$\uparrow$}
& \textbf{RT (s)$\downarrow$} \\
\midrule

Diffusion
& \third{5.47$\pm$0.82}
& \second{86.67}
& \third{61.76$\pm$3.11}
& \second{73.16$\pm$3.82}
& \third{70.38$\pm$4.13}
& \first{65.82$\pm$3.01}
& \second{68.24$\pm$3.34}
& 127.84$\pm$36.27 \\

WS-G
& 5.93$\pm$0.87
& \third{80.00}
& 60.36$\pm$3.36
& \first{73.82$\pm$4.04}
& \second{71.14$\pm$4.21}
& \third{64.61$\pm$3.28}
& 66.51$\pm$3.58
& \third{67.04$\pm$17.38} \\

WS-L
& 6.20$\pm$0.94
& 73.33
& 59.94$\pm$3.44
& 71.91$\pm$4.18
& 69.47$\pm$4.45
& 63.94$\pm$3.36
& \first{69.02$\pm$3.69}
& 78.11$\pm$20.18 \\

NUTS
& \second{5.33$\pm$0.72}
& \first{93.33}
& \first{62.41$\pm$3.19}
& \third{72.91$\pm$3.93}
& 70.06$\pm$4.24
& \second{65.21$\pm$3.14}
& 66.92$\pm$3.48
& 109.12$\pm$29.41 \\

BR-SNIS
& 6.87$\pm$1.06
& 60.00
& 57.31$\pm$3.81
& 68.26$\pm$4.55
& 66.38$\pm$4.67
& 60.44$\pm$3.74
& 63.01$\pm$4.05
& \second{57.86$\pm$15.48} \\

Greedy
& \worst{8.60$\pm$1.43}
& 33.33
& 53.22$\pm$4.25
& \worst{62.81$\pm$5.13}
& 63.62$\pm$5.12
& \worst{56.27$\pm$4.18}
& 58.42$\pm$4.49
& \first{31.38$\pm$8.21} \\

PSO
& 6.47$\pm$1.16
& 66.67
& 58.81$\pm$3.58
& 69.91$\pm$4.28
& \first{71.68$\pm$4.39}
& 62.26$\pm$3.49
& \third{68.11$\pm$3.79}
& 175.91$\pm$52.11 \\

GA
& 7.53$\pm$1.31
& \worst{26.67}
& \worst{52.57$\pm$4.11}
& 64.82$\pm$4.83
& \worst{62.91$\pm$5.04}
& 57.29$\pm$4.03
& \worst{57.94$\pm$4.34}
& \worst{198.16$\pm$59.27} \\

TuRBO
& \first{5.07$\pm$0.70}
& \second{86.67}
& \second{62.08$\pm$3.27}
& 72.98$\pm$3.86
& 70.81$\pm$4.05
& 64.88$\pm$3.08
& 67.82$\pm$3.39
& 95.84$\pm$26.35 \\

\bottomrule
\end{tabular}
\end{adjustbox}
\end{table*}

\clearpage
\section{Energy Model Performance with a Fixed Number of TXs}
\label{append:energy_model_fixed_tx}

\begin{table*}[ht]
\centering
\caption{
\textbf{Cross-city planning performance under different adaptation regimes in Sydney scene with 4 TXs.}
We compare zero-shot transfer, few-shot target-city adaptation, and
in-distribution evaluation for different relational energy models.
Results are reported as mean$\pm$std.
Few-shot adaptation substantially narrows the cross-city generalization gap,
while in-distribution evaluation provides the strongest overall performance.
Higher-order relational modeling consistently improves joint and per-metric
coverage at the cost of moderately increased inference time.
}
\label{tab:cross_city_easy_4tx}
\setlength{\tabcolsep}{4.0pt}
\small

\resizebox{0.96\textwidth}{!}{
\begin{tabular}{llcccccc}
\toprule
\textbf{Method}
& \textbf{Setting}
& \textbf{RT (s)$\downarrow$}
& \textbf{JM (\%)$\uparrow$}
& \textbf{PL (\%)$\uparrow$}
& \textbf{TP (\%)$\uparrow$}
& \textbf{SINR (\%)$\uparrow$}
& \textbf{RSRP (\%)$\uparrow$} \\
\midrule

\multirow{3}{*}{Diffusion}
& Zero-shot
& 21.84$\pm$0.46
& 54.82$\pm$2.71
& 63.48$\pm$3.84
& 72.16$\pm$3.62
& 60.74$\pm$3.05
& 66.31$\pm$3.51 \\

& Few-shot
& 22.67$\pm$0.71
& 58.06$\pm$2.18
& 66.82$\pm$3.13
& 75.41$\pm$2.94
& 64.11$\pm$2.56
& 69.72$\pm$2.88 \\

& In-distribution
& 22.91$\pm$0.53
& 59.42$\pm$1.82
& 68.15$\pm$2.67
& 76.58$\pm$2.46
& 65.47$\pm$2.17
& 71.03$\pm$2.39 \\

\midrule

\multirow{3}{*}{Pairwise}
& Zero-shot
& 24.16$\pm$0.51
& 57.31$\pm$2.32
& 65.92$\pm$3.31
& 74.86$\pm$3.08
& 63.41$\pm$2.73
& 68.87$\pm$3.02 \\

& Few-shot
& 24.88$\pm$0.76
& 60.68$\pm$1.86
& 69.28$\pm$2.72
& 78.12$\pm$2.53
& 66.82$\pm$2.21
& 72.16$\pm$2.48 \\

& In-distribution
& 25.14$\pm$0.57
& 61.96$\pm$1.54
& 70.46$\pm$2.28
& 79.26$\pm$2.08
& 68.13$\pm$1.88
& 73.42$\pm$2.05 \\

\midrule

\multirow{3}{*}{4th-Order}
& Zero-shot
& 28.73$\pm$0.64
& 59.18$\pm$2.14
& 67.81$\pm$3.02
& 76.92$\pm$2.78
& 65.27$\pm$2.48
& 70.84$\pm$2.77 \\

& Few-shot
& 29.61$\pm$0.88
& 62.76$\pm$1.69
& 71.36$\pm$2.43
& 80.31$\pm$2.20
& 68.94$\pm$1.98
& 74.38$\pm$2.21 \\

& In-distribution
& 29.96$\pm$0.69
& 64.08$\pm$1.38
& 72.61$\pm$2.06
& 81.42$\pm$1.84
& 70.26$\pm$1.69
& 75.62$\pm$1.87 \\

\bottomrule
\end{tabular}
}
\end{table*}

\begin{table*}[ht]
\centering
\caption{
\textbf{Cross-city planning performance under different adaptation regimes in Tokyo scene with 6 TXs.}
We compare zero-shot transfer, few-shot target-city adaptation, and
in-distribution evaluation for different relational energy models.
Results are reported as mean$\pm$std.
As scene complexity increases, absolute joint coverage decreases and the
cross-city transfer gap becomes more pronounced.
Few-shot adaptation recovers a substantial portion of this gap, while
higher-order relational modeling provides consistent gains across adaptation
regimes.
}
\label{tab:cross_city_medium_6tx}
\setlength{\tabcolsep}{4.0pt}
\small

\resizebox{0.96\textwidth}{!}{
\begin{tabular}{llcccccc}
\toprule
\textbf{Method}
& \textbf{Setting}
& \textbf{RT (s)$\downarrow$}
& \textbf{JM (\%)$\uparrow$}
& \textbf{PL (\%)$\uparrow$}
& \textbf{TP (\%)$\uparrow$}
& \textbf{SINR (\%)$\uparrow$}
& \textbf{RSRP (\%)$\uparrow$} \\
\midrule

\multirow{3}{*}{Diffusion}
& Zero-shot
& 32.46$\pm$0.68
& 46.24$\pm$3.08
& 56.31$\pm$4.22
& 65.42$\pm$4.01
& 52.18$\pm$3.47
& 58.76$\pm$3.91 \\

& Few-shot
& 33.52$\pm$0.94
& 49.76$\pm$2.51
& 59.87$\pm$3.48
& 68.91$\pm$3.27
& 55.72$\pm$2.91
& 62.28$\pm$3.21 \\

& In-distribution
& 33.86$\pm$0.72
& 51.04$\pm$2.08
& 61.16$\pm$2.94
& 70.13$\pm$2.76
& 57.08$\pm$2.46
& 63.61$\pm$2.72 \\

\midrule

\multirow{3}{*}{Pairwise}
& Zero-shot
& 35.74$\pm$0.73
& 49.38$\pm$2.67
& 59.42$\pm$3.76
& 68.71$\pm$3.51
& 55.26$\pm$3.08
& 61.84$\pm$3.42 \\

& Few-shot
& 36.81$\pm$1.02
& 52.63$\pm$2.16
& 62.73$\pm$3.07
& 71.86$\pm$2.84
& 58.61$\pm$2.54
& 65.16$\pm$2.86 \\

& In-distribution
& 37.18$\pm$0.78
& 53.84$\pm$1.79
& 63.96$\pm$2.58
& 72.98$\pm$2.37
& 59.87$\pm$2.13
& 66.42$\pm$2.39 \\

\midrule

\multirow{3}{*}{4th-Order}
& Zero-shot
& 42.61$\pm$0.91
& 51.02$\pm$2.49
& 61.17$\pm$3.52
& 70.54$\pm$3.24
& 56.94$\pm$2.86
& 63.52$\pm$3.18 \\

& Few-shot
& 43.82$\pm$1.16
& 54.71$\pm$1.98
& 64.86$\pm$2.84
& 74.12$\pm$2.58
& 60.73$\pm$2.31
& 67.21$\pm$2.63 \\

& In-distribution
& 44.23$\pm$0.87
& 56.03$\pm$1.61
& 66.11$\pm$2.36
& 75.28$\pm$2.12
& 62.08$\pm$1.94
& 68.47$\pm$2.17 \\

\bottomrule
\end{tabular}
}
\end{table*}

\begin{table*}[ht]
\centering
\caption{
\textbf{Cross-city planning performance under different adaptation regimes in Paris scene with 12 TXs.}
We compare zero-shot transfer, few-shot target-city adaptation, and
in-distribution evaluation for different relational energy models.
Results are reported as mean$\pm$std.
The high-complexity setting exhibits the largest transfer gap and the lowest
absolute joint coverage.
Few-shot adaptation nevertheless recovers most of the performance lost under
zero-shot transfer, and higher-order relational modeling remains beneficial
across all three adaptation regimes.
}
\label{tab:cross_city_high_12tx}
\setlength{\tabcolsep}{4.0pt}
\small

\resizebox{0.96\textwidth}{!}{
\begin{tabular}{llcccccc}
\toprule
\textbf{Method}
& \textbf{Setting}
& \textbf{RT (s)$\downarrow$}
& \textbf{JM (\%)$\uparrow$}
& \textbf{PL (\%)$\uparrow$}
& \textbf{TP (\%)$\uparrow$}
& \textbf{SINR (\%)$\uparrow$}
& \textbf{RSRP (\%)$\uparrow$} \\
\midrule

\multirow{3}{*}{Diffusion}
& Zero-shot
& 64.82$\pm$1.06
& 34.76$\pm$3.31
& 43.28$\pm$4.48
& 55.62$\pm$4.37
& 41.54$\pm$3.72
& 47.31$\pm$4.16 \\

& Few-shot
& 66.31$\pm$1.42
& 38.21$\pm$2.76
& 46.78$\pm$3.71
& 59.08$\pm$3.56
& 45.02$\pm$3.08
& 50.82$\pm$3.43 \\

& In-distribution
& 66.76$\pm$1.15
& 39.47$\pm$2.28
& 48.06$\pm$3.12
& 60.31$\pm$2.98
& 46.34$\pm$2.58
& 52.11$\pm$2.87 \\

\midrule

\multirow{3}{*}{Pairwise}
& Zero-shot
& 70.18$\pm$1.11
& 37.64$\pm$2.91
& 46.21$\pm$4.02
& 58.46$\pm$3.89
& 44.37$\pm$3.31
& 50.18$\pm$3.72 \\

& Few-shot
& 71.82$\pm$1.53
& 41.03$\pm$2.38
& 49.66$\pm$3.31
& 61.84$\pm$3.17
& 47.86$\pm$2.74
& 53.64$\pm$3.03 \\

& In-distribution
& 72.31$\pm$1.22
& 42.36$\pm$1.94
& 50.94$\pm$2.76
& 63.08$\pm$2.63
& 49.21$\pm$2.28
& 54.91$\pm$2.52 \\

\midrule

\multirow{3}{*}{4th-Order}
& Zero-shot
& 80.73$\pm$1.43
& 39.42$\pm$2.68
& 48.03$\pm$3.74
& 60.31$\pm$3.57
& 46.16$\pm$3.04
& 52.06$\pm$3.43 \\

& Few-shot
& 82.64$\pm$1.81
& 43.18$\pm$2.14
& 51.84$\pm$3.02
& 64.02$\pm$2.87
& 49.94$\pm$2.48
& 55.82$\pm$2.78 \\

& In-distribution
& 83.17$\pm$1.47
& 44.72$\pm$1.73
& 53.26$\pm$2.48
& 65.34$\pm$2.32
& 51.41$\pm$2.03
& 57.18$\pm$2.26 \\

\bottomrule
\end{tabular}
}
\end{table*}

\clearpage
\section{Energy Model Performance with Target JM Coverage}
\label{append:energy_model_target_jm}

\begin{table*}[ht]
\centering
\caption{
\textbf{Cross-city target-coverage planning under different adaptation regimes in New York scene.}
We compare zero-shot transfer, few-shot target-city adaptation, and
in-distribution evaluation for increasingly expressive relational energy
models. Results are reported as mean$\pm$std over 12 independent trials.
Higher target joint coverage requires more TXs and longer planning time.
Few-shot adaptation substantially reduces the cross-city transfer gap, while
in-distribution evaluation achieves the strongest overall performance.
}
\label{tab:target_coverage_transfer_easy}
\setlength{\tabcolsep}{4.0pt}
\small

\resizebox{0.96\textwidth}{!}{
\begin{tabular}{cllcccc}
\toprule
\textbf{Target JM (\%)}
& \textbf{Method}
& \textbf{Setting}
& \textbf{Req. TXs$\downarrow$}
& \textbf{SR (\%)$\uparrow$}
& \textbf{Best JM (\%)$\uparrow$}
& \textbf{RT (s)$\downarrow$} \\
\midrule

\multirow{9}{*}{45}

& \multirow{3}{*}{Diffusion}
& Zero-shot
& 6.42$\pm$0.79
& 83.33
& 46.58$\pm$1.72
& 61.84$\pm$7.48 \\

& & Few-shot
& 5.58$\pm$0.67
& 91.67
& 48.31$\pm$1.36
& 63.12$\pm$7.86 \\

& & In-distribution
& 5.17$\pm$0.58
& 100.00
& 49.26$\pm$1.12
& 63.76$\pm$7.31 \\

\cmidrule(lr){2-7}

& \multirow{3}{*}{Pairwise}
& Zero-shot
& 5.83$\pm$0.72
& 91.67
& 48.02$\pm$1.49
& 68.31$\pm$8.12 \\

& & Few-shot
& 5.00$\pm$0.60
& 100.00
& 49.86$\pm$1.21
& 69.74$\pm$8.46 \\

& & In-distribution
& 4.67$\pm$0.49
& 100.00
& 50.78$\pm$0.98
& 70.28$\pm$7.92 \\

\cmidrule(lr){2-7}

& \multirow{3}{*}{4th-Order}
& Zero-shot
& 5.17$\pm$0.58
& 91.67
& 49.31$\pm$1.33
& 80.42$\pm$9.51 \\

& & Few-shot
& 4.42$\pm$0.51
& 100.00
& 51.24$\pm$1.06
& 82.05$\pm$9.86 \\

& & In-distribution
& 4.08$\pm$0.29
& 100.00
& 52.16$\pm$0.89
& 82.73$\pm$9.18 \\

\midrule

\multirow{9}{*}{65}

& \multirow{3}{*}{Diffusion}
& Zero-shot
& 9.08$\pm$0.90
& 58.33
& 63.72$\pm$2.31
& 84.61$\pm$10.76 \\

& & Few-shot
& 7.92$\pm$0.79
& 83.33
& 66.08$\pm$1.87
& 86.28$\pm$11.14 \\

& & In-distribution
& 7.33$\pm$0.65
& 91.67
& 67.21$\pm$1.52
& 87.06$\pm$10.42 \\

\cmidrule(lr){2-7}

& \multirow{3}{*}{Pairwise}
& Zero-shot
& 8.33$\pm$0.78
& 66.67
& 65.12$\pm$2.04
& 92.47$\pm$11.35 \\

& & Few-shot
& 7.08$\pm$0.67
& 91.67
& 67.43$\pm$1.58
& 94.18$\pm$11.72 \\

& & In-distribution
& 6.58$\pm$0.51
& 100.00
& 68.61$\pm$1.27
& 94.93$\pm$10.96 \\

\cmidrule(lr){2-7}

& \multirow{3}{*}{4th-Order}
& Zero-shot
& 7.42$\pm$0.67
& 83.33
& 67.08$\pm$1.71
& 108.36$\pm$13.42 \\

& & Few-shot
& 6.25$\pm$0.62
& 100.00
& 69.62$\pm$1.29
& 110.58$\pm$13.91 \\

& & In-distribution
& 5.83$\pm$0.39
& 100.00
& 70.74$\pm$1.03
& 111.42$\pm$12.84 \\

\midrule

\multirow{9}{*}{85}

& \multirow{3}{*}{Diffusion}
& Zero-shot
& 12.17$\pm$1.11
& 33.33
& 81.62$\pm$3.28
& 108.74$\pm$14.63 \\

& & Few-shot
& 10.75$\pm$0.97
& 58.33
& 84.18$\pm$2.36
& 111.06$\pm$15.14 \\

& & In-distribution
& 10.08$\pm$0.79
& 75.00
& 85.48$\pm$1.86
& 112.14$\pm$14.28 \\

\cmidrule(lr){2-7}

& \multirow{3}{*}{Pairwise}
& Zero-shot
& 11.08$\pm$1.00
& 50.00
& 83.54$\pm$2.57
& 118.83$\pm$15.71 \\

& & Few-shot
& 9.67$\pm$0.89
& 75.00
& 85.72$\pm$1.82
& 121.27$\pm$16.22 \\

& & In-distribution
& 9.00$\pm$0.74
& 83.33
& 86.83$\pm$1.47
& 122.36$\pm$15.08 \\

\cmidrule(lr){2-7}

& \multirow{3}{*}{4th-Order}
& Zero-shot
& 9.75$\pm$0.87
& 75.00
& 85.96$\pm$1.91
& 139.24$\pm$18.36 \\

& & Few-shot
& 8.42$\pm$0.67
& 91.67
& 88.14$\pm$1.43
& 142.17$\pm$18.92 \\

& & In-distribution
& 7.83$\pm$0.58
& 100.00
& 89.31$\pm$1.12
& 143.45$\pm$17.61 \\

\bottomrule
\end{tabular}
}
\end{table*}

\begin{table*}[ht]
\centering
\caption{
\textbf{Cross-city target-coverage planning under different adaptation regimes in London scene.}
We compare zero-shot transfer, few-shot target-city adaptation, and
in-distribution evaluation for increasingly expressive relational energy
models. Results are reported as mean$\pm$std over 12 independent trials.
Compared with the easy scene, higher deployment complexity leads to lower
success rates and a larger number of required TXs, particularly at aggressive
coverage targets.
}
\label{tab:target_coverage_transfer_medium}
\setlength{\tabcolsep}{4.0pt}
\small

\resizebox{0.96\textwidth}{!}{
\begin{tabular}{cllcccc}
\toprule
\textbf{Target JM (\%)}
& \textbf{Method}
& \textbf{Setting}
& \textbf{Req. TXs$\downarrow$}
& \textbf{SR (\%)$\uparrow$}
& \textbf{Best JM (\%)$\uparrow$}
& \textbf{RT (s)$\downarrow$} \\
\midrule

\multirow{9}{*}{35}

& \multirow{3}{*}{Diffusion}
& Zero-shot
& 7.17$\pm$0.72
& 75.00
& 35.92$\pm$1.69
& 72.16$\pm$8.84 \\

& & Few-shot
& 6.25$\pm$0.62
& 91.67
& 37.68$\pm$1.38
& 73.72$\pm$9.18 \\

& & In-distribution
& 5.83$\pm$0.58
& 100.00
& 38.62$\pm$1.15
& 74.35$\pm$8.57 \\

\cmidrule(lr){2-7}

& \multirow{3}{*}{Pairwise}
& Zero-shot
& 6.50$\pm$0.67
& 83.33
& 37.21$\pm$1.51
& 79.64$\pm$9.36 \\

& & Few-shot
& 5.58$\pm$0.51
& 100.00
& 39.17$\pm$1.22
& 81.26$\pm$9.74 \\

& & In-distribution
& 5.17$\pm$0.39
& 100.00
& 40.08$\pm$1.01
& 81.91$\pm$9.02 \\

\cmidrule(lr){2-7}

& \multirow{3}{*}{4th-Order}
& Zero-shot
& 5.83$\pm$0.58
& 91.67
& 38.84$\pm$1.38
& 93.58$\pm$11.42 \\

& & Few-shot
& 5.00$\pm$0.43
& 100.00
& 40.96$\pm$1.08
& 95.47$\pm$11.86 \\

& & In-distribution
& 4.67$\pm$0.49
& 100.00
& 41.93$\pm$0.91
& 96.15$\pm$10.93 \\

\midrule

\multirow{9}{*}{55}

& \multirow{3}{*}{Diffusion}
& Zero-shot
& 9.83$\pm$0.94
& 41.67
& 52.31$\pm$2.64
& 96.24$\pm$12.17 \\

& & Few-shot
& 8.50$\pm$0.80
& 66.67
& 55.47$\pm$2.02
& 98.36$\pm$12.72 \\

& & In-distribution
& 7.92$\pm$0.67
& 83.33
& 56.82$\pm$1.61
& 99.27$\pm$11.86 \\

\cmidrule(lr){2-7}

& \multirow{3}{*}{Pairwise}
& Zero-shot
& 9.00$\pm$0.85
& 58.33
& 54.64$\pm$2.18
& 105.38$\pm$13.42 \\

& & Few-shot
& 7.75$\pm$0.75
& 83.33
& 57.18$\pm$1.66
& 107.76$\pm$13.91 \\

& & In-distribution
& 7.17$\pm$0.58
& 91.67
& 58.43$\pm$1.34
& 108.82$\pm$12.97 \\

\cmidrule(lr){2-7}

& \multirow{3}{*}{4th-Order}
& Zero-shot
& 8.00$\pm$0.74
& 75.00
& 56.91$\pm$1.84
& 124.73$\pm$15.68 \\

& & Few-shot
& 6.92$\pm$0.67
& 91.67
& 59.62$\pm$1.39
& 127.56$\pm$16.11 \\

& & In-distribution
& 6.42$\pm$0.51
& 100.00
& 60.91$\pm$1.12
& 128.74$\pm$15.06 \\

\midrule

\multirow{9}{*}{75}

& \multirow{3}{*}{Diffusion}
& Zero-shot
& 13.58$\pm$1.16
& 16.67
& 71.48$\pm$3.04
& 122.61$\pm$16.08 \\

& & Few-shot
& 11.92$\pm$1.00
& 41.67
& 73.92$\pm$2.32
& 125.47$\pm$16.72 \\

& & In-distribution
& 11.08$\pm$0.90
& 58.33
& 75.31$\pm$1.82
& 126.61$\pm$15.54 \\

\cmidrule(lr){2-7}

& \multirow{3}{*}{Pairwise}
& Zero-shot
& 12.42$\pm$1.08
& 25.00
& 73.64$\pm$2.43
& 134.26$\pm$17.42 \\

& & Few-shot
& 10.83$\pm$0.94
& 58.33
& 75.96$\pm$1.83
& 137.48$\pm$18.06 \\

& & In-distribution
& 10.08$\pm$0.79
& 66.67
& 77.18$\pm$1.47
& 138.75$\pm$16.84 \\

\cmidrule(lr){2-7}

& \multirow{3}{*}{4th-Order}
& Zero-shot
& 10.92$\pm$0.90
& 58.33
& 77.82$\pm$1.68
& 157.94$\pm$20.31 \\

& & Few-shot
& 9.42$\pm$0.67
& 83.33
& 79.86$\pm$1.28
& 161.72$\pm$20.94 \\

& & In-distribution
& 8.75$\pm$0.62
& 91.67
& 81.06$\pm$1.04
& 163.18$\pm$19.47 \\

\bottomrule
\end{tabular}
}
\end{table*}

\begin{table*}[ht]
\centering
\caption{
\textbf{Cross-city target-coverage planning under different adaptation regimes in Shanghai scene.}
We compare zero-shot transfer, few-shot target-city adaptation, and
in-distribution evaluation for increasingly expressive relational energy
models. Results are reported as mean$\pm$std over 12 independent trials.
The high-complexity scene exhibits the largest transfer gap and the strongest
degradation in success rate at demanding target coverage levels.
Few-shot adaptation recovers a substantial fraction of the lost performance,
while higher-order relational models remain consistently advantageous.
}
\label{tab:target_coverage_transfer_high}
\setlength{\tabcolsep}{4.0pt}
\small

\resizebox{0.96\textwidth}{!}{
\begin{tabular}{cllcccc}
\toprule
\textbf{Target JM (\%)}
& \textbf{Method}
& \textbf{Setting}
& \textbf{Req. TXs$\downarrow$}
& \textbf{SR (\%)$\uparrow$}
& \textbf{Best JM (\%)$\uparrow$}
& \textbf{RT (s)$\downarrow$} \\
\midrule

\multirow{9}{*}{25}

& \multirow{3}{*}{Diffusion}
& Zero-shot
& 7.92$\pm$0.79
& 66.67
& 25.84$\pm$1.78
& 89.73$\pm$10.86 \\

& & Few-shot
& 6.92$\pm$0.67
& 83.33
& 27.62$\pm$1.43
& 91.64$\pm$11.24 \\

& & In-distribution
& 6.42$\pm$0.51
& 91.67
& 28.51$\pm$1.19
& 92.48$\pm$10.52 \\

\cmidrule(lr){2-7}

& \multirow{3}{*}{Pairwise}
& Zero-shot
& 7.25$\pm$0.75
& 75.00
& 27.14$\pm$1.59
& 98.16$\pm$11.73 \\

& & Few-shot
& 6.25$\pm$0.62
& 91.67
& 29.04$\pm$1.23
& 100.27$\pm$12.14 \\

& & In-distribution
& 5.83$\pm$0.58
& 100.00
& 30.02$\pm$1.01
& 101.14$\pm$11.32 \\

\cmidrule(lr){2-7}

& \multirow{3}{*}{4th-Order}
& Zero-shot
& 6.50$\pm$0.67
& 83.33
& 28.93$\pm$1.41
& 115.62$\pm$13.98 \\

& & Few-shot
& 5.58$\pm$0.51
& 100.00
& 31.05$\pm$1.09
& 118.16$\pm$14.46 \\

& & In-distribution
& 5.17$\pm$0.39
& 100.00
& 32.11$\pm$0.90
& 119.23$\pm$13.41 \\

\midrule

\multirow{9}{*}{40}

& \multirow{3}{*}{Diffusion}
& Zero-shot
& 10.58$\pm$0.90
& 25.00
& 37.12$\pm$2.51
& 112.48$\pm$14.13 \\

& & Few-shot
& 9.25$\pm$0.87
& 50.00
& 39.74$\pm$2.02
& 114.93$\pm$14.68 \\

& & In-distribution
& 8.67$\pm$0.78
& 66.67
& 41.03$\pm$1.66
& 115.96$\pm$13.72 \\

\cmidrule(lr){2-7}

& \multirow{3}{*}{Pairwise}
& Zero-shot
& 9.67$\pm$0.89
& 41.67
& 39.16$\pm$2.16
& 122.36$\pm$15.21 \\

& & Few-shot
& 8.42$\pm$0.67
& 66.67
& 41.87$\pm$1.69
& 125.18$\pm$15.74 \\

& & In-distribution
& 7.83$\pm$0.58
& 83.33
& 43.12$\pm$1.37
& 126.31$\pm$14.63 \\

\cmidrule(lr){2-7}

& \multirow{3}{*}{4th-Order}
& Zero-shot
& 8.58$\pm$0.79
& 58.33
& 41.86$\pm$1.76
& 143.85$\pm$17.62 \\

& & Few-shot
& 7.42$\pm$0.67
& 83.33
& 44.48$\pm$1.34
& 147.16$\pm$18.13 \\

& & In-distribution
& 6.83$\pm$0.58
& 91.67
& 45.72$\pm$1.09
& 148.52$\pm$16.88 \\

\midrule

\multirow{9}{*}{55}

& \multirow{3}{*}{Diffusion}
& Zero-shot
& 14.08$\pm$1.11
& 8.33
& 49.62$\pm$3.37
& 137.94$\pm$17.84 \\

& & Few-shot
& 12.42$\pm$1.08
& 25.00
& 52.13$\pm$2.72
& 141.06$\pm$18.46 \\

& & In-distribution
& 11.58$\pm$0.90
& 33.33
& 53.72$\pm$2.21
& 142.41$\pm$17.12 \\

\cmidrule(lr){2-7}

& \multirow{3}{*}{Pairwise}
& Zero-shot
& 13.00$\pm$1.04
& 16.67
& 51.28$\pm$2.92
& 149.63$\pm$19.24 \\

& & Few-shot
& 11.42$\pm$0.90
& 41.67
& 54.42$\pm$2.13
& 153.08$\pm$19.86 \\

& & In-distribution
& 10.58$\pm$0.79
& 50.00
& 55.81$\pm$1.73
& 154.46$\pm$18.41 \\

\cmidrule(lr){2-7}

& \multirow{3}{*}{4th-Order}
& Zero-shot
& 11.42$\pm$0.90
& 33.33
& 54.91$\pm$2.15
& 175.42$\pm$21.73 \\

& & Few-shot
& 9.92$\pm$0.79
& 58.33
& 57.43$\pm$1.61
& 179.61$\pm$22.46 \\

& & In-distribution
& 9.17$\pm$0.72
& 75.00
& 58.72$\pm$1.31
& 181.18$\pm$20.82 \\

\bottomrule
\end{tabular}
}
\end{table*}

\clearpage
\section{Ablation Studies}
\label{app:ablation}

\subsection{Potential Model Capacity}
\label{app:ablation_capacity}

We examine whether increasing the capacity of the learned potential
systematically improves utility prediction.
We vary the hidden and hypernetwork dimensions while keeping the remaining
training configuration fixed.

\begin{table}[ht]
\centering
\caption{
\textbf{Potential-model capacity ablation.}
Coverage prediction errors under different hidden and hypernetwork dimensions.
Values are reported as mean$\pm$standard deviation; lower prediction error is
better.
}
\label{tab:ablation_hidden_hyper_scene01}
\resizebox{\textwidth}{!}{
\begin{tabular}{cccccccc}
\toprule
Hidden & Hyper & Runtime (s) & Joint (\%) & Pathloss (\%) &
Throughput (\%) & SINR (\%) & RSRP (\%) \\
\midrule
64  & 128 & $1.08 \pm 0.13$ & $20.58 \pm 3.48$ & $20.91 \pm 3.32$ & $34.83 \pm 5.00$ & $33.61 \pm 5.20$ & $29.70 \pm 4.39$ \\
128 & 128 & $1.08 \pm 0.14$ & $19.75 \pm 3.81$ & $20.03 \pm 3.79$ & $33.36 \pm 4.27$ & $32.21 \pm 4.27$ & $28.59 \pm 3.88$ \\
256 & 64  & $1.09 \pm 0.11$ & $\mathbf{17.52 \pm 1.69}$ & $\mathbf{18.58 \pm 2.28}$ & $\mathbf{27.70 \pm 1.75}$ & $\mathbf{25.88 \pm 1.84}$ & $\mathbf{24.22 \pm 1.77}$ \\
256 & 128 & $1.09 \pm 0.10$ & $18.55 \pm 2.92$ & $19.51 \pm 2.55$ & $29.76 \pm 4.28$ & $28.14 \pm 5.02$ & $25.80 \pm 3.52$ \\
256 & 256 & $1.05 \pm 0.09$ & $19.05 \pm 2.83$ & $19.58 \pm 2.26$ & $31.07 \pm 3.83$ & $29.70 \pm 4.60$ & $26.48 \pm 3.29$ \\
\bottomrule
\end{tabular}
}
\end{table}

Increasing model capacity does not yield monotonic gains.
The $256/64$ configuration produces the lowest error across all five service
targets, while further enlarging the hypernetwork degrades prediction despite
similar runtime.
This suggests that utility quality depends on how capacity is allocated rather
than parameter count alone.

\subsection{Inference Budget and Interaction Order}
\label{app:ablation_inference}

We next vary relational order and inference budget while keeping the remaining
planning procedure fixed.

\begin{table}[t]
\centering
\caption{
\textbf{Interaction order and inference-budget ablation.}
Simulator-verified JM coverage (\%) under different maximum interaction orders,
particle populations, and sampling steps.
All settings use the same planning protocol on the New York scene with
$N=3$ TXs.
}
\label{tab:ablation_interaction_order}
\small
\setlength{\tabcolsep}{6pt}

\begin{tabular}{c|c||cccc}
\toprule
\multicolumn{2}{c||}{Inference Budget}
&
\multicolumn{4}{c}{Maximum Interaction Order}
\\
\cmidrule(lr){1-2}
\cmidrule(lr){3-6}
Particles & Steps & Unary & Pairwise & 3rd-Order & 4th-Order \\
\midrule

\multirow{5}{*}{10}
& 20  & 38.90 & 40.15 & 41.47 & \textbf{42.82} \\
& 40  & 39.62 & 40.91 & 42.28 & \textbf{43.67} \\
& 60  & 40.18 & 41.52 & 42.94 & \textbf{44.40} \\
& 80  & 40.52 & 41.89 & 43.35 & \textbf{44.84} \\
& 100 & 40.68 & 42.08 & 43.55 & \textbf{45.06} \\
\midrule

\multirow{5}{*}{15}
& 20  & 39.58 & 40.91 & 42.31 & \textbf{43.74} \\
& 40  & 40.31 & 41.69 & 43.14 & \textbf{44.63} \\
& 60  & 40.88 & 42.31 & 43.82 & \textbf{45.36} \\
& 80  & 41.23 & 42.69 & 44.23 & \textbf{45.81} \\
& 100 & 41.40 & 42.88 & 44.44 & \textbf{46.04} \\
\midrule

\multirow{5}{*}{20}
& 20  & 40.19 & 41.58 & 43.04 & \textbf{44.55} \\
& 40  & 40.93 & 42.37 & 43.89 & \textbf{45.45} \\
& 60  & 41.51 & 43.01 & 44.58 & \textbf{46.19} \\
& 80  & 41.87 & 43.40 & 45.01 & \textbf{46.65} \\
& 100 & 42.05 & 43.60 & 45.22 & \textbf{46.88} \\
\midrule

\multirow{5}{*}{25}
& 20  & 40.72 & 42.16 & 43.69 & \textbf{45.27} \\
& 40  & 41.47 & 42.96 & 44.55 & \textbf{46.18} \\
& 60  & 42.06 & 43.61 & 45.25 & \textbf{46.93} \\
& 80  & 42.43 & 44.01 & 45.69 & \textbf{47.39} \\
& 100 & 42.61 & 44.21 & 45.90 & \textbf{47.63} \\
\bottomrule
\end{tabular}
\end{table}

Across the tested budget grid, higher interaction order produces a consistent
advantage, and the gap becomes more pronounced as the inference budget grows.
Increasing particles or sampling steps also improves coverage within each
interaction order, indicating that representation capacity and search effort
provide complementary gains in this setting.
Importantly, the improvement from higher-order structure persists at matched
inference budgets, so it cannot be attributed solely to stronger search.

\section{Aggregate Planning Diagnostics}
\label{app:planning_diagnostics}

\subsection{Physical Scaling and Service Coupling}
\label{app:diagnostics_scaling}

\begin{figure}[ht]
    \centering
    \includegraphics[width=0.98\linewidth]{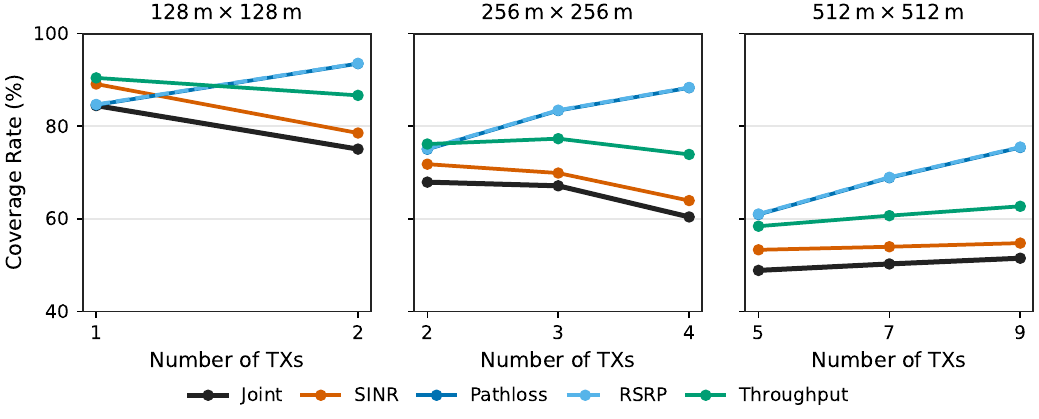}
    \caption{
    \textbf{Service trade-offs under increasing TX cardinality.}
    Each point aggregates six cities, nine planners, four utility representations,
    and five random seeds.
    Additional TXs improve serving-link coverage, while the resulting SINR and JM
    behavior depends on the balance between coverage gain and co-channel interference
    at each deployment scale.
    }
    \label{fig:tx_metric_tradeoff}
\end{figure}

Figure~\ref{fig:tx_metric_tradeoff} summarizes how the service metrics evolve
with TX cardinality across representative deployment scales.
Increasing the number of TXs generally improves serving-link coverage, as reflected
by path-loss and SS-RSRP coverage.
The behavior of SINR and JM coverage is more strongly scene dependent:
in compact deployments, additional co-channel interference can offset the gain
in serving-link coverage, whereas larger regions can remain predominantly
coverage limited over the evaluated TX range.
The resulting non-monotonicity motivates treating multi-TX deployment as a
coupled set-planning problem rather than reasoning about TX locations independently.

\subsection{Sensitivity to Utility Representation}
\label{app:diagnostics_reward}

\begin{figure}[ht]
    \centering
    \includegraphics[width=0.95\linewidth]{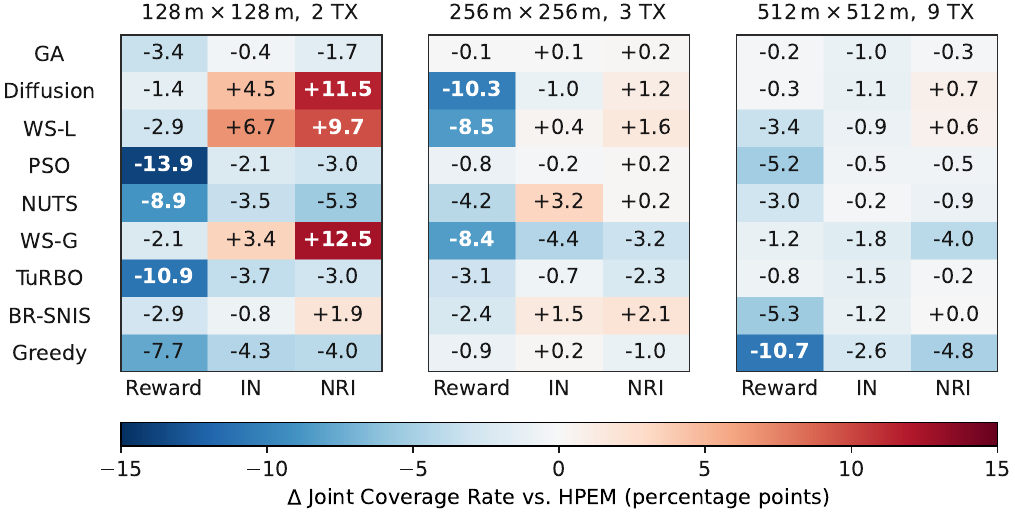}
    \caption{
    \textbf{Sensitivity of planning quality to utility representation.}
    Each cell reports the mean change in simulator-verified JM coverage
    (percentage points) when HPEM is replaced by the pooled Regressor, IN, or NRI
    under the same planner.
    Results are averaged over five random seeds.
    Positive values favor the alternative representation; negative values favor HPEM.
    }
    \label{fig:reward_sensitivity}
\end{figure}

Figure~\ref{fig:reward_sensitivity} isolates the effect of utility representation
while holding the planner fixed.
The preferred representation changes across deployment regimes:
pairwise or latent-relational models can be competitive at low TX cardinality,
while higher-order structure becomes increasingly useful in more strongly coupled
settings.
The variation across planner rows further shows that representation quality and
search dynamics are not independent factors; the same reward replacement can
help one planner while providing little gain to another.

\subsection{Quality, Efficiency, and Stability}
\label{app:diagnostics_efficiency}

\begin{figure}[ht]
    \centering
    \includegraphics[width=0.98\linewidth]{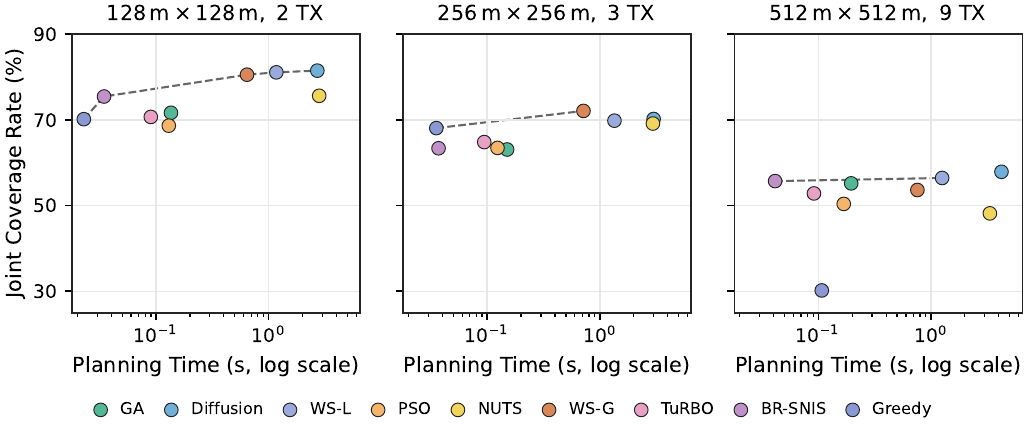}
    \caption{
    \textbf{Quality--efficiency trade-off across planning methods.}
    The horizontal axis reports median planning time on a logarithmic scale and
    the vertical axis reports mean simulator-verified JM coverage.
    Results aggregate six cities, four utility representations, and five random
    seeds for representative deployment regimes.
    Dashed curves connect the empirical Pareto-optimal methods.
    }
    \label{fig:quality_efficiency_pareto}
\end{figure}

Planning quality alone does not determine the usefulness of a search method.
Figure~\ref{fig:quality_efficiency_pareto} compares simulator-verified JM coverage
against planning time and shows that the empirical Pareto frontier changes with
deployment scale.
Fast local procedures can remain competitive in simpler settings, while more
structured stochastic exploration becomes increasingly valuable as the deployment
problem becomes more strongly coupled.

Figure~\ref{fig:quality_seed_stability} provides a complementary view of
repeatability.
Stochastic search does not necessarily imply unstable outcomes: several sampling
methods combine high average JM coverage with relatively small variation across
random seeds.
Together, these results motivate evaluating planning methods jointly in terms of
quality, cost, and stability rather than using a single performance axis.

\end{document}